\documentclass[journal]{IEEEtran}
\usepackage{algorithm}
\usepackage{algorithmic}
\usepackage{caption}
\usepackage{color}
\usepackage{subcaption}
\usepackage{authblk}

\usepackage{amsmath,amssymb,graphicx,epstopdf,textpos,rotating}

\usepackage[square,sort,comma,numbers,compress]{natbib}
\usepackage{hyperref}
\hypersetup{colorlinks=True,linkcolor=black,anchorcolor=black,citecolor=black,urlcolor=blue}
\usepackage{url}
\newcommand{\batch}{\mathcal{B}}

\newcommand{\expect}{\mathbb{E}}
\newtheorem{assumption}{\bf{Assumption}}

\newtheorem{theorem}{\bf{Theorem}}
\newtheorem{definition}{\bf{Definition}}
\allowdisplaybreaks

\ifCLASSINFOpdf
\else
\fi
\begin{document}

\title{Drill the Cork of Information Bottleneck by Inputting the Most Important Data}

\author{Xinyu~Peng,~\IEEEmembership{}
	    Jiawei~Zhang,~\IEEEmembership{}
        Fei-Yue~Wang,~\IEEEmembership{Fellow,~IEEE}%
        ~and~Li~Li,~\IEEEmembership{Fellow,~IEEE}
\thanks{Manuscript received; accepted; Date of publication; This work was supported in part by. \emph{(Corresponding author: Li Li.)}}}

\maketitle

\begin{abstract}
Deep learning has become the most powerful machine learning tool in the last decade. However, how to efficiently train deep neural networks remains to be thoroughly solved. The widely used minibatch stochastic gradient descent (SGD) still needs to be accelerated. As a promising tool to better understand the learning dynamic of minibatch SGD, the information bottleneck (IB) theory claims that the optimization process consists of an initial fitting phase and the following compression phase. Based on this principle, we further study typicality sampling, an efficient data selection method, and propose a new explanation of how it helps accelerate the training process of the deep networks. We show that the fitting phase depicted in the IB theory will be boosted with a high signal-to-noise ratio of gradient approximation if the typicality sampling is appropriately adopted. Furthermore, this finding also implies that the prior information of the training set is critical to the optimization process and the better use of the most important data can help the information flow through the bottleneck faster. Both theoretical analysis and experimental results on synthetic and real-world datasets demonstrate our conclusions.
\end{abstract}

\begin{IEEEkeywords}
Machine learning, minibatch stochastic gradient descent, information bottleneck theory, typicality sampling.
\end{IEEEkeywords}

%
\IEEEpeerreviewmaketitle

\section{Introduction}

\IEEEPARstart{D}{eep} learning has achieved the state-of-the-art for many machine learning tasks ranging from computer vision to natural language processing and reinforcement learning \citep{ren2015faster,krizhevsky2012imagenet,bahdanau2014neural,oord2016wavenet,mnih2015human}, but the large training data size \citep{hinton2012deep,deng2009imagenet,lin2017stardata} and high model complexity \citep{he2016deep,lin2013network,devlin2018bert} used in these successful applications make the training time becomes unmanageable \cite{goyal2017accurate,you2019reducing}. Although minibatch SGD enables the training of most deep learning models and often generalizes better than the other algorithms, it still needs to be accelerated to facilitate the development progress.

As a general framework for the information extraction problem in information theory and modern statistics, the IB theory \citep{tishby2000information} has promoted the analysis \citep{khadivi2016flow,achille1706emergence,saxe2019information} and improvements \citep{alemi2016deep,achille2018information} of deep learning systems by giving a fresh perspective of describing the training dynamic of minibatch SGD. It considers the training of deep networks as an information extraction process that squeezes the most relevant information about the label from the input, and claims that each layer achieves a trade-off between the mutual information about the input and the label. The training dynamics of layers can be visualized as curves in the information plane: the first plane that the mutual information about the input and the label can be quantified by its coordinates \citep{tishby2015deep}. Further investigation in this information plane shows that the training process of minibatch SGD often consists of two phases: the initial fitting phase with a high gradient signal-to-noise ratio (SNR) and a subsequent compression phase with a low gradient SNR \citep{shwartz2017opening}. Thus, different architecture and training algorithms can be compared by the pattern of information curves \citep{kolchinsky2019nonlinear,li2019information,cheng2019utilizing}.


Follow up this idea, we revisit the typicality sampling scheme proposed in previous work \citep{peng2019accelerating}. Different from the works that concentrate on the update rule of model parameters \citep{Qian1999On,Johnson2013Accelerating,Roux2012A} or adaptive hyperparameters \citep{Duchi2011Adaptive,Zeiler2012ADADELTA,Kingma2014Adam}, the typicality sampling scheme improves the convergence rate of conventional minibatch SGD by building a more efficient batch selection method. It has been hypothesized that not all samples are of the same importance \citep{Swayamdipta2020dataset}. In each iteration of the parameters update, only a group of most important training samples, namely the typical samples, need to be focused to find the correct search direction. Increasing the probability of these samples appearing in the batch can accelerate conventional minibatch SGD. Both theoretical analysis and experimental results verified the improvement of the resulting typical batch SGD in the averaged convergence rate due to the small noise in gradient approximation, but there is no further discussion about the dynamic performance of typicality sampling in different phases of the learning and how it achieves the inside improvement. Therefore, we employ the IB theory to give a new explanation of how typicality sampling helps the training.

From the viewpoint of the IB theory, we inquire into the learning process and compare the training dynamics between conventional minibatch SGD and typical batch SGD in the information plane. Both theoretical analysis and experimental results show that the typicality sampling scheme accelerates conventional minibatch SGD by mainly speeding up the fitting phase with more informative gradient features, i.e., a higher SNR of gradient approximation. Furthermore, we find that by simply utilizing the prior information of the training set, one can expedite the underlying information extraction process and approach the lower bound of IB theory faster. The rest of this paper is arranged as follows. In Section \ref{section2}, we review the framework of the IB theory and the typicality sampling scheme. In Section \ref{section3}, we theoretically prove that the typicality sampling scheme speeds up the fitting phase. In Section \ref{section4}, we set up numerical experiments on various datasets and validate our analysis with empirical results. The whole paper is concluded in Section \ref{section5}.

\section{Methods}
\label{section2}

\subsection{Information Bottleneck Theory}
\label{subsection21}
The IB theory was first presented by \cite{tishby2000information} to tackle the information extraction problem in modern statistics and information theory. The target of this kind of problem is to get the minimal sufficient statistic $T(X)$, which captures all relevant information that one variable $X\in \mathcal{X}$ contains about another variable $Y \in \mathcal{Y}$ while preserving as little information about $X$ as possible. A general definition of minimal sufficient statistic is as follows
\begin{definition}[\citep{Shamir2010Learning}]
	\label{def_of_TX}
	The statistic $T(X)$ is a minimal sufficient statistic if and only if it is the solution of the following constrained optimization problem
	\begin{eqnarray}
		\min_{S\left(X\right):I\left(S\left(X\right);Y\right)=I\left(X;Y\right)} I\left(S\left(X\right);X\right), \label{eq1}
	\end{eqnarray}
	where $X$ and $Y$ are two given random variables, $I(X;Y)=\expect [ \log \frac{p(X,Y)}{p(X)p(Y)}]$ is their mutual information.
\end{definition}

The IB theory provides an extensible framework to find the minimal sufficient statistic $T(X)$. This approach suggests solving a Lagrange relaxation of \eqref{eq1} that is formulated as the form
\begin{eqnarray}
	\min_{p(t \mid x),p(y \mid t),p(t)} \{ I(X;T)-\beta I(T;Y)  \}, \label{eq2}
\end{eqnarray}
where $t\in T$ is the compressed representation of $x \in X$. If the joint distribution $P(X,Y)$ is known, one can increase the Lagrange multiplier $\beta \rightarrow \infty$ to get the approximate minimal sufficient statistic $T(X)$; if the joint distribution $P(X,Y)$ is unknown, then $\beta$ determines a trade-off between the accuracy of Lagrange relaxation of \eqref{eq1} and the accuracy of the computation of mutual information with finite samples. Thus, it should be set to a value that gives the best trade-off.

With the rapid development of deep learning algorithms, many researchers focus on revealing the mechanisms behind the deep neural network (DNN) structure \citep{alain2016understanding,zhang2016understanding,nakkiran2019deep}, which usually involves finding an approximate solution of the following empirical risk minimization (ERM) problem
\begin{eqnarray}
\min_{\theta \in \Theta} J(\theta) =  \frac{1}{N}\sum_{i=1}^N \ell\left(f \left(x_i ;\theta \right),y_i\right),
\label{eq3}
\end{eqnarray}
where $f(x_i;\theta)$ is the output predicted by the DNN with parameter $\theta$ and input $x_i$. The loss function $\ell(\cdot)$ measures the difference between the prediction and the true label $y_i$, and $J(\theta)$ describes the overall losses of the network on the whole training set $X$. The work in \citep{tishby2015deep} first used the idea of the IB theory to describe the above training process of DNN. The authors made a hypothesis that each layer is a compressed representation variable $T$ and solves an IB optimization problem \eqref{eq2} during the training process. In this case, the Lagrange multiplier $\beta$ controls a trade-off between the accuracy of prediction, $I(T;Y)$, and the compression of the input information, $I(X;T)$. Therefore, the draining dynamic of DNN can be visualized as curves in the information plane: a plane shows the mutual information value between any variable and the input variable $X$ and the label variable $Y$. Each curve represents the training process of a specific layer in DNN.

The authors in \citep{shwartz2017opening} further investigated the case where the DNN is trained by minibatch SGD. Instead of evaluating the true gradient $\nabla J(\theta)$ of objective function $J(\theta)$ in \eqref{eq3} directly, minibatch SGD samples a mini-batch $\batch$ of training samples from $X$ by simple random sampling (SRS) scheme \citep{lohr2009sampling} on each iteration $k$ and performs the update rule as the form
\begin{equation}
\label{eq4}
\begin{aligned}
\theta_{k+1} 
~=~& \theta_k - \eta_k \frac{1}{m}\sum_{i \in\batch}\nabla J_{i}(\theta_k)\\
~=~& \theta_k - \eta_k\nabla J_\batch (\theta_k),
\end{aligned}
\end{equation}
where $m$ is batch size and $\nabla J_{i}(\theta)$ denotes $\nabla\ell(f (x_i ;\theta),y_i)$. 

\begin{figure}[!t]
	\includegraphics[width=0.47\textwidth, scale=1]{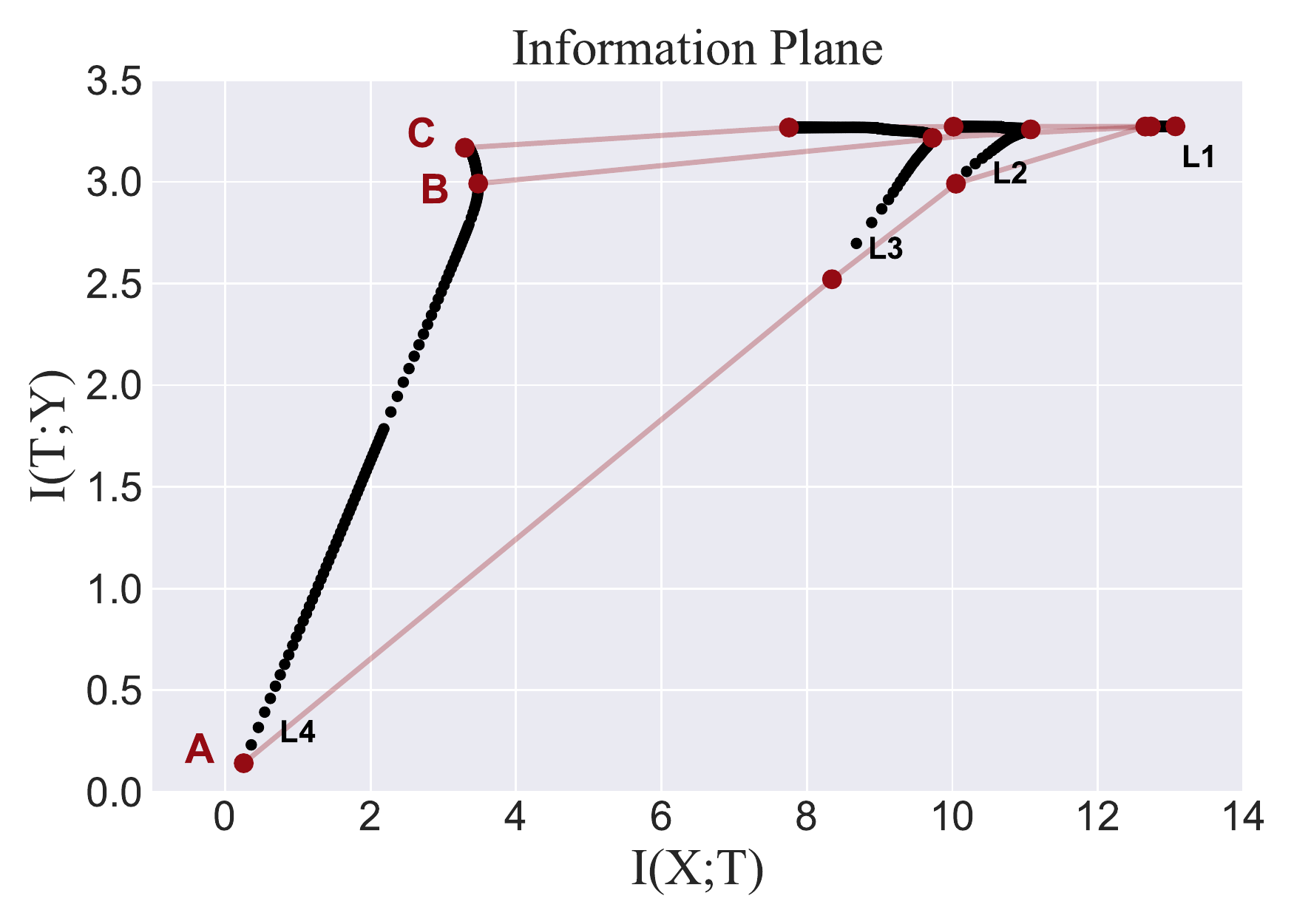}
	\caption{The evolution of layers in the information plane during the minibatch SGD optimization. A corresponds to the initial point, B corresponds to the transition point and C corresponds to the final point. Same points of different layers are connected by the red thin lines}
	\label{fig:0}
\end{figure}

Figure \ref{fig:0} shows the evolution of layers in the information plane during the minibatch SGD optimization. The x-label $I(X;T)$ represents mutual information about the input. The larger the value, the more $T$ compresses the input $X$. The y-label $I(T;Y)$ represents mutual information about the label. The larger the value, the more information $T$ contains about the label $Y$. Since the lower layer includes more information about both the input and the label at the initial point, the curves in the information plane are located from right to left in order from the first layer to the final layer. The most striking phenomenon shown in Figure \ref{fig:0} is that all curves exhibit the same two-phase pattern: the initial fitting phase marked by the rapid growth of $I(T;Y)$ from initial point A to transition point B, and the following compression phase where $I(X;T)$ decreases slowly from transition point B to final point C. 


They also found that these two optimization phases are closely connected to and can be explained by the gradient SNR value, which is defined as
\begin{equation}
	\label{def:gsnr}
	\textsc{GSNR} := \frac{\left\| \expect\left[\nabla J_\batch (\theta_k)\right] \right\|}{{\mathrm{STD} \left[\nabla J_\batch (\theta_k)\right]}},
\end{equation}
In the fitting phase, the mean of the gradient is much larger than the standard deviation of the gradient. The resulting high gradient SNR increases the mutual information about the label and leads to a small training error. In the compression phase, the gradient mean becomes relatively smaller than its standard deviation, yielding a low gradient SNR that searches more compressed representation for each layer.

Therefore, the IB theory can serve as a promising tool to analyze the performance of different training settings, compare different deep learning models, and guide architecture design by comparing the pattern of curves in the information plane.

\subsection{Typicality Sampling Scheme}
\label{subsection22}

The typicality sampling scheme was first presented in previous work \citep{peng2019accelerating} for accelerating conventional minibatch SGD. Researchers have pointed out that one cause for the long training time of conventional minibatch SGD is the applied SRS scheme. This inefficient sampling method treats all training samples as of the same importance and yields a gradient estimator with large noise that impedes the approaching to the optimal solution, slowing down the convergence rate \citep{de2016big,friedlander2012hybrid,alain2015variance,zhao2014accelerating}. 
One of the key concepts proposed in work \citep{peng2019accelerating} is typical samples. The authors suggested that these samples are the most important part of the whole training set in gradient estimation, as they can reflect the crucial pattern of the optimization space and roughly approximate the true gradient. In work \citep{peng2019accelerating}, a formal definition of typical samples was given as follows
\begin{definition}
	\label{eq5}
	A group of training samples is called typical samples if the overall gradients of the subset $\mathcal{H}$, which consists of all these samples, satisfy:
	\begin{align}
	\frac{1}{N}\sum_{i\in \mathcal{H}}\nabla {{J}_{i}}\left( \theta  \right) = \nabla J\left( \theta  \right)\,\,\,\,\,\, \text{for any} \,\,\,\,\,\theta \in \Theta.
	\end{align}
	Subset $\mathcal{H}$ is called highly representative subset.
\end{definition}

In this view, it is natural that one can speed up minibatch SGD by letting the typical samples which provide the most informative gradient features dominate the mini-batch $\batch$ to reduce the gradient noise. Therefore, the typicality sampling scheme consisting of three steps was proposed:
\begin{enumerate}
	\item partition subset $\mathcal{H}$ of all typical samples from training set and group the rest samples as subset $\mathcal{L}$.
	\item set proper sampling density $n_1/N_1\geq m/N$  so that the typical samples are with greater probabilities to be selected.
	\item on each iteration $k$, sample sub-batch $h_k$ with size $n_1$ and sub-batch $l_k$ with size $n_2$ from subset $\mathcal{H}$ with size $N_1$ and subset $\mathcal{L}$ with size $N_2$ by SRS , respectively, and combine them to get mini-batch $\batch$.
\end{enumerate} 

The authors also analyzed the convergence of the resulting typical batch SGD. Both theoretical analysis and experimental results proved that typical batch SGD achieves a faster convergence rate than conventional minibatch SGD. But it still lacks a deeper understanding of the inside performance of typical batch SGD, especially at which stage of the learning dynamics that typicality sampling plays the most crucial role and what the mechanism is behind this improvement. To this end, we try to apply the IB theory to propose a new explanation of the performance of typicality sampling.

\section{Theoretical Analysis}
\label{section3}

We start by making the following assumption
\begin{assumption}
	\label{assumption1}
	We assume that at each iteration $k$, the averaged squared L2 norm of samples gradient in subset $\mathcal{H}$ and $\mathcal{L}$ satisfy
	\begin{equation}
		\underset{i\in \mathcal{H}}{\mathop \sum }\,\frac{ {\|\nabla {{J}_{i}}{( {{\theta }_{k}})}\|}^{2}}{N_1}   =   \underset{j\in \mathcal{L}}{\mathop \sum }\,\frac{ {\|\nabla {{J}_{j}}{( {{\theta }_{k}})}\|}^{2}}{N_2}
	\end{equation}
\end{assumption}
This assumption implies that there is no significant difference in the squared L2 norm of the gradient of the samples in subset $\mathcal{H}$ and $\mathcal{L}$, respectively. It is rational to make this assumption since we only rely on the gradient direction to demarcate highly representative subset $\mathcal{H}$ from the whole training set. Furthermore, we believe this assumption is independent of Definition \ref{eq5} of typical samples and cannot be derived from it directly. In Section \ref{section4}, we will use numerical examples to verify this assumption.

Together with Definition \ref{eq5} of typical samples, we can prove that the gradient SNR of typical batch SGD (denoted as $\textsc{GSNR}_{\textsc{TS}}$) is larger than the gradient SNR of conventional minibatch SGD (denoted as $\textsc{GSNR}_{\textsc{SRS}}$).
\begin{theorem}
	\label{theorem1}
	Suppose Assumption \ref{assumption1} holds. Suppose further $n_1$ is large enough that the typicality sampling density satisfies $n_1/N_1 \geq m/N$. Then at each iteration $k$, we have
	\begin{eqnarray}
		\textsc{GSNR}_{\textsc{TS}} \geq \textsc{GSNR}_{\textsc{SRS}}
	\end{eqnarray}
\end{theorem}
\begin{IEEEproof}
	We begin by defining random variable $$ Z_i:=\left\{ \begin{aligned} 1  &~~~~ i \in \batch \\ 0 &~~~ otherwise \end{aligned} \right.$$ for all training samples. Based on this definition, we can derive the expression of the gradient mean ${\expect \left[\nabla J_\batch (\theta_k)\right]}_{\textsc{SRS}}$ for conventional minibatch SGD
	\begin{equation}
		\begin{aligned}
			&{\expect \left[\nabla J_\batch (\theta_k)\right]}_{\textsc{SRS}}\\
			&=\expect \left[ \sum_{i=1}^{N} Z_i \frac{\nabla J_{i}(\theta_k)}{m} \right]
			=\sum_{i=1}^{N} \expect \left[ Z_i \right] \frac{\nabla J_{i}(\theta_k)}{m}=\nabla J(\theta_k)~~~~~ \label{eq7} 
		\end{aligned}
	\end{equation}
	where the third equation uses the fact that $\expect \left[ Z_i \right]={m}/{N}$. Similarly, we can also derive the variance of the gradient, denoted as ${\mathrm{V}\left[\nabla J_\batch (\theta_k)\right]}_{\textsc{SRS}}$, for conventional minibatch SGD
	\begin{equation}
		\begin{aligned}
			&{\mathrm{V}\left[\nabla J_\batch (\theta_k)\right]}_{\textsc{SRS}}\\
			&=\frac{1}{m^2}\mathrm{Cov}\left[\sum_{i=1}^{N} Z_i \nabla J_{i}(\theta_k), \sum_{j=1}^{N} Z_j \nabla J_{j}(\theta_k)    \right] \\
			&=\frac{1}{m^2}\Bigg[\sum_{i=1}^{N} \nabla J_{i} (\theta_k)^2 \mathrm{V}(Z_i)\\
			&~~~~~~~~~~~~~~~~~~~~~~+\sum_{i=1}^{N}\sum_{j \neq i}\nabla J_{i} (\theta_k)\nabla J_{j} (\theta_k)\mathrm{Cov}(Z_i, Z_j)  \Bigg]  \\
			&=\frac{1}{mN(N-1)}\left(1-\frac{m}{N}\right)\\
			&~~~~~~~~~~~~~~~~~~~~*\left[ N\sum_{i=1}^{N}{\left\| \nabla J_{i}(\theta_k)\right\|}^2  + {\left\| \sum_{i=1}^{N}\nabla J_{i}(\theta_k)\right\|}^2\right]
		\end{aligned}\label{eq8}
	\end{equation}
	where the first equation uses the definition of variance and the third equation applies the result of Lemma 2 in \citep{peng2019accelerating}.
	
	Proceeding as the proof above, we can also get the expression of gradient mean ${\expect \left[\nabla J_\batch (\theta_k)\right]}_{\textsc{TS}}$ and its variance, denoted as ${\mathrm{V}\left[\nabla J_\batch (\theta_k)\right]}_{\textsc{TS}}$, for typical batch SGD. First we define random variables $$ {Z}_i^h:=\left\{ \begin{aligned} 1  &~~~~ i \in h_k \\ 0 &~~~ otherwise \end{aligned} \right. ~~\text{and}~~  {Z}_j^l:=\left\{ \begin{aligned} 1  &~~~~ j \in l_k \\ 0 &~~~ otherwise \end{aligned} \right.$$ for the training samples in subset $\mathcal{H}$ and $\mathcal{L}$, respectively. Then we can obtain
	\begin{align}
	 &{\expect \left[\nabla J_\batch (\theta_k)\right]}_{\textsc{TS}}\nonumber\\
	 &~~~~~~~~~~= \expect\left[ \underset{i\in \mathcal{H}}{\mathop \sum }\,{{Z}_{i}^h}\frac{\nabla {{J}_{i}}( {{\theta }_{k}})}{m}+\underset{j\in \mathcal{L}}{\mathop \sum }\,{{Z}_{j}^l}\frac{\nabla {{J}_{j}}({{\theta }_{k}} )}{m} \right]~~~~~~~~~~~~~~~~~~~~~~~~~~~~~~~~~~ \nonumber\\ 
	 &~~~~~~~~~~=\underset{i\in \mathcal{H}}{\mathop \sum }\,\frac{{{n}_{1}}}{{{N}_{1}}}\frac{\nabla {{J}_{i}}( {{\theta }_{k}})}{m}+\underset{j\in \mathcal{L}}{\mathop \sum }\,\frac{{{n}_{2}}}{{{N}_{2}}}\frac{\nabla {{J}_{j}}( {{\theta }_{k}} )}{m}\nonumber\\   
	 &~~~~~~~~~~=\underset{i\in \mathcal{H}}{\mathop \sum }\frac{{{n}_{1}}}{{{N}_{1}}}\frac{\nabla {{J}_{i}}( {{\theta }_{k}})}{m}=\frac{n_1 N}{N_1 m} \nabla J( {{\theta }_{k}}),\label{eq9}
	\end{align}
     where the first equation follows from the framework of typicality sampling scheme and the third equation uses Definition \ref{eq5} of typical samples. 
	 \begin{align}
	 &{\mathrm{V}\left[\nabla J_\batch (\theta_k)\right]}_{\textsc{TS}}\nonumber\\ 
	 &= \frac{1}{{{m}^{2}}}\mathrm{V}\left[ \underset{i\in \mathcal{H}}{\mathop \sum }\,{{Z}_{i}^h}\nabla {{J}_{i}}( {{\theta }_{k}} )+\underset{j\in \mathcal{L}}{\mathop \sum }\,{{Z}_{j}^l}\nabla {{J}_{j}}( {{\theta }_{k}} ) \right] \nonumber \\
	 &=\frac{1}{{{m}^{2}}}\Bigg\{ \mathrm{Cov}\left[ \underset{i\in \mathcal{H}}{\mathop \sum }\,{{Z}_{i}^h}\nabla {{J}_{i}}( {{\theta }_{k}}),\underset{i\in \mathcal{H}}{\mathop \sum }\,{{Z}_{i}^h}\nabla {{J}_{i}}( {{\theta }_{k}}) \right]\nonumber\\
	 &~~~~~~~~~~~~~~~~+\mathrm{Cov}\left[ \underset{j\in \mathcal{L}}{\mathop \sum }\,{{Z}_{j}^l}\nabla {{J}_{j}}( {{\theta }_{k}}),\underset{j\in \mathcal{L}}{\mathop \sum }\,{{Z}_{j}^l}\nabla {{J}_{j}}( {{\theta }_{k}}) \right] \Bigg\} \nonumber \\
	 &=\frac{1}{{{m}^{2}}}\frac{{{n}_{1}}}{{{N}_{1}}\left( {{N}_{1}}-1 \right)}\left( 1-\frac{{{n}_{1}}}{{{N}_{1}}} \right)\nonumber\\
	 &~~~~~~~~~~~~~~~~~~*\left[ {{N}_{1}}\underset{i\in \mathcal{H}}{\mathop \sum }\,{\|\nabla {{J}_{i}}{{( {{\theta }_{k}})}\|}^{2}}-{\left\| \underset{i\in \mathcal{H}}{\mathop \sum }\,\nabla {{J}_{i}}( {{\theta }_{k}} ) \right\|^{2}} \right]~~~ \nonumber \\
	 &~~~+\frac{1}{{{m}^{2}}}\frac{{{n}_{2}}}{{{N}_{2}}\left( {{N}_{2}}-1 \right)}\left( 1-\frac{{{n}_{2}}}{{{N}_{2}}} \right)*\left[ {{N}_{2}}\underset{j\in \mathcal{L}}{\mathop \sum }\,{\|\nabla {{J}_{j}}{{\left( {{\theta }_{k}} \right)}\|}^{2}} \right] \label{eq10}
	 \end{align}
     where the first equation follows from the framework of the typicality sampling scheme, the second equation uses the definition of variance and the third equation applies the result of Lemma 3 in \citep{peng2019accelerating} and Definition \ref{eq5} of typical samples.
     
     Now we compare the gradient mean and its variance between conventional minibatch SGD and typical batch SGD, respectively. Subtract ${\expect \left[\nabla J_\batch (\theta_k)\right]}_{\textsc{SRS}}$ from ${\expect \left[\nabla J_\batch (\theta_k)\right]}_{\textsc{TS}}$ to get
     \begin{align}
     &{\left\| {\expect \left[\nabla J_\batch (\theta_k)\right]}_{\textsc{TS}} \right\|}-{\left\| {\expect \left[\nabla J_\batch (\theta_k)\right]}_{\textsc{SRS}} \right\|}\nonumber\\
     &={\left\| \frac{n_1 N}{N_1 m} \nabla J( {{\theta }_{k}}) \right\|}-{\left\| \nabla J( {{\theta }_{k}}) \right\|} \geq 0\label{eq11}
     \end{align}
     where the inequation is based on the restriction on sampling density that $n_1/N_1 \geq m/N$. We then subtract ${\mathrm{V}\left[\nabla J_\batch (\theta_k)\right]}_{\textsc{TS}}$ from ${\mathrm{V}\left[\nabla J_\batch (\theta_k)\right]}_{\textsc{SRS}}$ to get, after simplifying
     \begin{align}
     &{\mathrm{V}\left[\nabla J_\batch (\theta_k)\right]}_{\textsc{SRS}}-{\mathrm{V}\left[\nabla J_\batch (\theta_k)\right]}_{\textsc{TS}}\nonumber\\
     &=\frac{1}{m^2}\left[ \frac{m}{N}\left(1-\frac{m}{N}\right)-\frac{{{n}_{1}}}{ {{N}_{1}} }\left( 1-\frac{{{n}_{1}}}{{{N}_{1}}} \right) \right]\underset{i\in \mathcal{H}}{\mathop \sum }\,{\|\nabla {{J}_{i}}{( {{\theta }_{k}})}\|}^{2}\nonumber\\
     &~~~+\frac{1}{m^2}\left[ \frac{m}{N}\left(1-\frac{m}{N}\right)-\frac{{{n}_{2}}}{ {{N}_{2}}}\left( 1-\frac{{{n}_{2}}}{{{N}_{2}}} \right) \right]\underset{j\in \mathcal{L}}{\mathop \sum }\,{\|\nabla {{J}_{j}}{( {{\theta }_{k}})}\|}^{2}~~~\nonumber\\
     &~~~+\frac{N^2}{m^2}\left[  \frac{{{n}_{1}}}{{{N}_{1}}^2}\left( 1-\frac{{{n}_{1}}}{{{N}_{1}}} \right) - \frac{m}{N^2}\left(1-\frac{m}{N}\right) \right]{\left\| \nabla J(\theta_{k})\right\|}^2 \label{eq12}
     \end{align}
      Following from Assumption \ref{assumption1} and the assumption made in \citep{peng2019accelerating} that for some constant $\beta_1 \geq 0$, $\beta_2 \geq 1$
     \begin{equation}
     {{\|\nabla {{J}_{i}}(\theta)\|}^{2}}\le {{\beta}_{1}}+{{\beta}_{2}}{{\|\nabla J(\theta)\|}^{2}} \,\,\,\,\,\,\,\,\, i=1,\ldots,N. \label{eq13}
     \end{equation}
      we have
      \begin{align}
      	&{\mathrm{V}\left[\nabla J_\batch (\theta_k)\right]}_{\textsc{SRS}}-{\mathrm{V}\left[\nabla J_\batch (\theta_k)\right]}_{\textsc{TS}}\nonumber\\
      	&\geq \frac{1}{m^2}\Bigg[ m\left(1-\frac{m}{N}\right)-n_1\left( 1-\frac{{{n}_{1}}}{{{N}_{1}}} \right)-n_2\left( 1-\frac{{{n}_{2}}}{{{N}_{2}}} \right)\Bigg]\nonumber\\
      	&~~~~~~~~~~~~~~~~~~~~~~~~~~~~~~~~~~~~~~~~*\left[{{\beta}_{1}}+{\beta}_{2}{\|\nabla J(\theta)\|}^{2}\right]~~~~~\nonumber\\
      	&~~~+\frac{N^2}{m^2}\left[  \frac{{{n}_{1}}}{{{N}_{1}}^2}*\left( 1-\frac{{{n}_{1}}}{{{N}_{1}}} \right) - \frac{m}{N^2}\left(1-\frac{m}{N}\right) \right]{\left\| \nabla J(\theta_{k})\right\|}^2 \label{eq14}
      \end{align}
       Rearrange the above inequality can obtain
       \begin{align}
       &{\mathrm{V}\left[\nabla J_\batch (\theta_k)\right]}_{\textsc{SRS}}-{\mathrm{V}\left[\nabla J_\batch (\theta_k)\right]}_{\textsc{TS}}\nonumber\\
       &\geq \frac{1}{m^2 N}\left[\left(\sqrt{\frac{N_2}{N_1}n_1}-\sqrt{\frac{N_1}{N_2}n_2}\right)^{2}\right]*\left[{{\beta}_{1}}+{\beta}_{2}{\|\nabla J(\theta)\|}^{2}\right]~~\nonumber\\
       &~~~+\frac{N^2}{m^2}\left[  \frac{{{n}_{1}}}{{{N}_{1}}^2}\left( 1-\frac{{{n}_{1}}}{{{N}_{1}}} \right) - \frac{m}{N^2}\left(1-\frac{m}{N}\right) \right]{\left\| \nabla J(\theta_{k})\right\|}^2 \nonumber\\
       & \geq 0 \label{eq15}
       \end{align}
       where the first inequation applies triangle inequality and the second inequation is based on the fact that $n_1/N_1 \geq m/N$. The equality \eqref{eq15} implies that the standard deviation satisfies
       \begin{equation}
       {\mathrm{STD} \left[\nabla J_\batch (\theta_k)\right]}_{\textsc{SRS}} \geq {\mathrm{STD} \left[\nabla J_\batch (\theta_k)\right]}_{\textsc{TS}} \label{eq16}
       \end{equation}
       According to the definition \eqref{def:gsnr} of gradient SNR, the equality \eqref{eq11} and \eqref{eq16} yield the conclusion of this theorem immediately.                  
\end{IEEEproof}

\begin{figure*}[!t]
	\centering
	\includegraphics[width=\textwidth, scale=1]{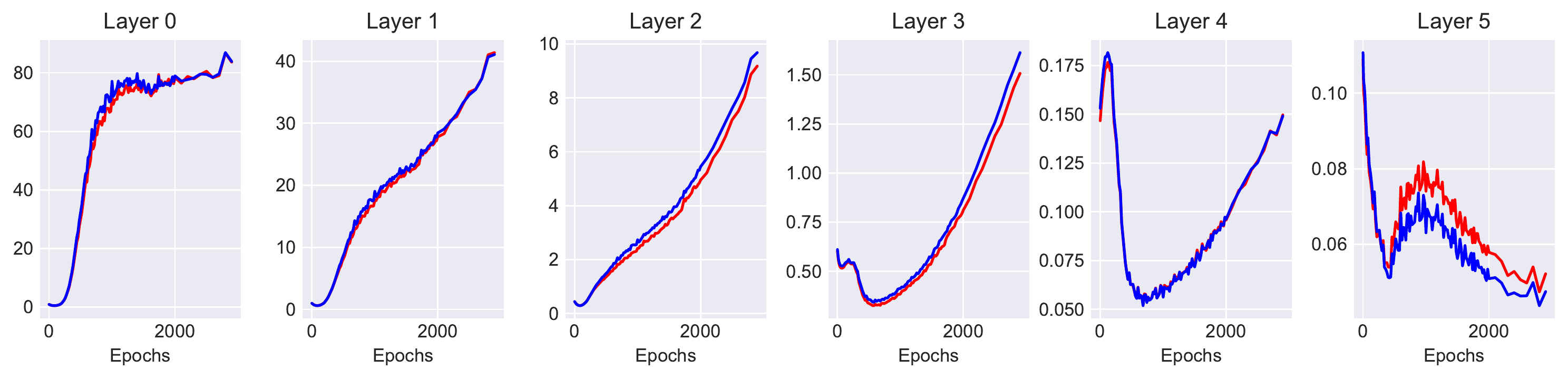}
	\caption{The averaged squared L2 norm of gradient across the samples in each subset $\mathcal{H}$ and $\mathcal{L}$. The red curves represent the results on subset  $\mathcal{H}$ and the blue curves represent the results on subset $\mathcal{L}$.}
	\label{fig:1}
\end{figure*}

Combined with the observation in Section \ref{subsection21} that high gradient SNR greatly increases the mutual information $I(T;Y)$, typical batch SGD is supposed to fit the training label with fewer epochs. In other words, the more informative gradient features provided by typical samples help to generate a sketch for the optimization landscape of the network in less time. Thus, we believe typicality sampling accelerates minibatch SGD by speeding up the fitting phase of the training process.

Note that there are some debates regarding the generality of IB theory. The authors in \citep{saxe2019information} suggested that the single-sided saturating activation functions may influence the two-phase phenomenon in the IB theory, and the compression phase may not exist explicitly. We are convinced that these doubts will not affect our findings. The reason is that instead of the compression phase, our theoretical analysis mainly focuses on the fitting phase which appears in all cases. Moreover, the existence of the transition from high gradient SNR phase to low gradient SNR phase in SGD optimization has been verified before \citep{murata1998statistical,chee2017convergence} and it will not be affected by the disappearance of the compression phase. In Section \ref{section4}, we will experimentally validate if our theoretical analysis holds in different cases.

Note also that we can treat the typicality sampling scheme as a data compression method that demarcates the most important samples from the training set. A similar work proposed before is the deep variational information bottleneck (VIB), which uses variational inference to approximate the IB trade-off \citep{alemi2016deep}. The difference is that the typicality sampling scheme concentrates on compressing to capture as much information as possible about the label before the optimization begins, while the deep VIB simplifies the internal representation to reduce model complexity during the training. Although typical batch SGD might not be able to reach the optimal bound given by the deep VIB, it saves a lot of training time and converges to the region close to the optimal point.

Moreover, our theoretical analysis also proves that one can leverage the prior information of the training set to speed up the information extraction process and approach the lower bound of IB theory faster. Although the typicality sampling scheme may not be the best way to use this prior information, it is a simple and straightforward method with a negligible additional computational cost to find a good compressed representation for each layer of the DNN.

\section{Experiments}
\label{section4}

\begin{table*}
	\centering
	\begin{tabular}{|l|r|r|r|r|r|}
		\hline
		name & \# class& \# features & \# training size & \# validation size &\# architecture \\
		\hline
		Pendigits&   10&           16&             7,494&               3,498&      16-256-56-10 \\
		Usps&        10&          256&             7,291&               2,007& 256-1024-200-20-20-10\\
		MNIST&       10&          784&            55,000&              10,000& 784-1024-200-20-20-10\\
		\hline
	\end{tabular}
	\caption{Summary of the real-world datasets and neural network architectures}
	\label{tab:1}
\end{table*}

\begin{figure*}[!t]
	\centering
	\begin{subfigure}[t]{0.24\textwidth}
		\includegraphics[width=\textwidth]{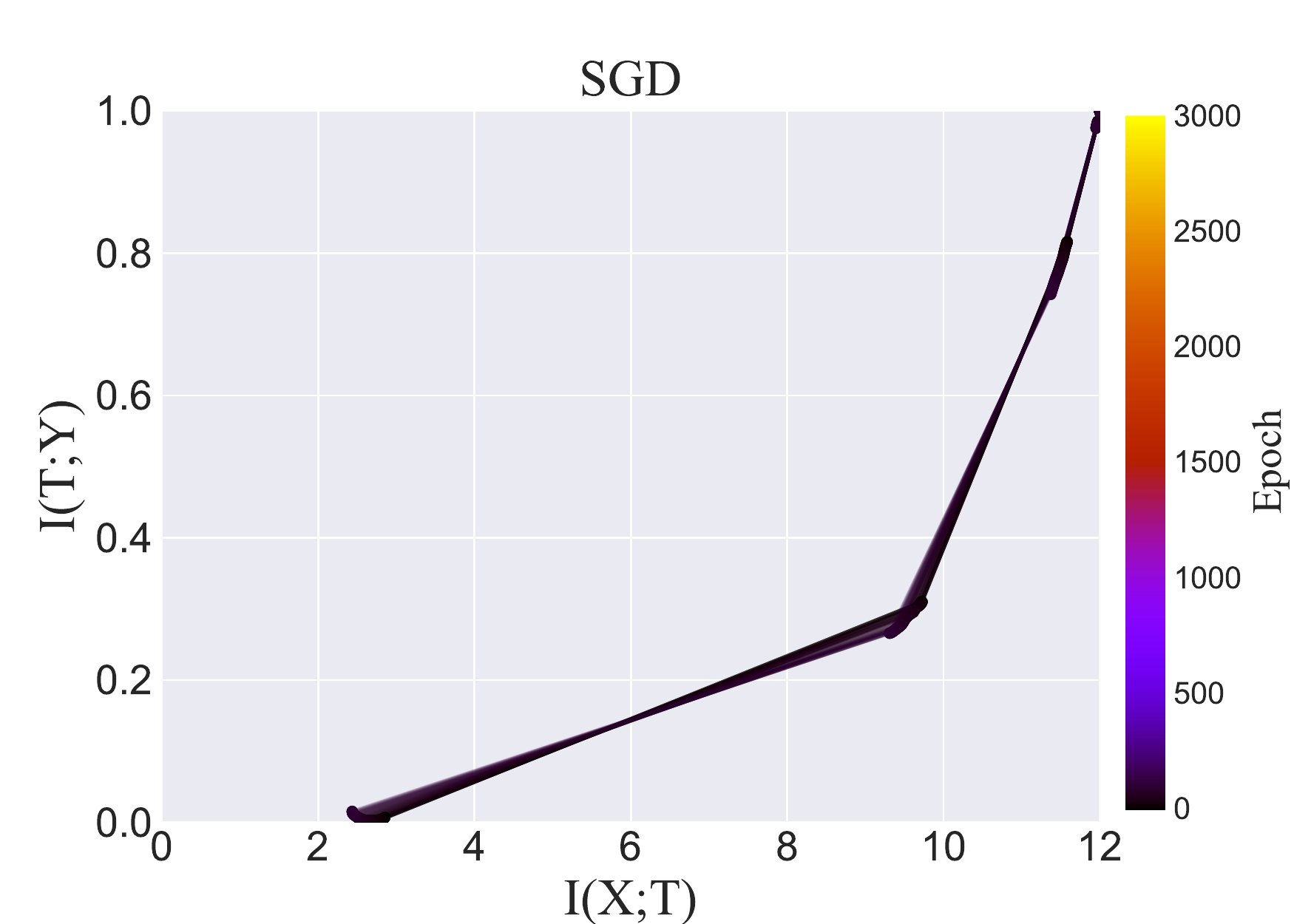}
	\end{subfigure}
	\hfill
	\begin{subfigure}[t]{0.24\textwidth}
		\includegraphics[width=\textwidth]{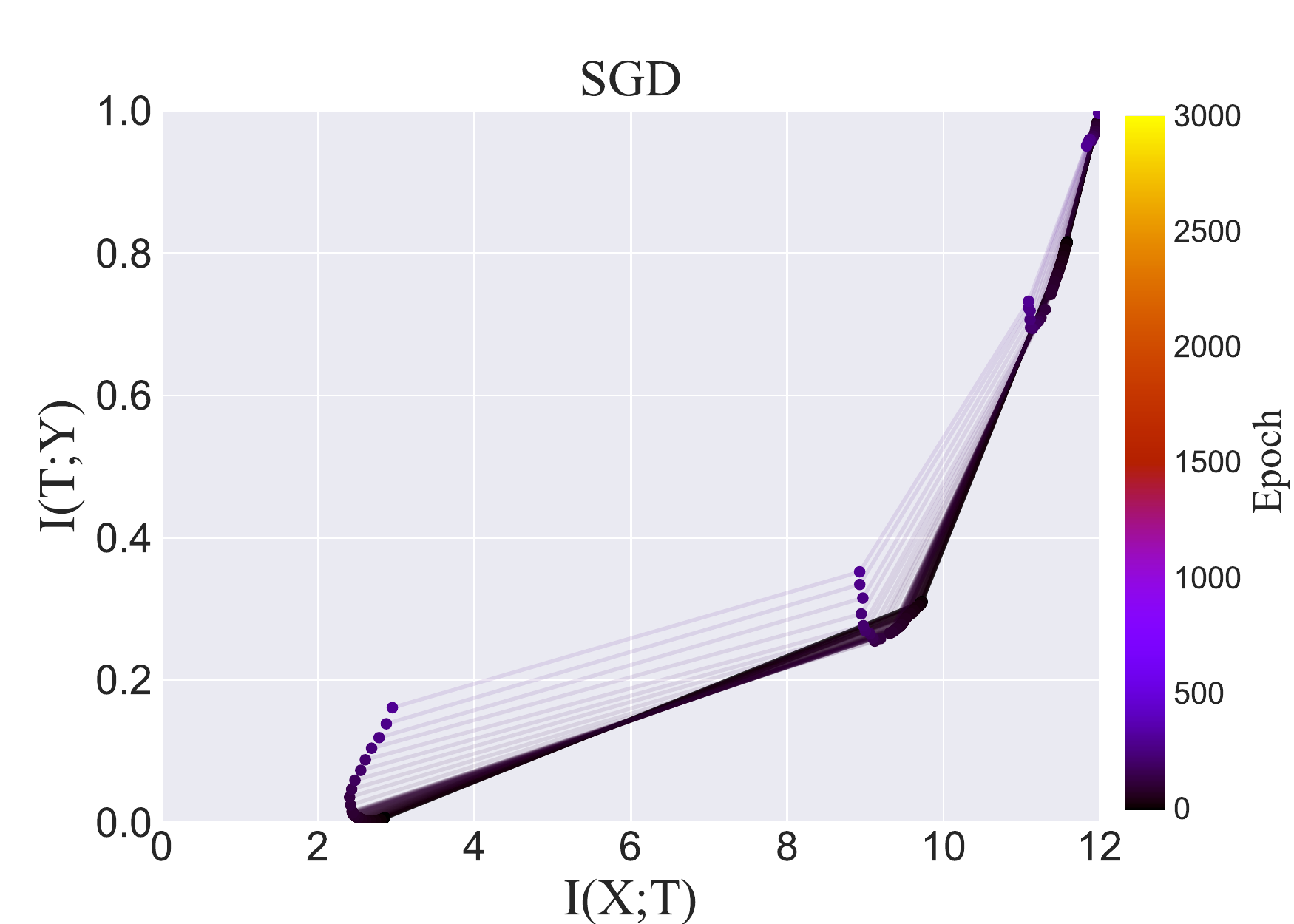}
	\end{subfigure}
	\hfill
	\begin{subfigure}[t]{0.24\textwidth}
		\includegraphics[width=\textwidth]{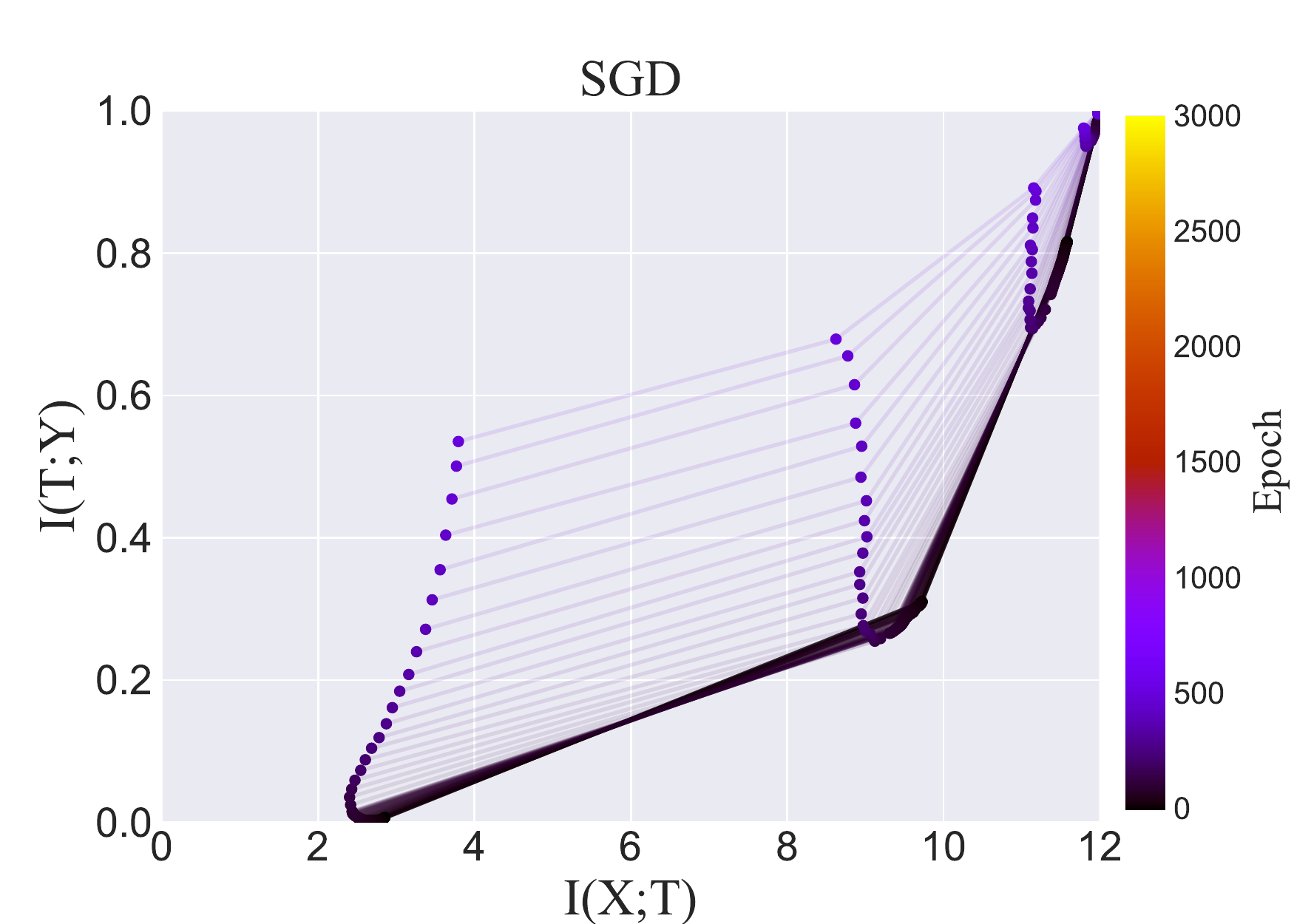}
	\end{subfigure} 
	\hfill
	\begin{subfigure}[t]{0.24\textwidth}
		\includegraphics[width=\textwidth]{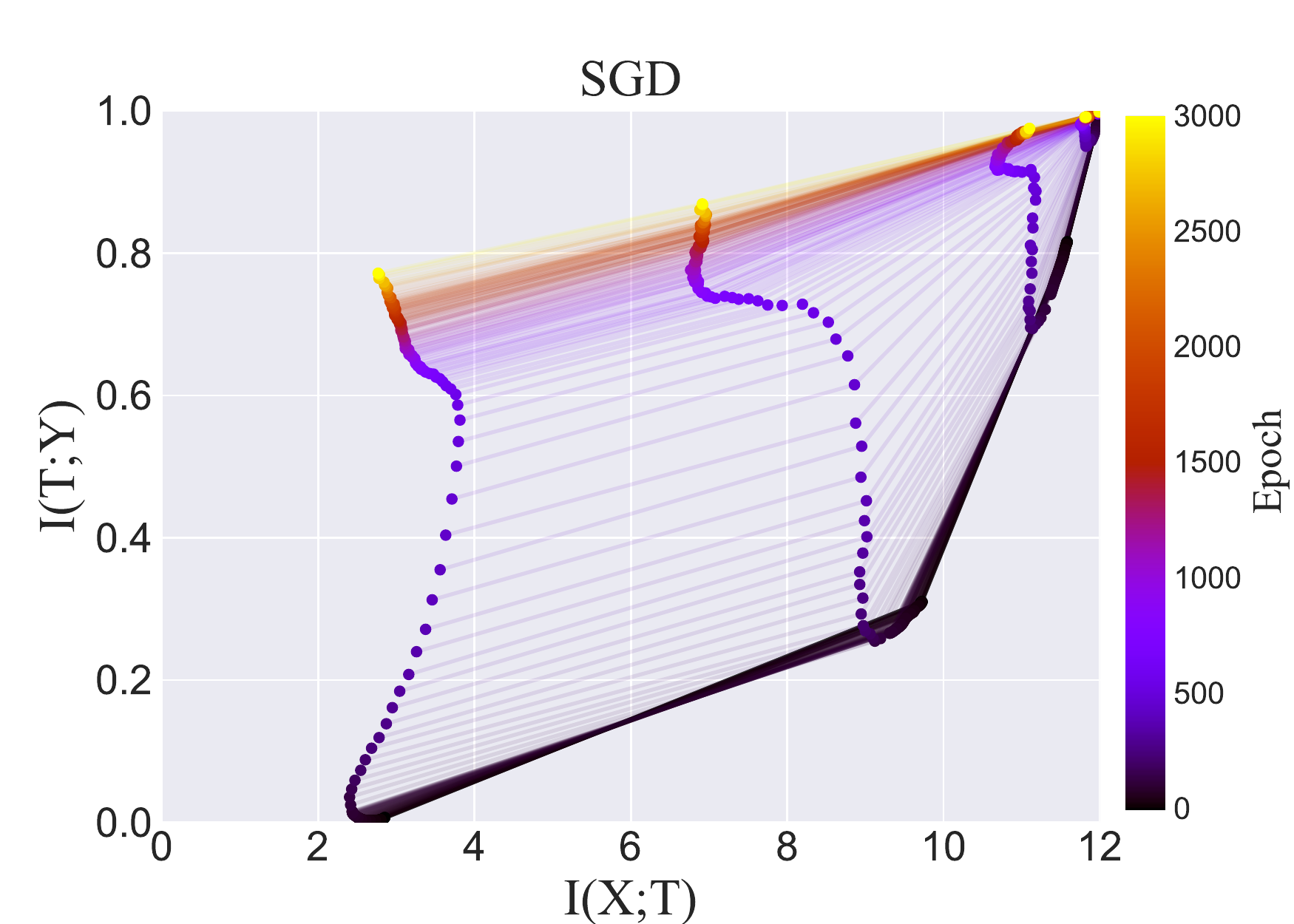}
	\end{subfigure} 
	\hfill
	\begin{subfigure}[t]{0.24\textwidth}
		\includegraphics[width=\textwidth]{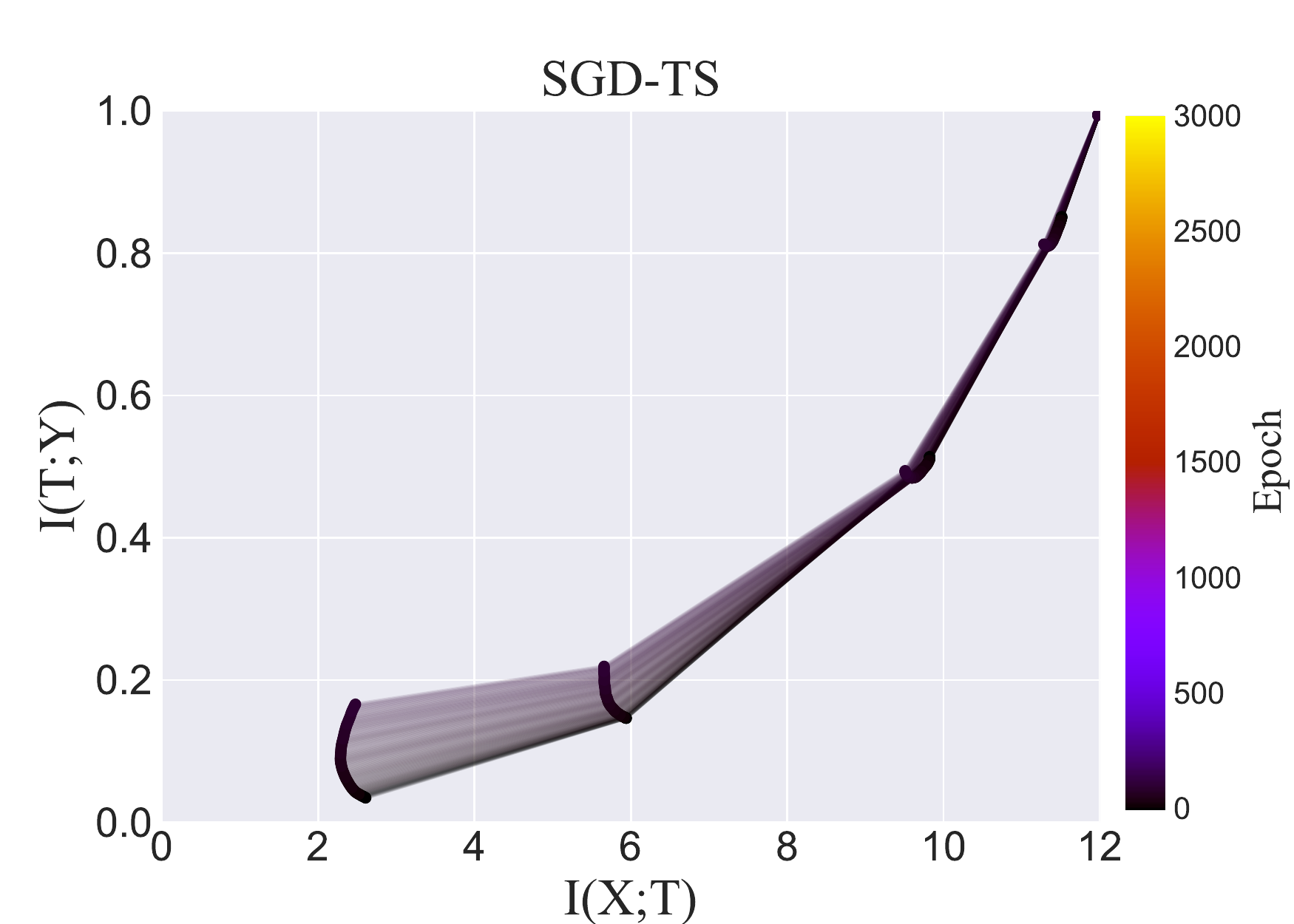}
	\end{subfigure}
	\hfill
	\begin{subfigure}[t]{0.24\textwidth}
		\includegraphics[width=\textwidth]{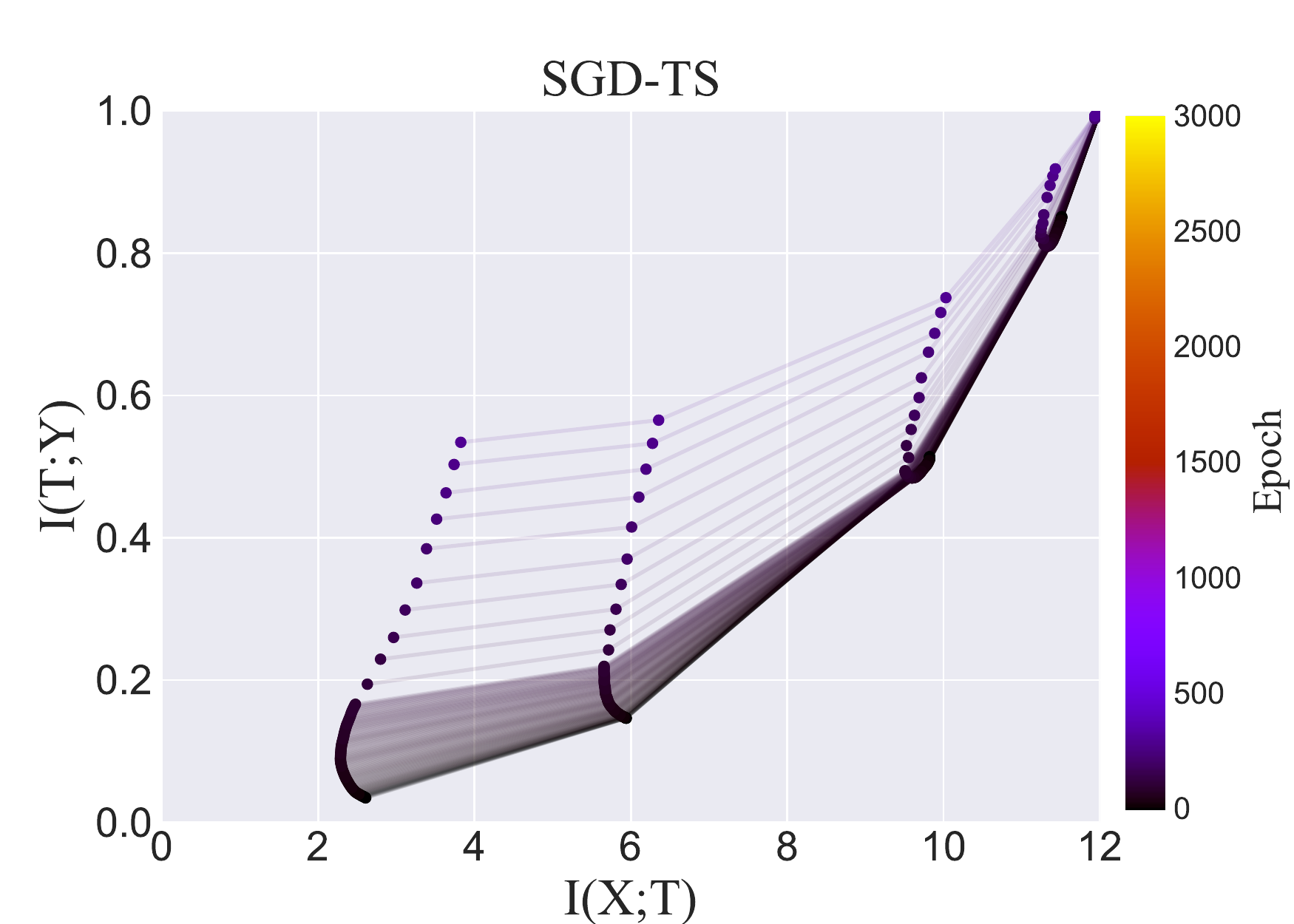}
	\end{subfigure}
	\hfill
	\begin{subfigure}[t]{0.24\textwidth}
		\includegraphics[width=\textwidth]{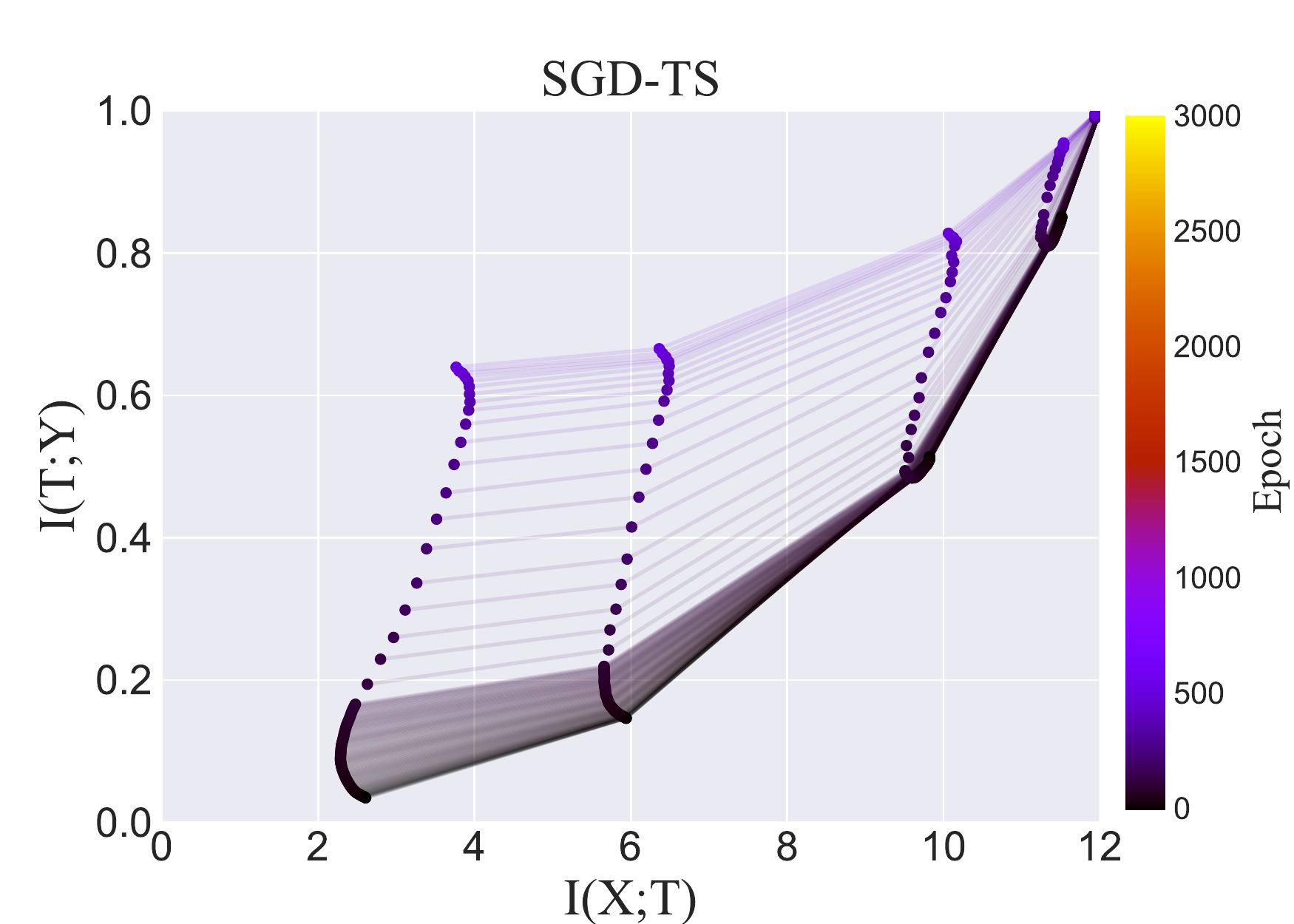}
	\end{subfigure} 
	\hfill
	\begin{subfigure}[t]{0.24\textwidth}
		\includegraphics[width=\textwidth]{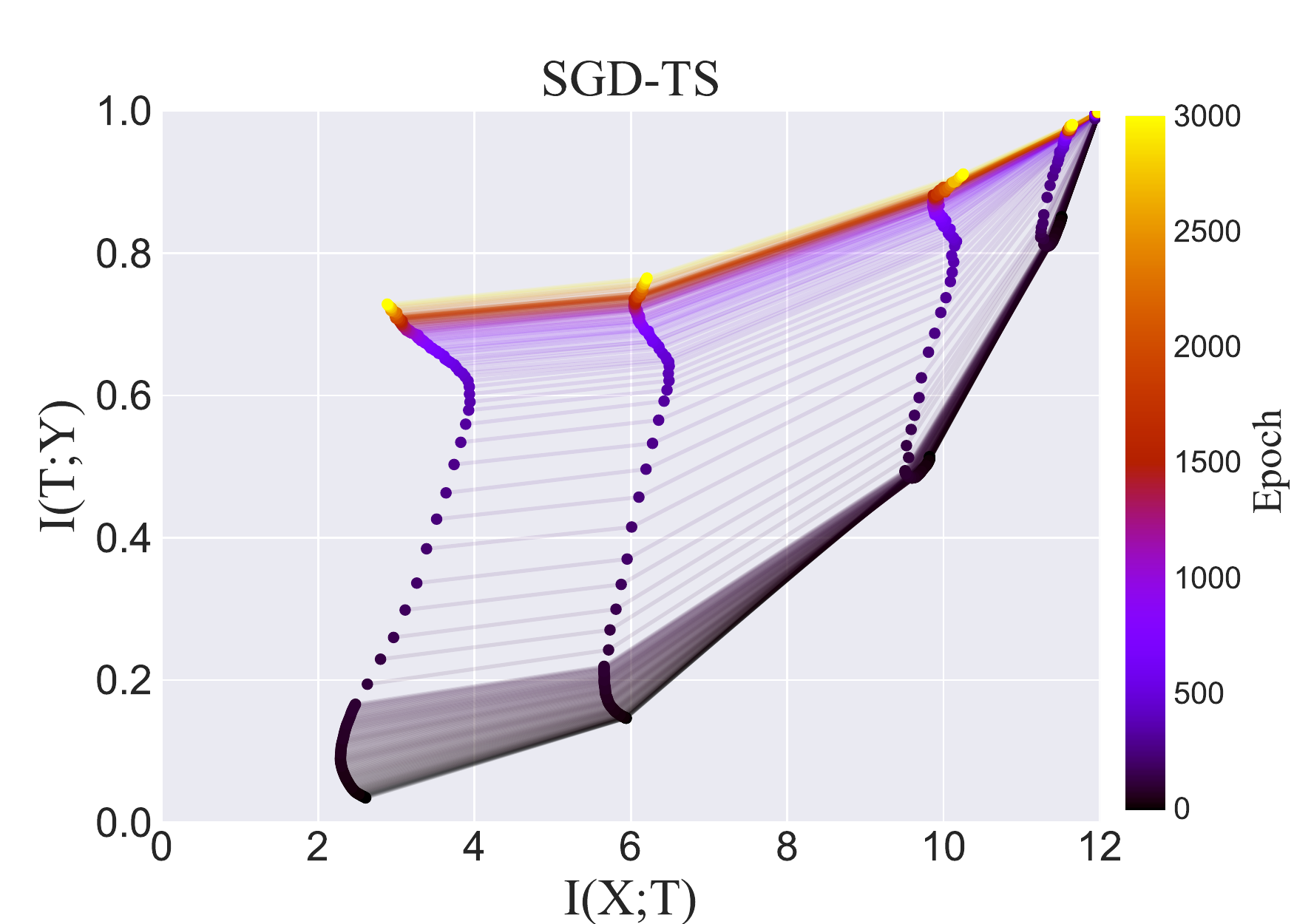}
	\end{subfigure}
	\begin{subfigure}[t]{\textwidth}
		\centering
		\includegraphics[width=\textwidth]{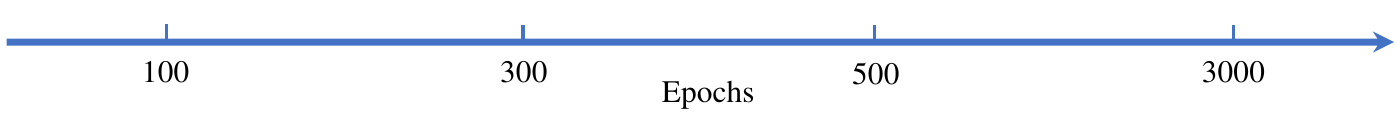}
	\end{subfigure}  
	\caption{The evolution of the layers information paths with different training epochs for synthetic dataset, during the optimization of conventional minibatch SGD and typical batch SGD. Same training steps of different layers are connected by the gray thin lines. Left to right: at 100, 300, 500 and 3000 epochs. The first row shows the optimization process of conventional minibatch SGD, the second row shows the optimization process of typical batch SGD.}
	\label{fig:2}
\end{figure*}

\begin{figure*}[!t]
	\centering
	\begin{subfigure}[t]{0.24\textwidth}
		\includegraphics[width=\textwidth]{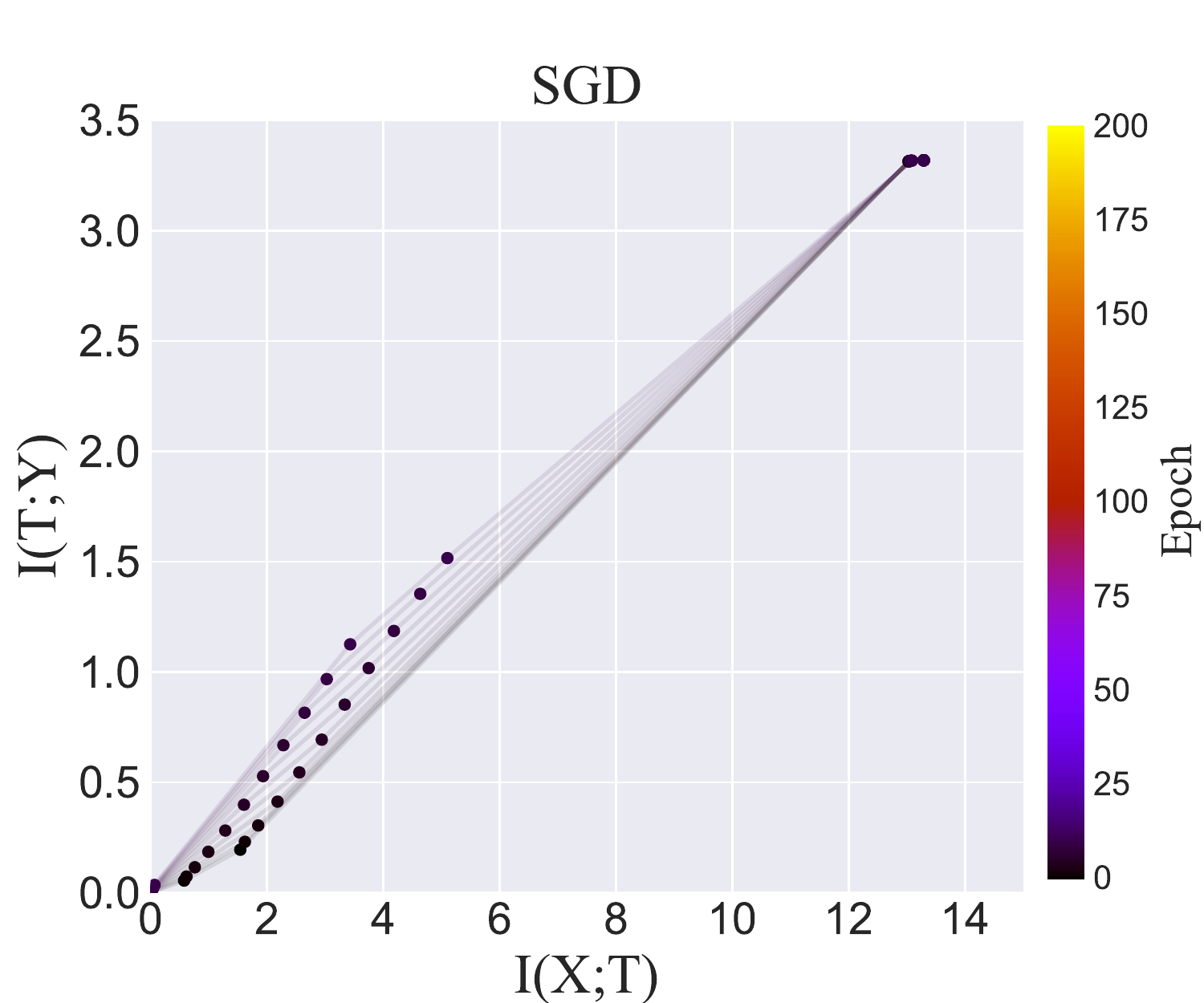}
	\end{subfigure}
	\hfill
	\begin{subfigure}[t]{0.24\textwidth}
		\includegraphics[width=\textwidth]{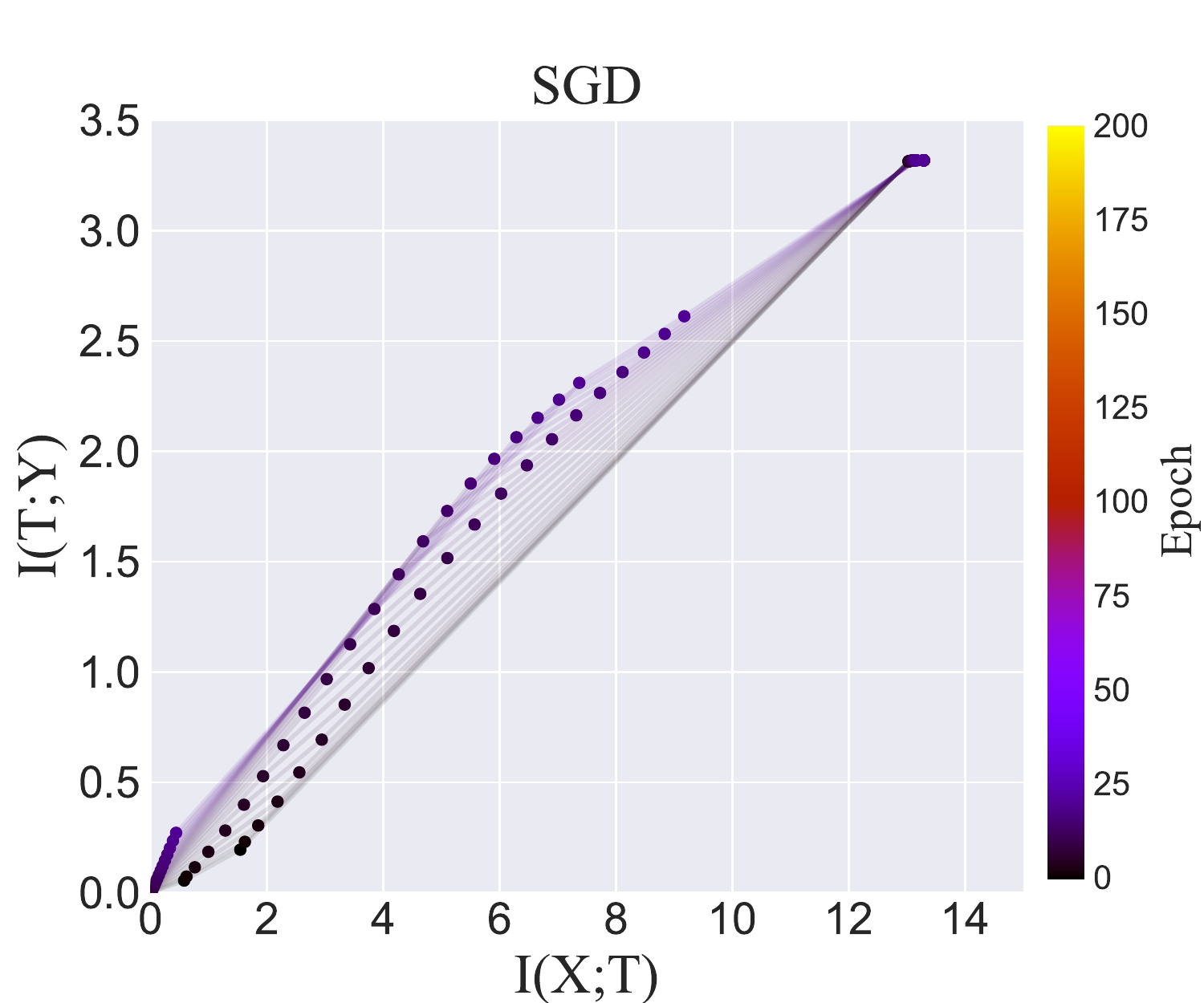}
	\end{subfigure}
	\hfill
	\begin{subfigure}[t]{0.24\textwidth}
		\includegraphics[width=\textwidth]{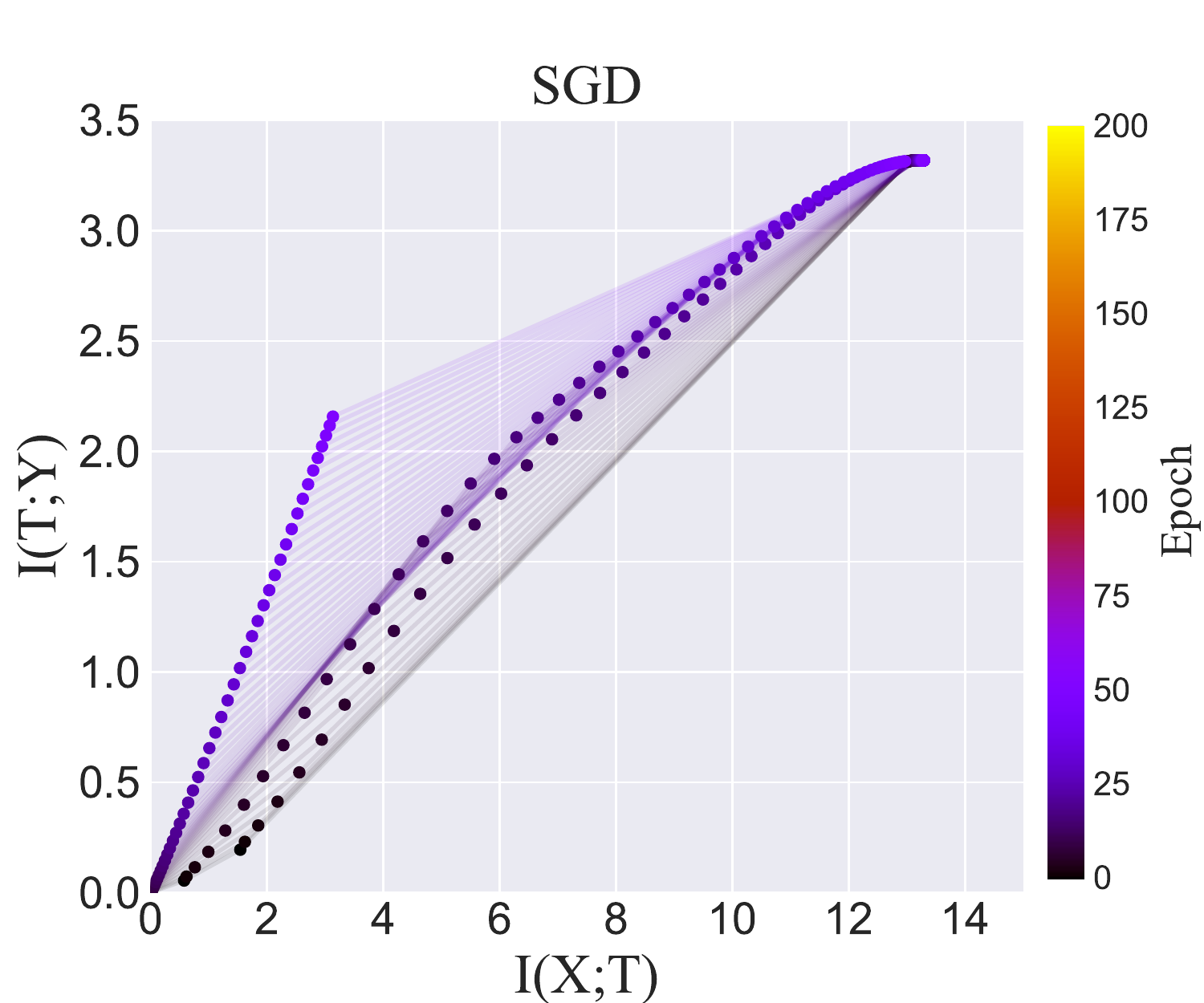}
	\end{subfigure} 
	\hfill
	\begin{subfigure}[t]{0.24\textwidth}
		\includegraphics[width=\textwidth]{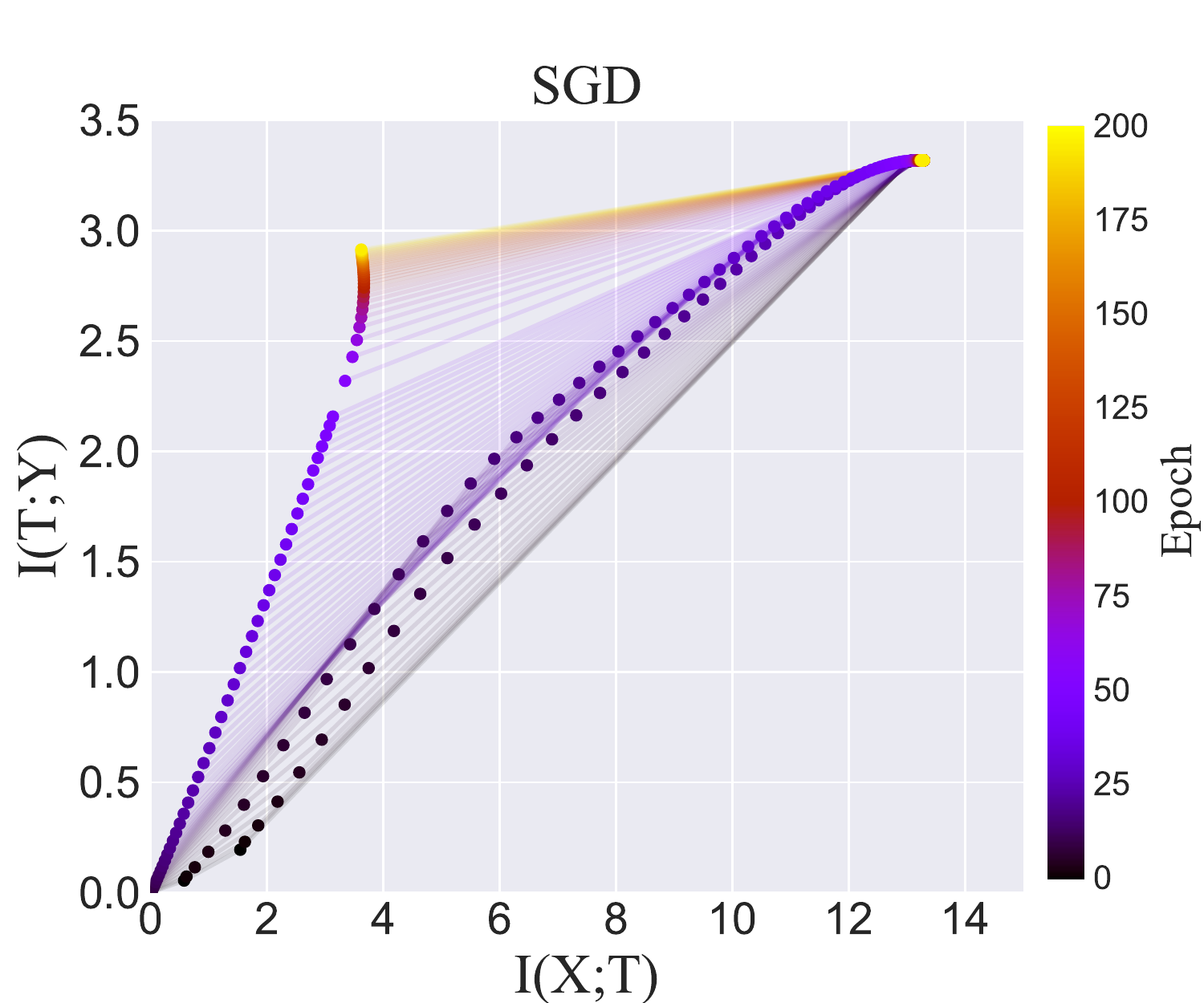}
	\end{subfigure} 
	\hfill
	\begin{subfigure}[t]{0.24\textwidth}
		\includegraphics[width=\textwidth]{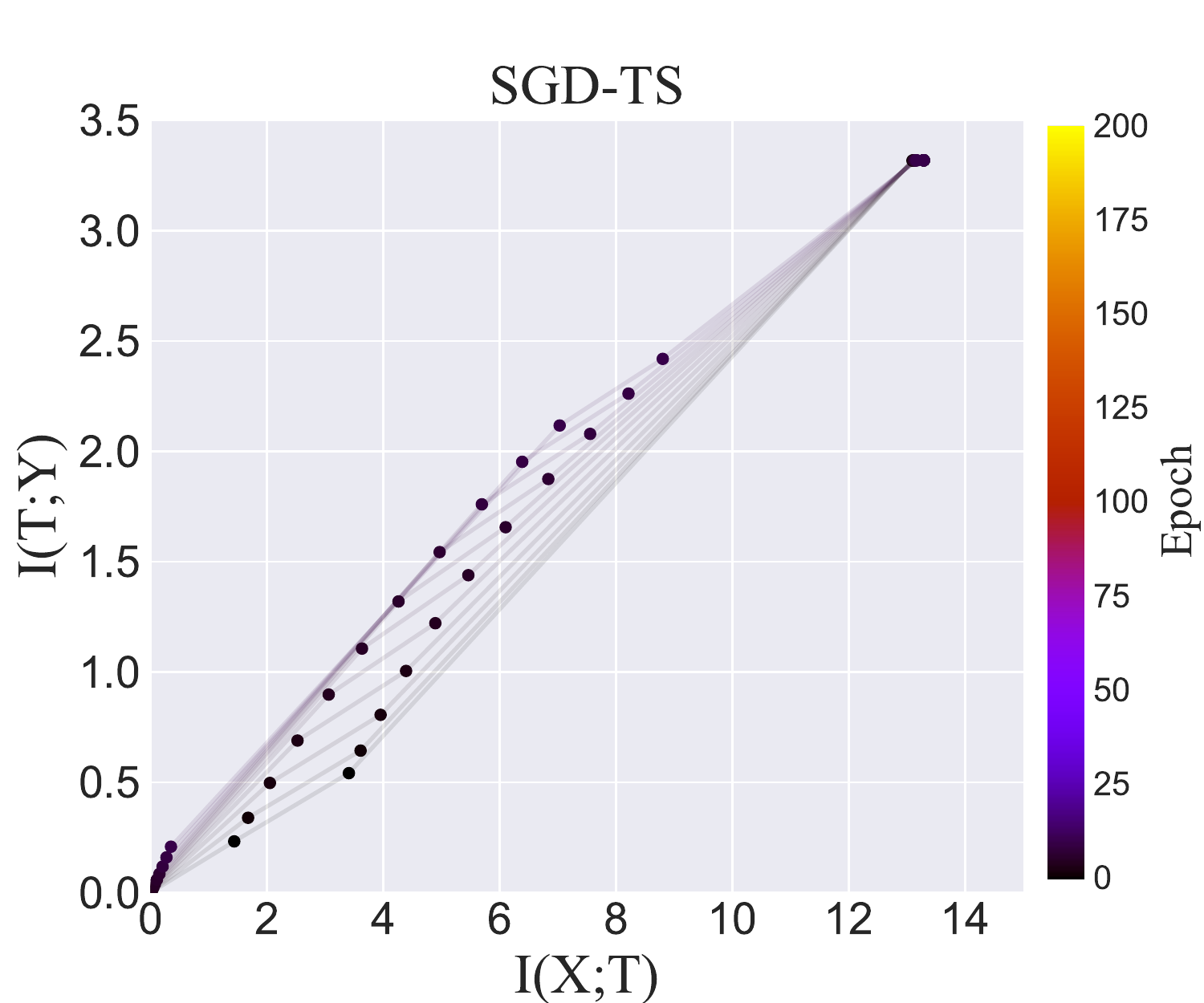}
	\end{subfigure}
	\hfill
	\begin{subfigure}[t]{0.24\textwidth}
		\includegraphics[width=\textwidth]{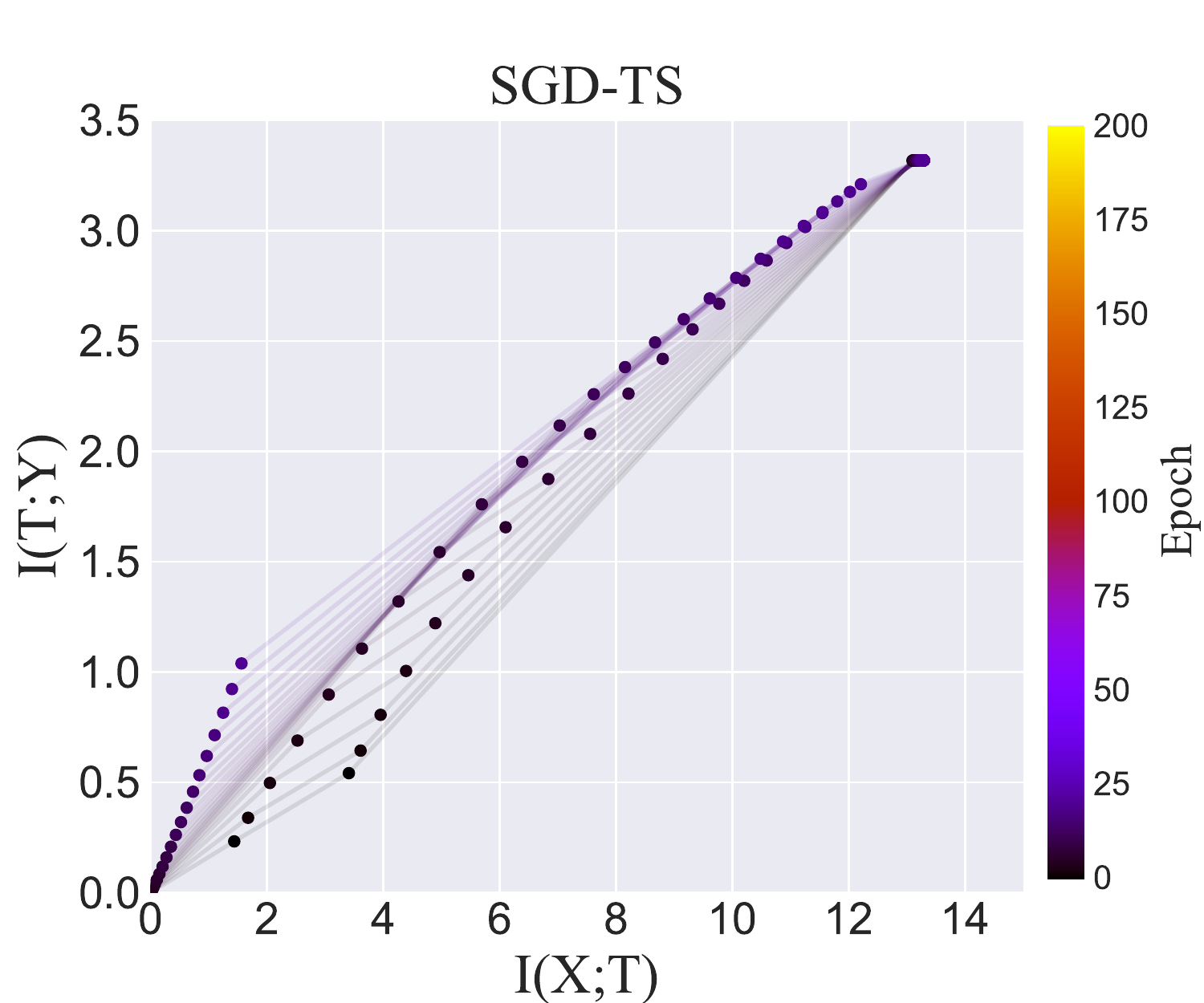}
	\end{subfigure}
	\hfill
	\begin{subfigure}[t]{0.24\textwidth}
		\includegraphics[width=\textwidth]{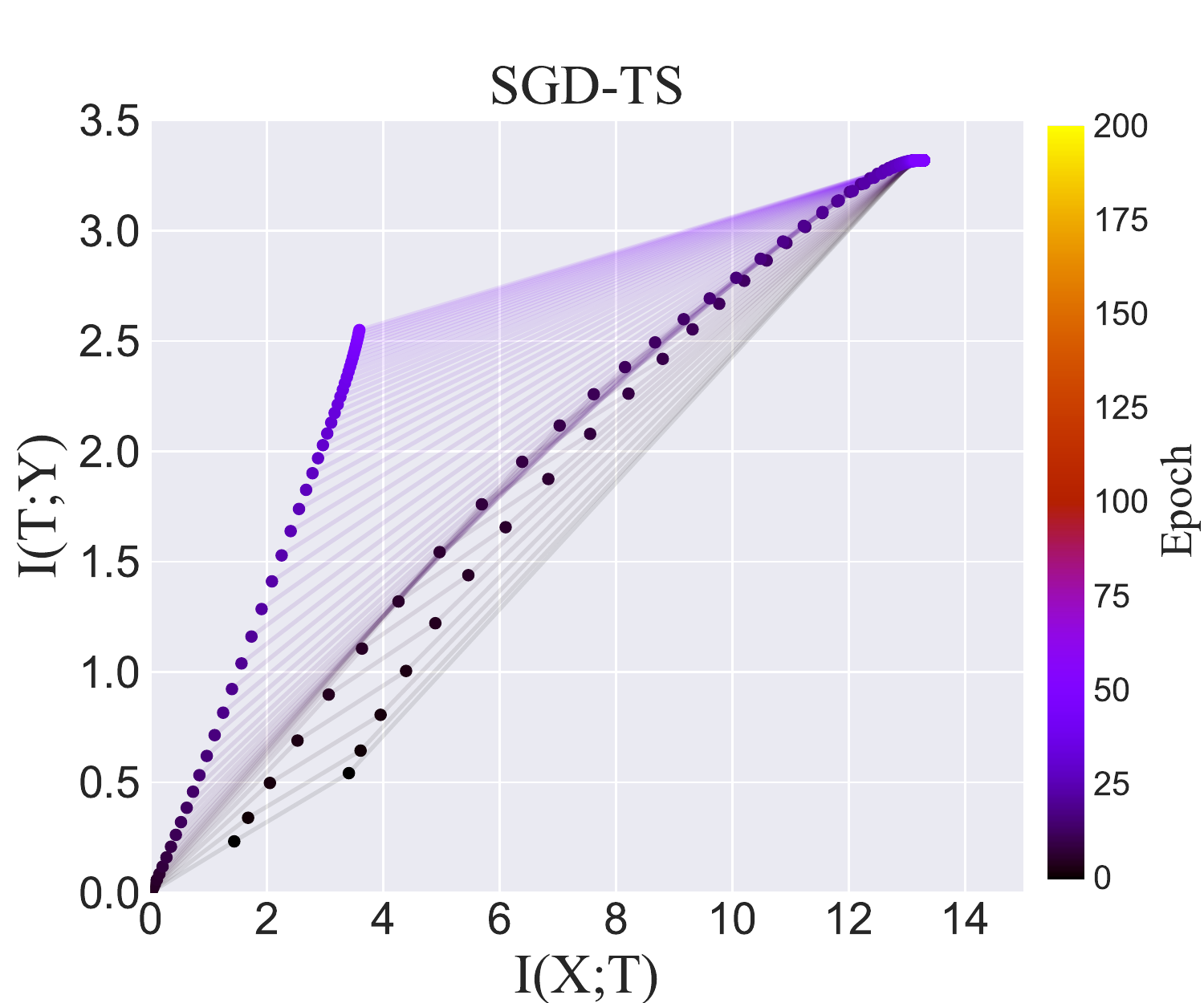}
	\end{subfigure} 
	\hfill
	\begin{subfigure}[t]{0.24\textwidth}
		\includegraphics[width=\textwidth]{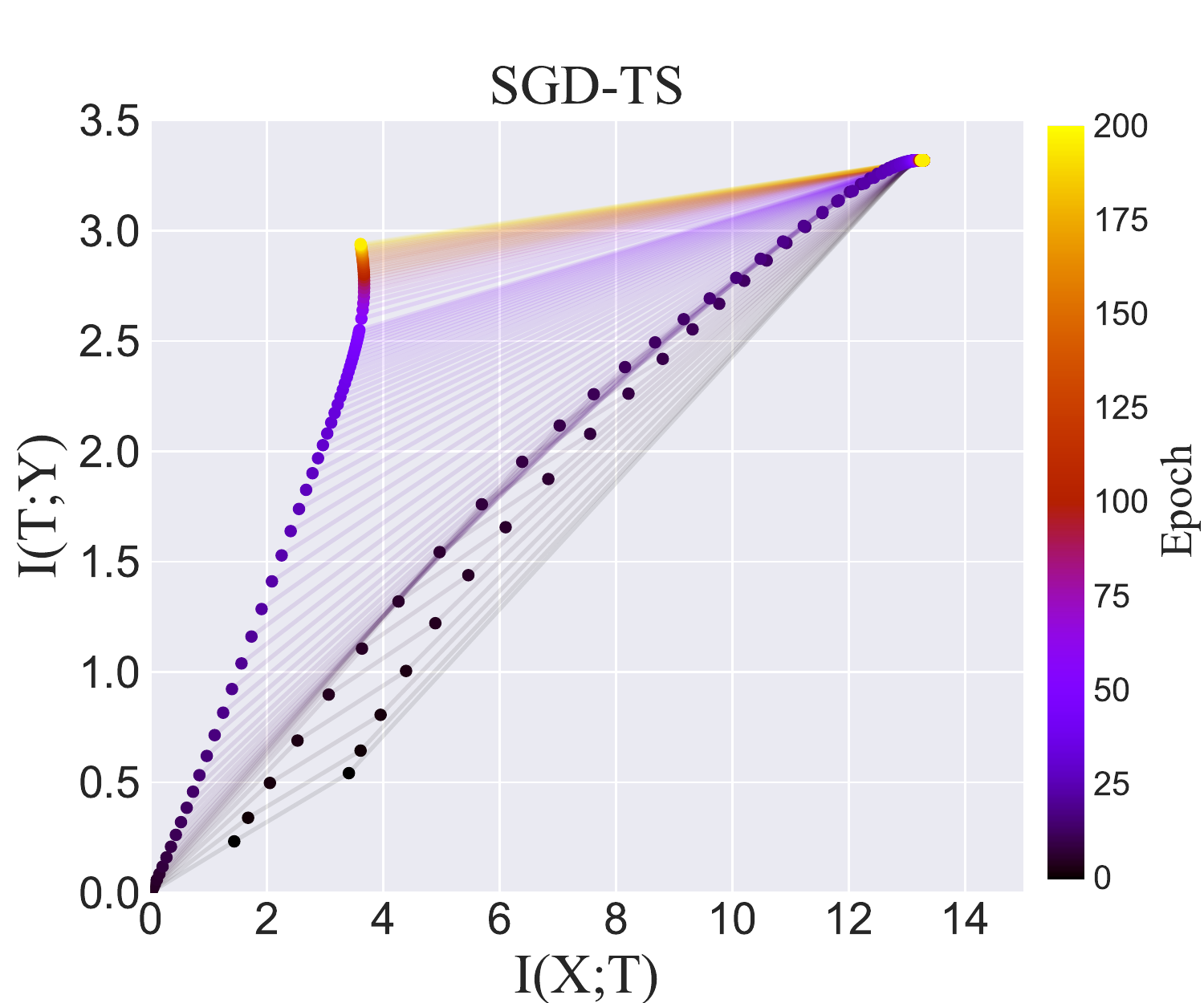}
	\end{subfigure}
	\begin{subfigure}[t]{\textwidth}
		\centering
		~\includegraphics[width=\textwidth]{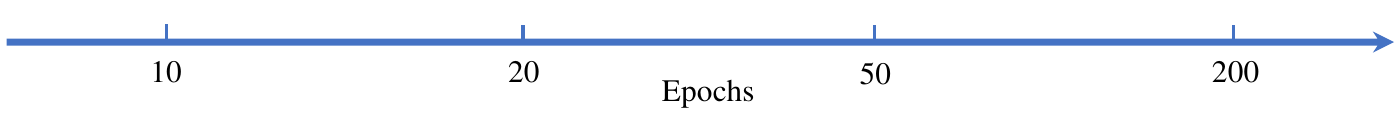}
	\end{subfigure} 
	\caption{The evolution of the layers information paths with different training epochs for MNIST dataset, during the optimization of conventional minibatch SGD and typical batch SGD. Same training steps of different layers are connected by the gray thin lines. Left to right: at 10, 20, 50 and 200 epochs. The first row shows the optimization process of conventional minibatch SGD, the second row shows the optimization process of typical batch SGD.}
	\label{fig:3}
\end{figure*}

\begin{figure*}[!t]
	\centering
	\begin{subfigure}[t]{0.24\textwidth}
		\includegraphics[width=\textwidth]{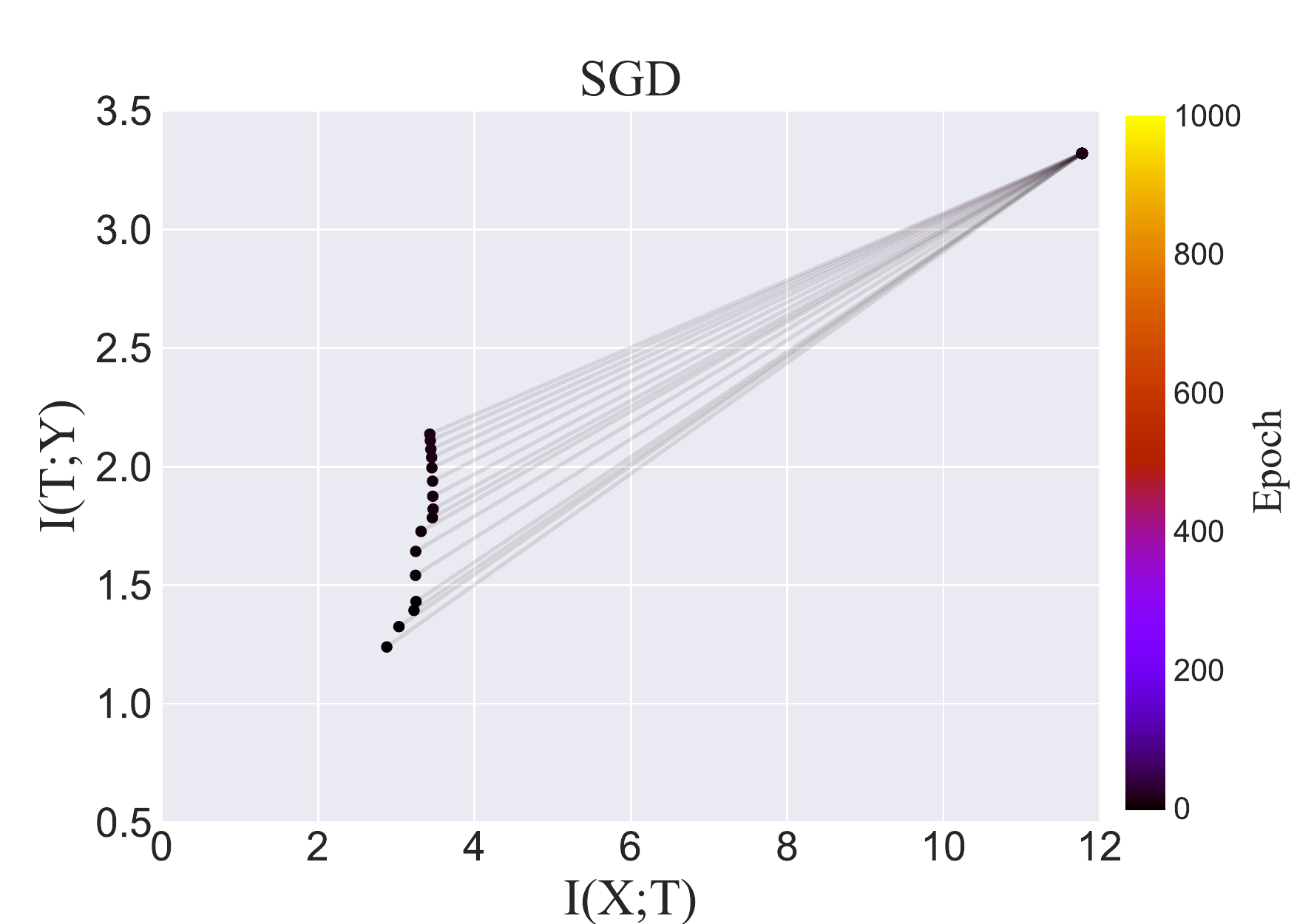}
	\end{subfigure}
	\hfill
	\begin{subfigure}[t]{0.24\textwidth}
		\includegraphics[width=\textwidth]{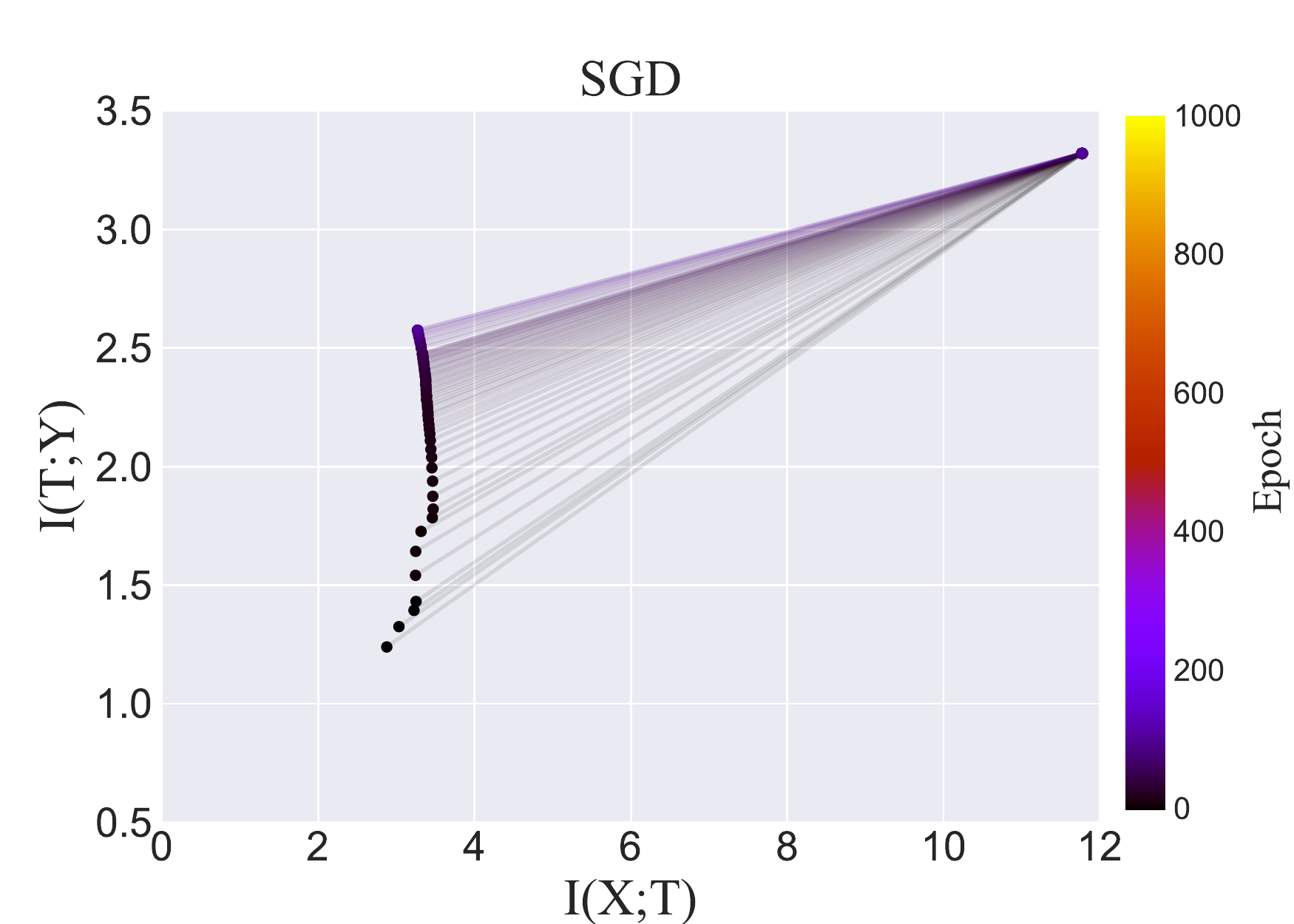}
	\end{subfigure}
	\hfill
	\begin{subfigure}[t]{0.24\textwidth}
		\includegraphics[width=\textwidth]{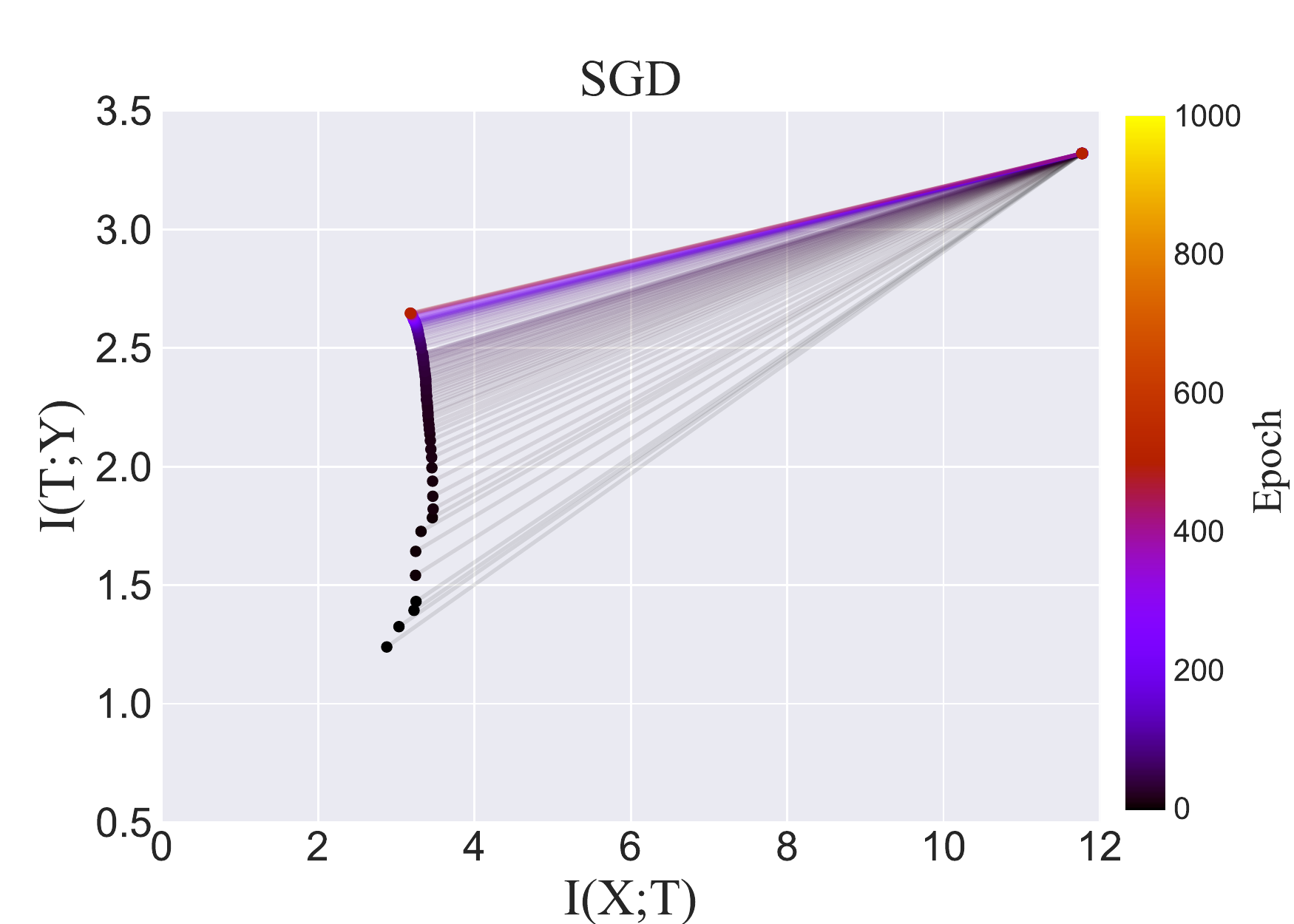}
	\end{subfigure} 
	\hfill
	\begin{subfigure}[t]{0.24\textwidth}
		\includegraphics[width=\textwidth]{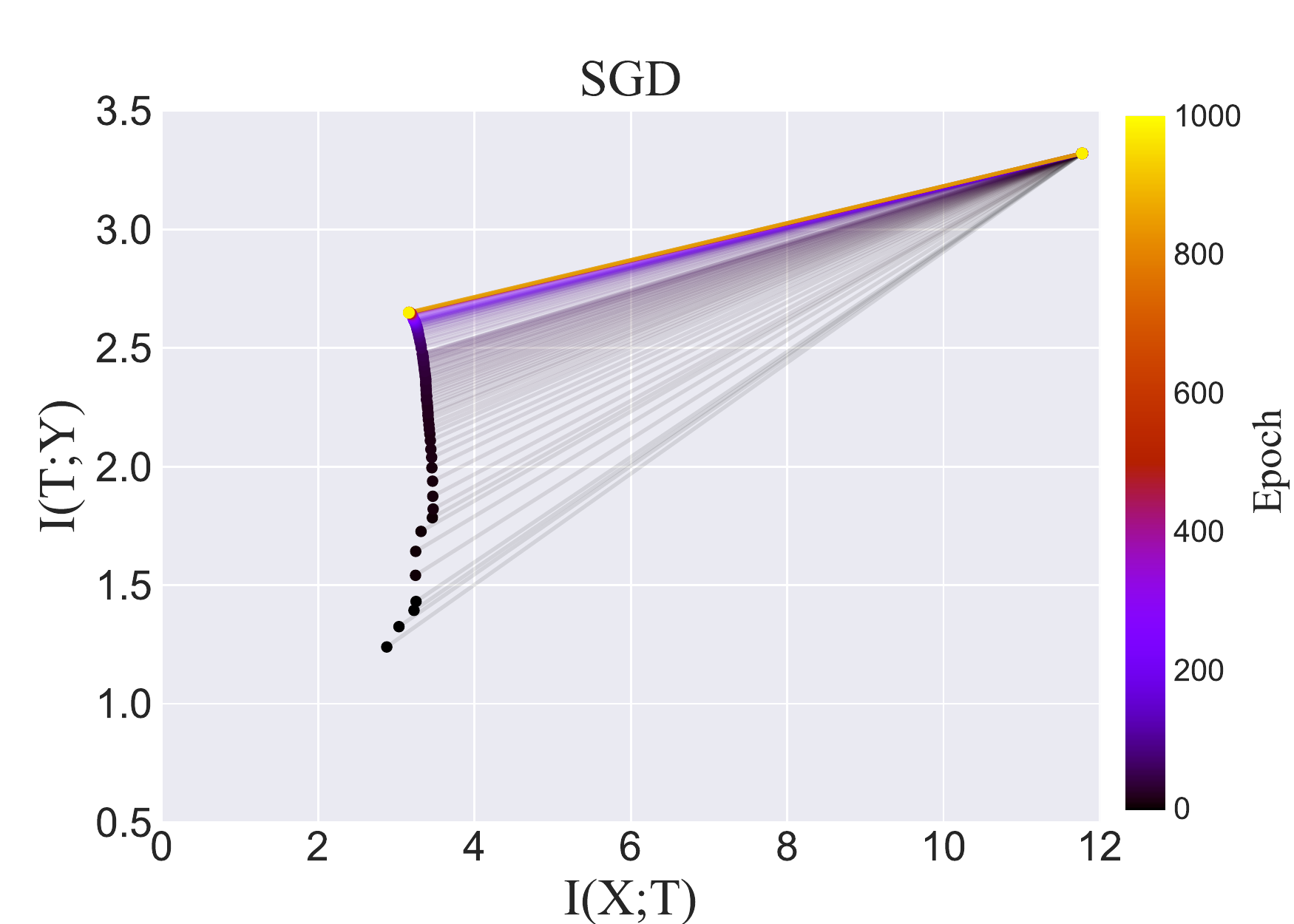}
	\end{subfigure} 
	\hfill
	\begin{subfigure}[t]{0.24\textwidth}
		\includegraphics[width=\textwidth]{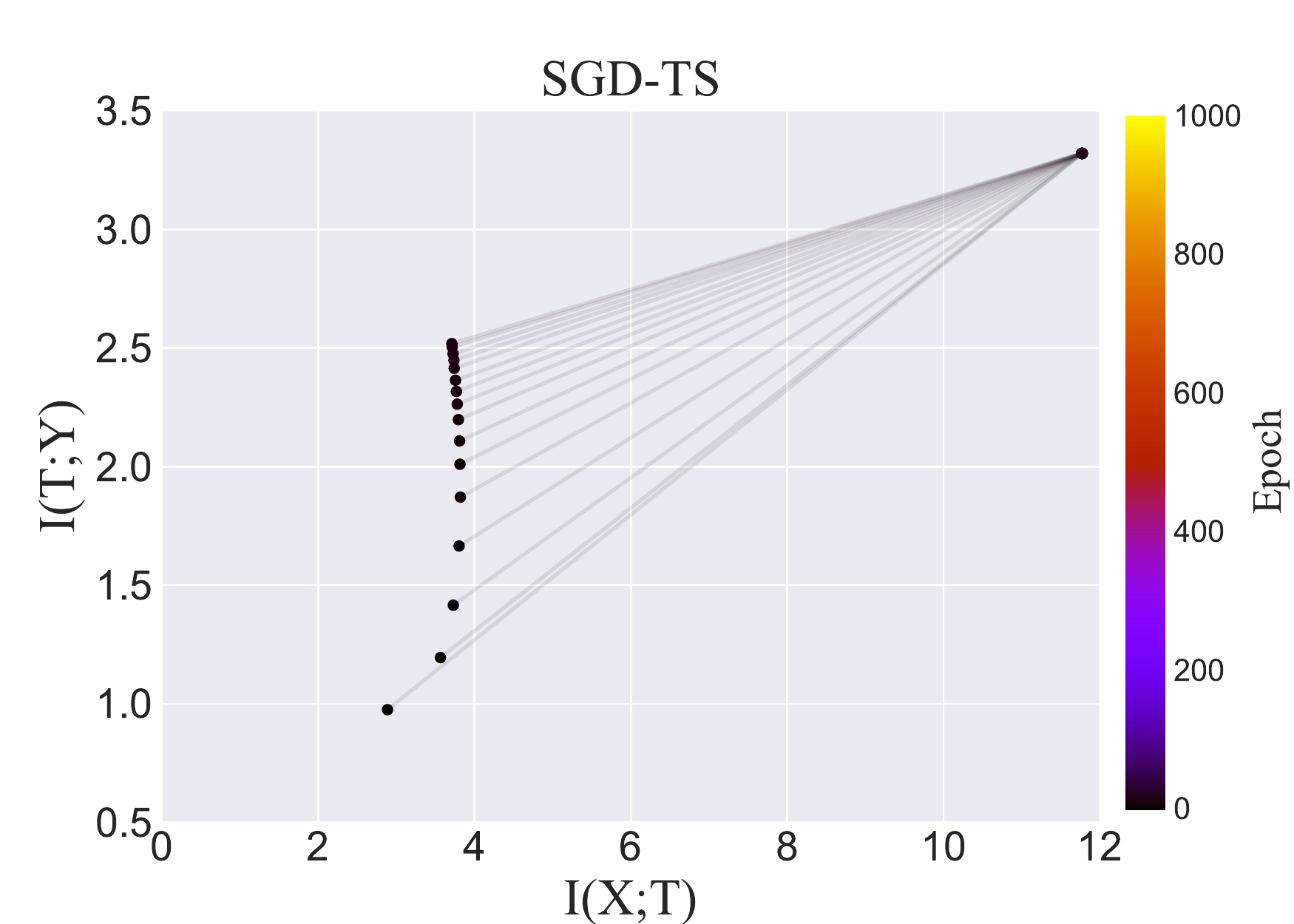}
	\end{subfigure}
	\hfill
	\begin{subfigure}[t]{0.24\textwidth}
		\includegraphics[width=\textwidth]{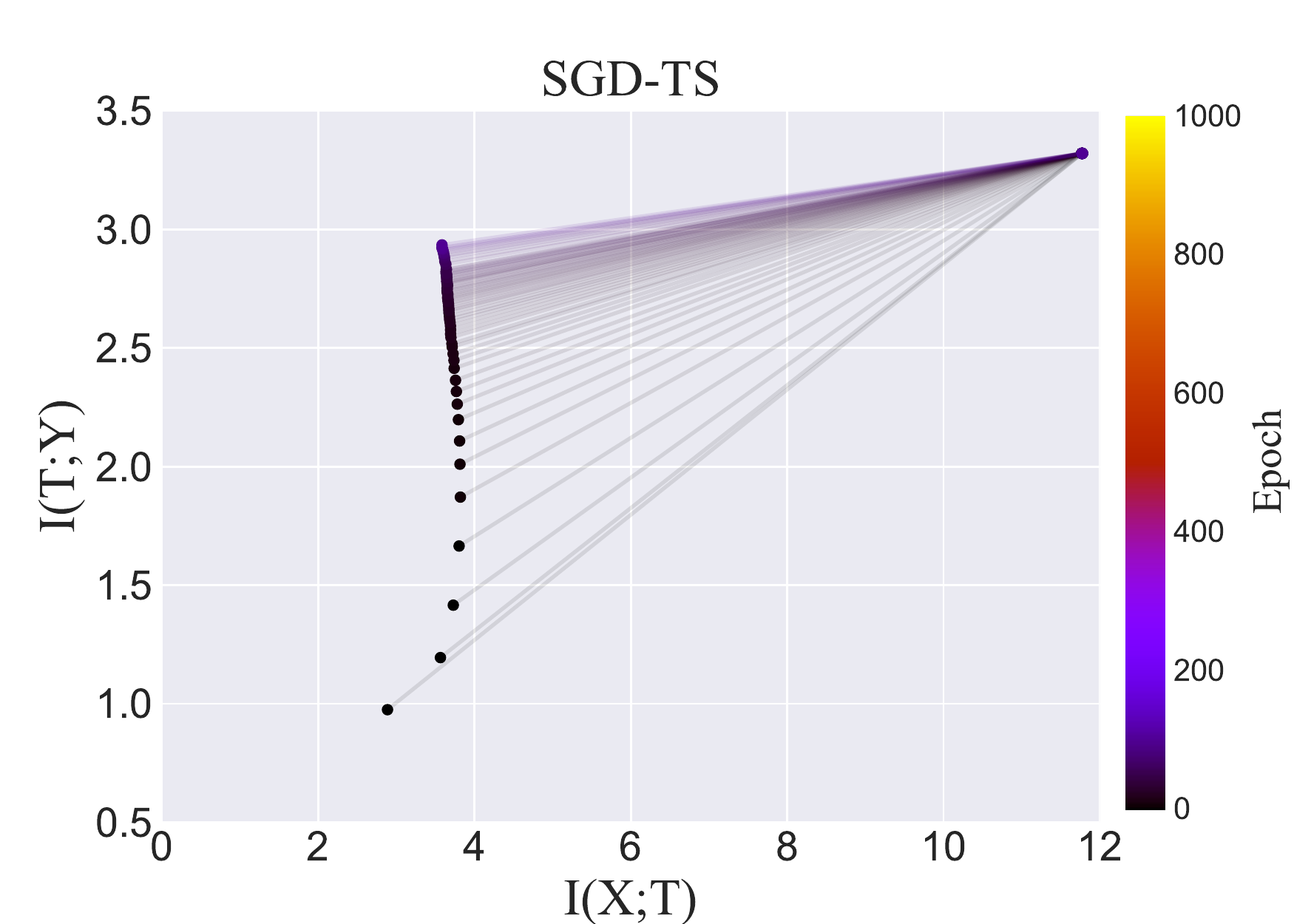}
	\end{subfigure}
	\hfill
	\begin{subfigure}[t]{0.24\textwidth}
		\includegraphics[width=\textwidth]{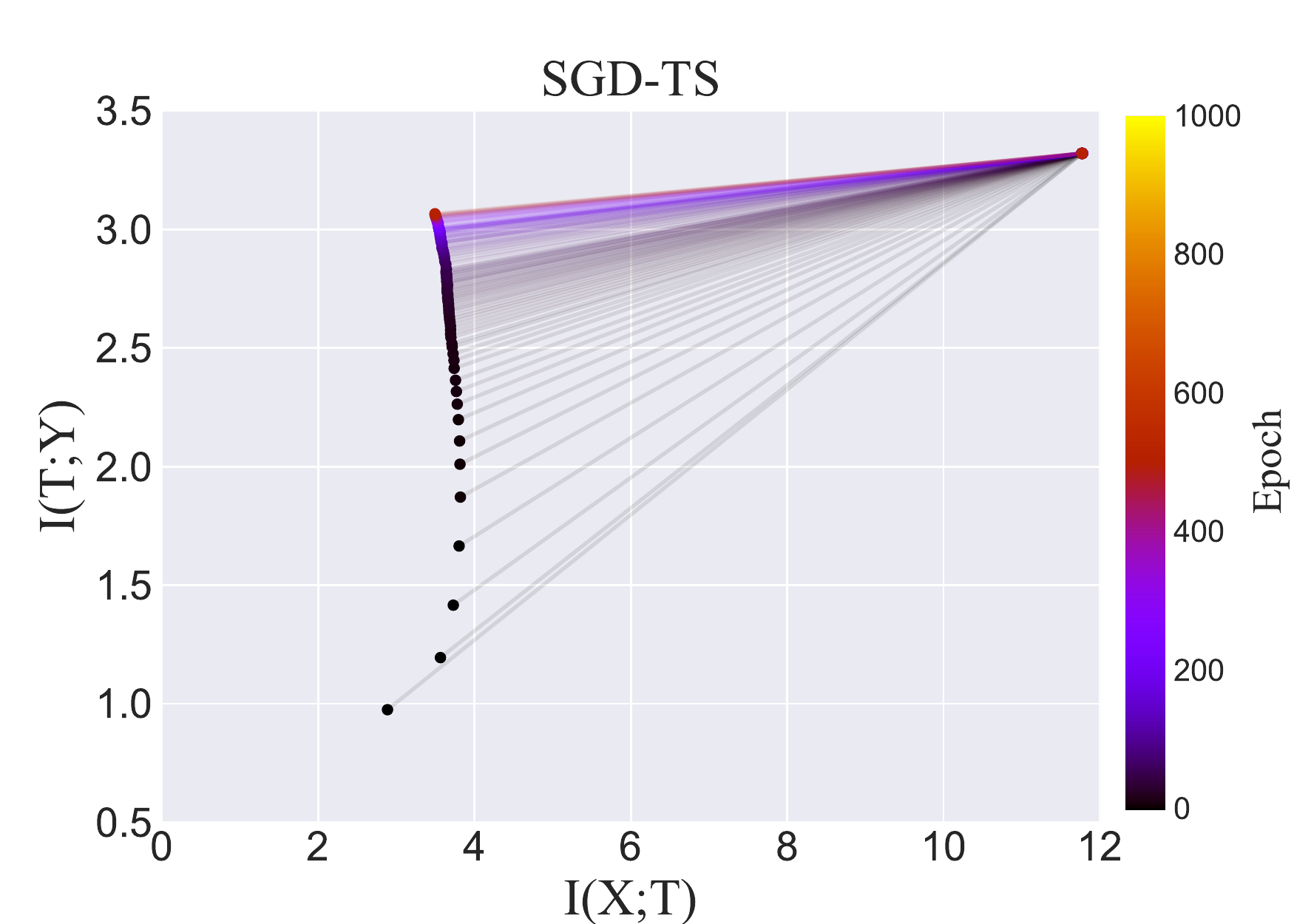}
	\end{subfigure} 
	\hfill
	\begin{subfigure}[t]{0.24\textwidth}
		\includegraphics[width=\textwidth]{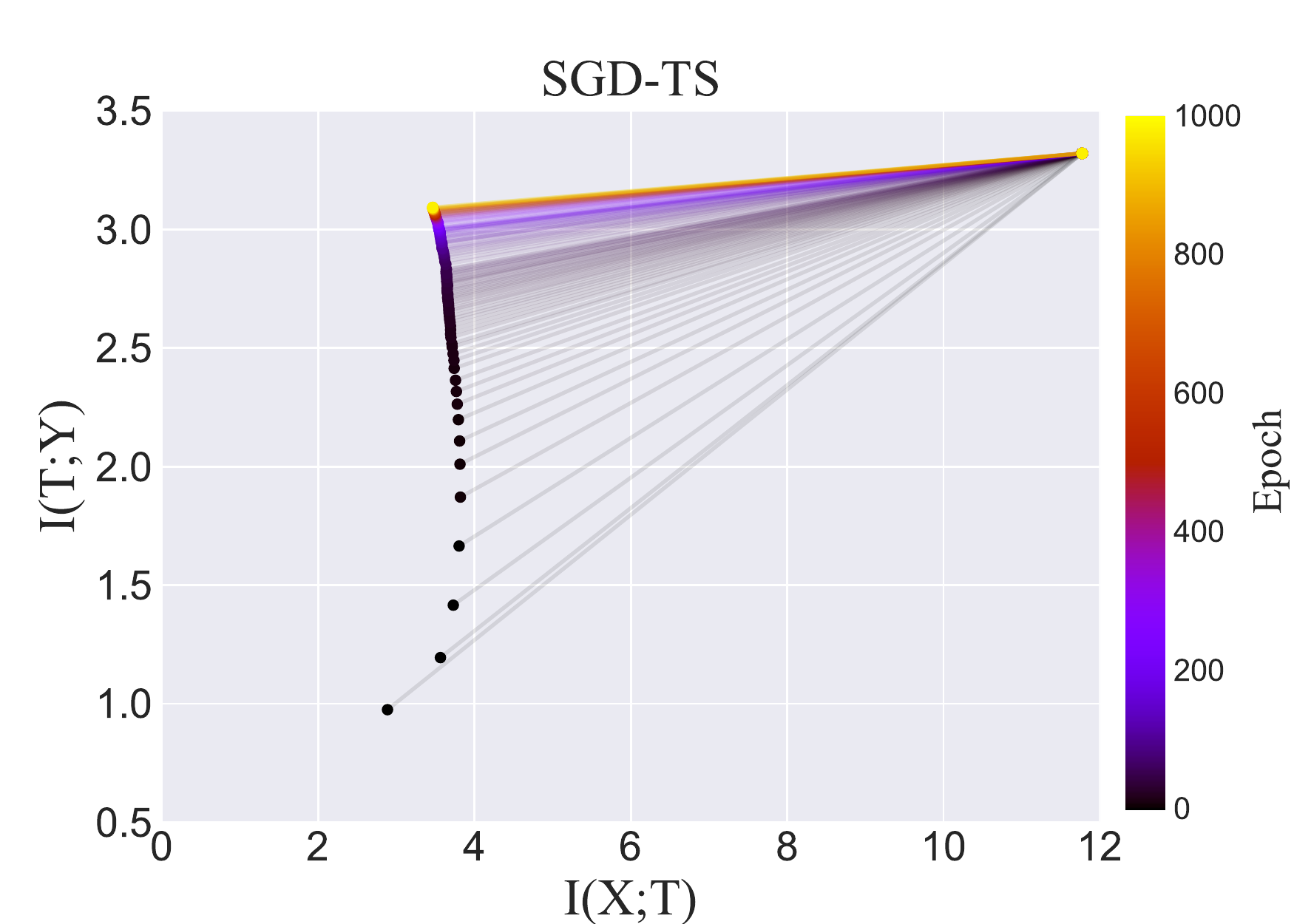}
	\end{subfigure}
	\begin{subfigure}[t]{\textwidth}
		\centering
		~\includegraphics[width=\textwidth]{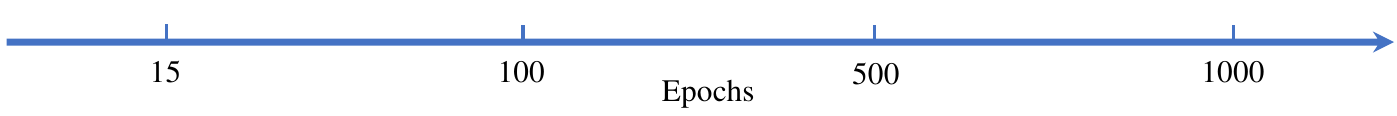}
	\end{subfigure}  
	\caption{The evolution of the layers information paths with different training epochs for Pendigits dataset, during the optimization of conventional minibatch SGD and typical batch SGD. Same training steps of different layers are connected by the gray thin lines. Left to right: at 15, 100, 500 and 1000 epochs. The first row shows the optimization process of conventional minibatch SGD, the second row shows the optimization process of typical batch SGD.}
	\label{fig:4}
\end{figure*}

\begin{figure*}[!t]
	\centering
	\begin{subfigure}[t]{0.24\textwidth}
		\includegraphics[width=\textwidth]{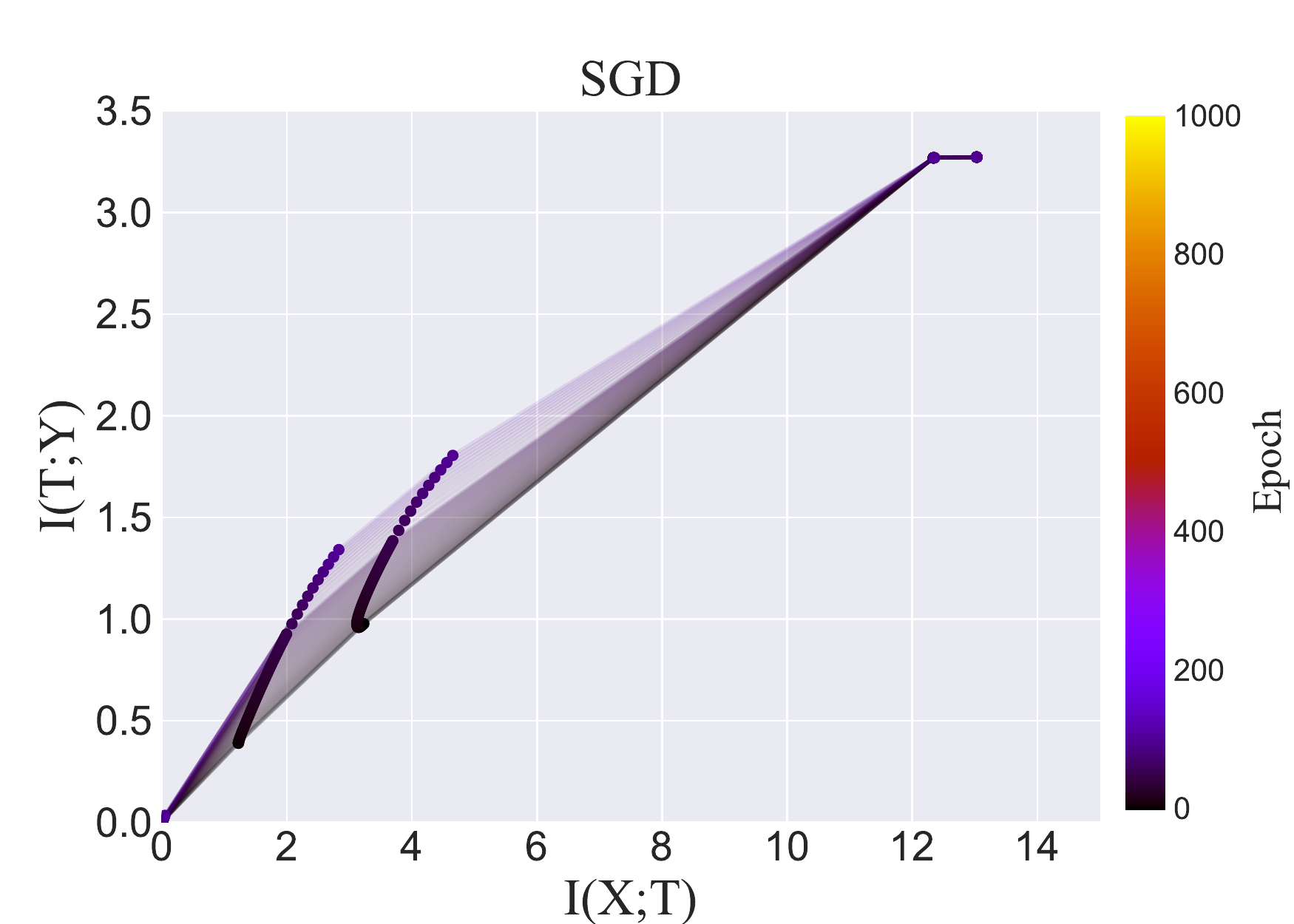}
	\end{subfigure}
	\hfill
	\begin{subfigure}[t]{0.24\textwidth}
		\includegraphics[width=\textwidth]{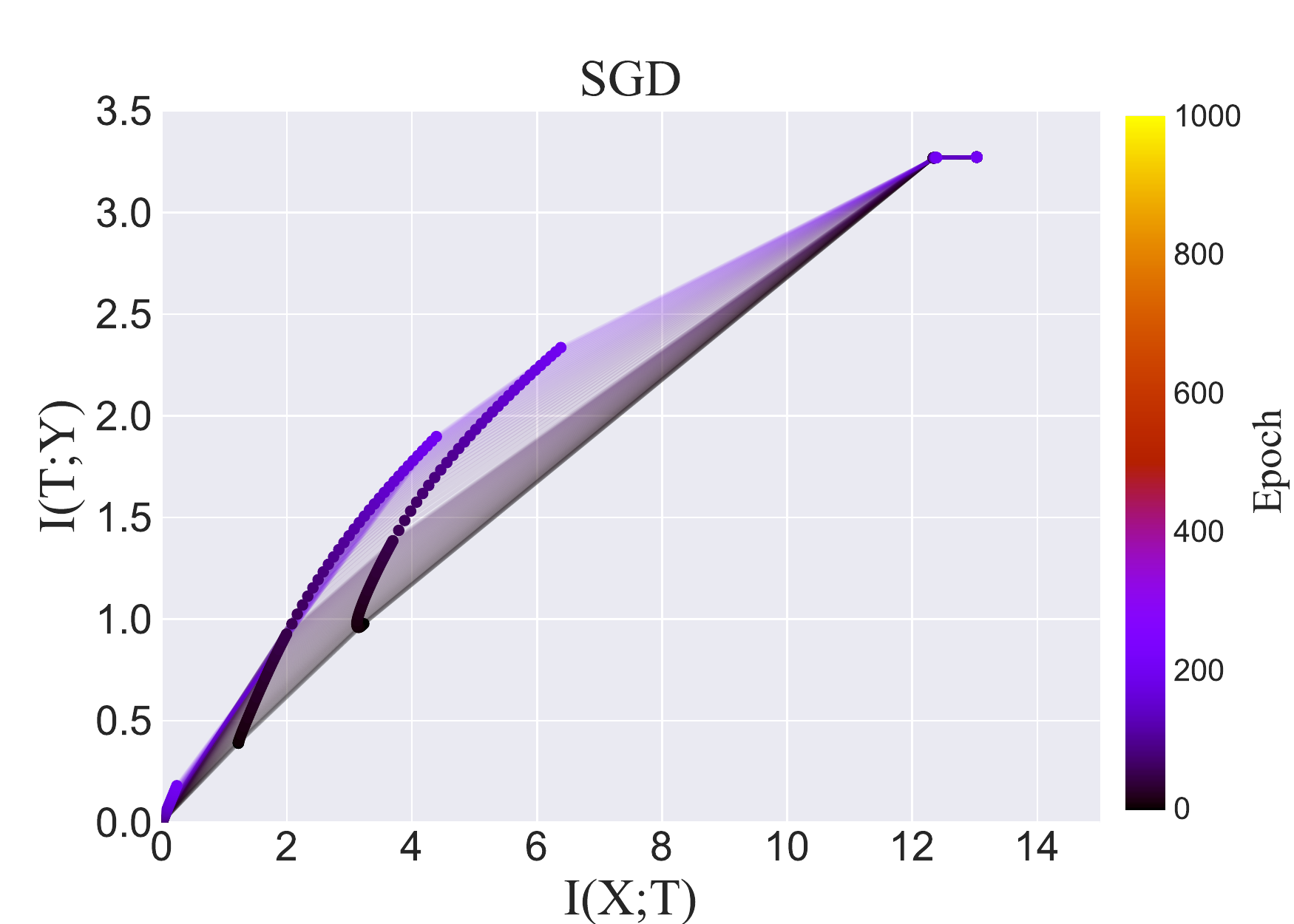}
	\end{subfigure}
	\hfill
	\begin{subfigure}[t]{0.24\textwidth}
		\includegraphics[width=\textwidth]{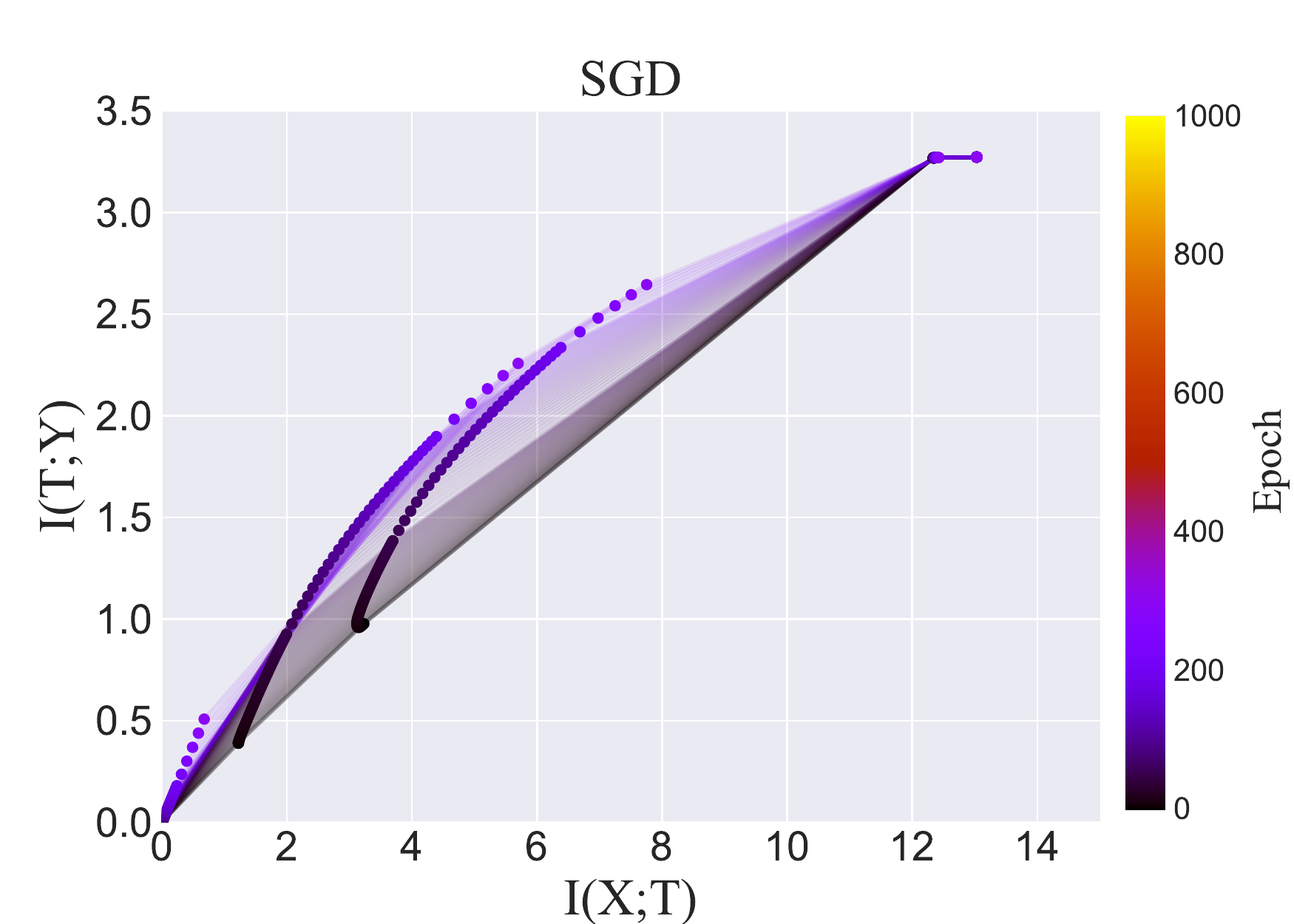}
	\end{subfigure} 
	\hfill
	\begin{subfigure}[t]{0.24\textwidth}
		\includegraphics[width=\textwidth]{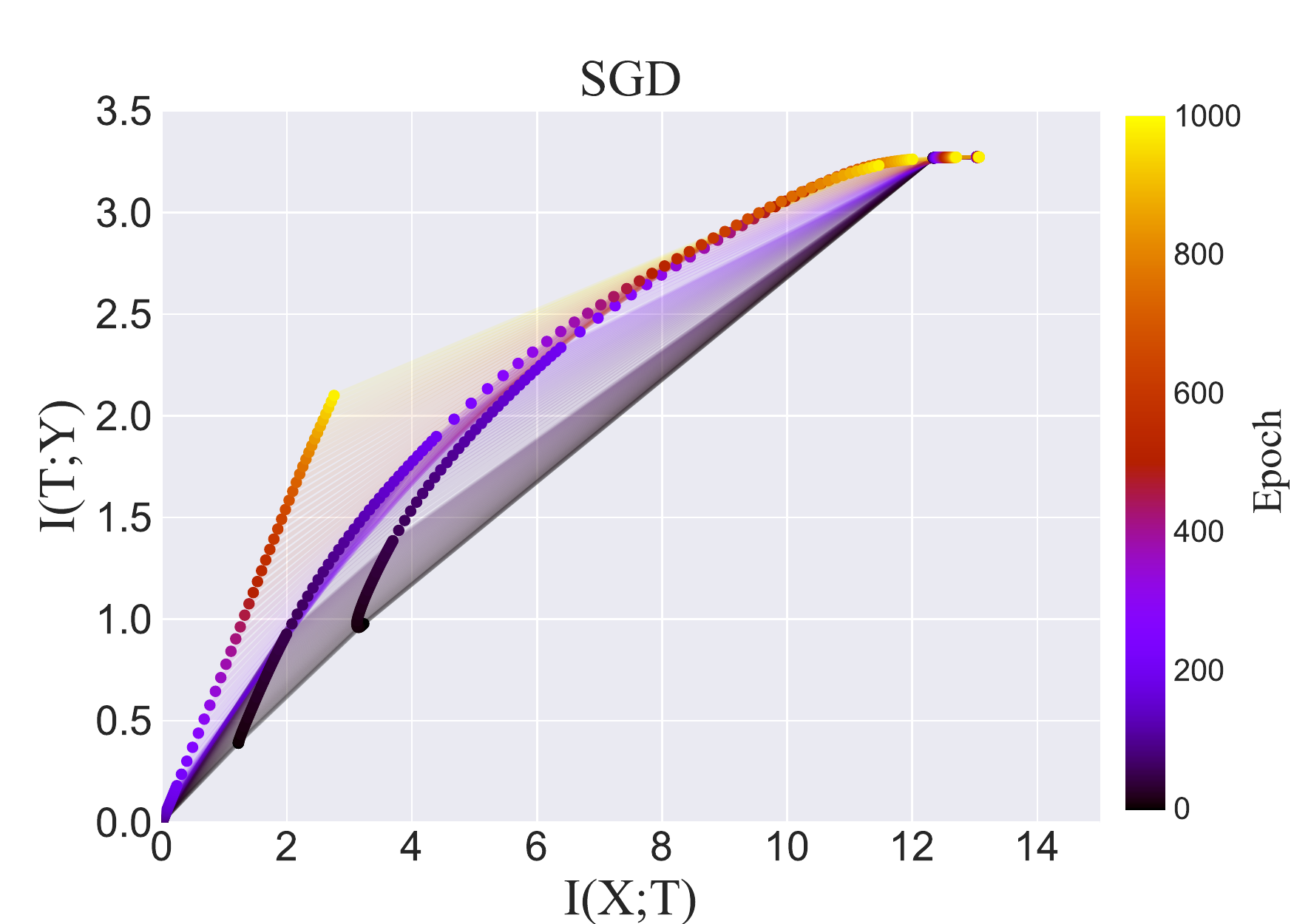}
	\end{subfigure} 
	\hfill
	\begin{subfigure}[t]{0.24\textwidth}
		\includegraphics[width=\textwidth]{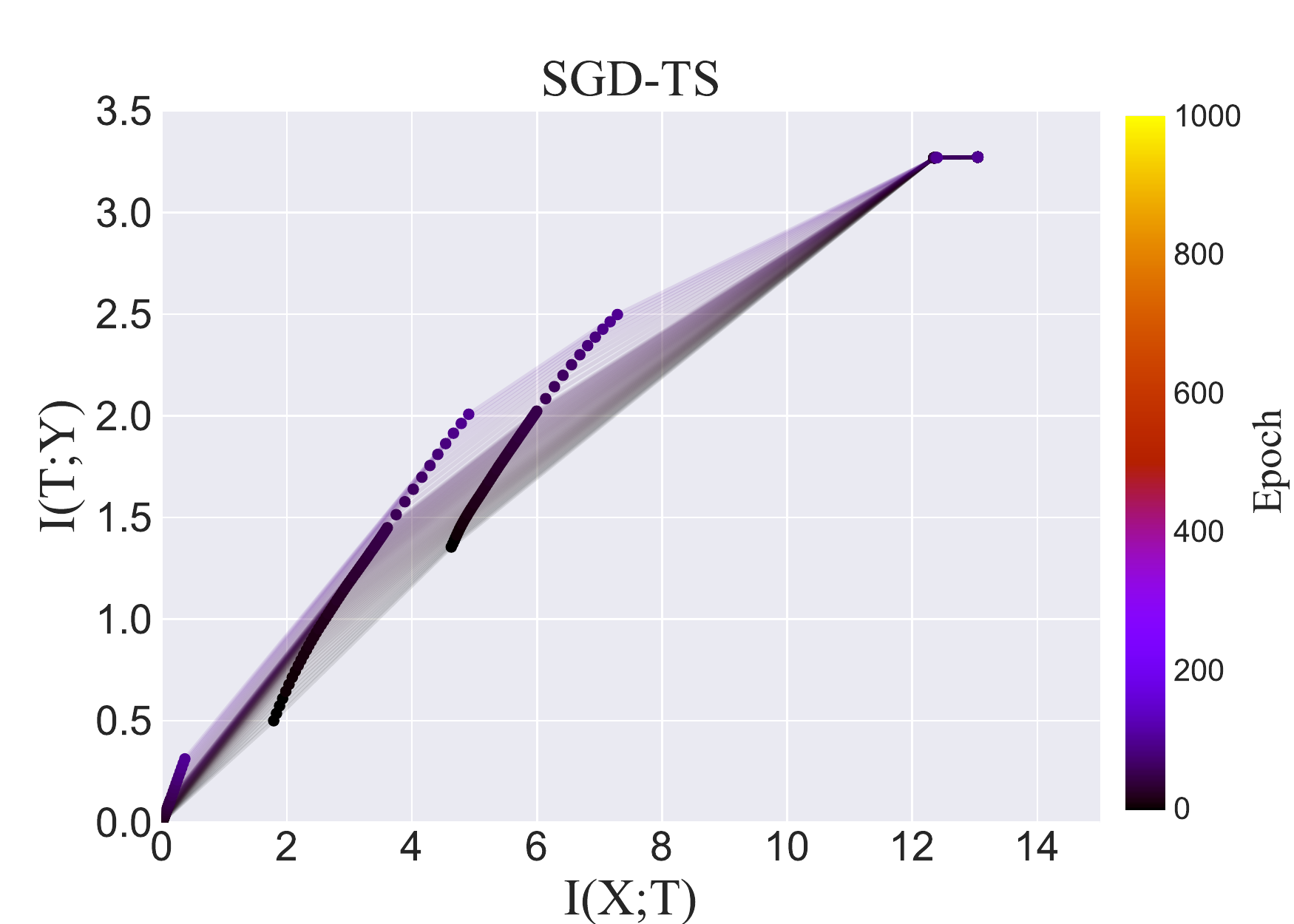}
	\end{subfigure}
	\hfill
	\begin{subfigure}[t]{0.24\textwidth}
		\includegraphics[width=\textwidth]{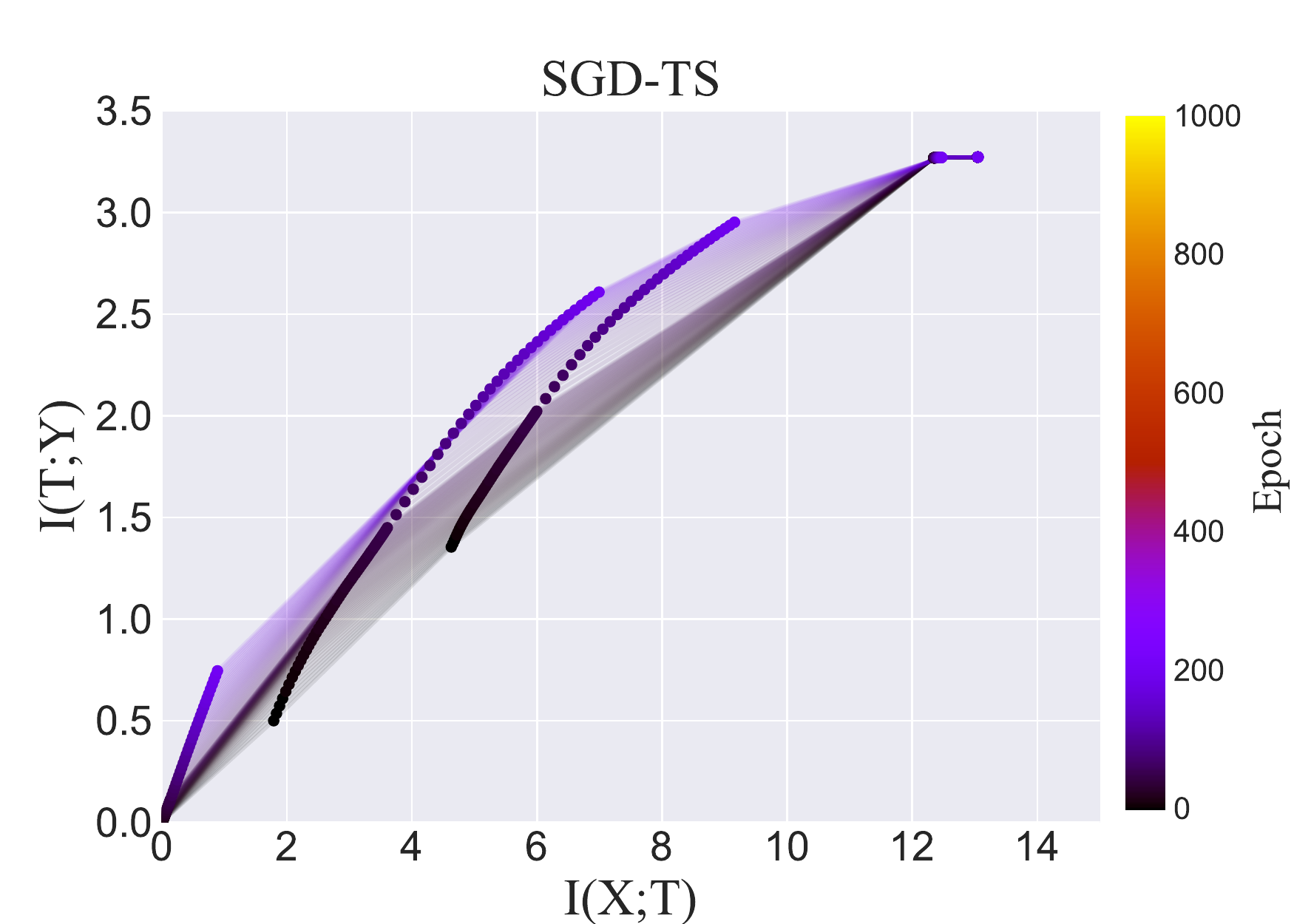}
	\end{subfigure}
	\hfill
	\begin{subfigure}[t]{0.24\textwidth}
		\includegraphics[width=\textwidth]{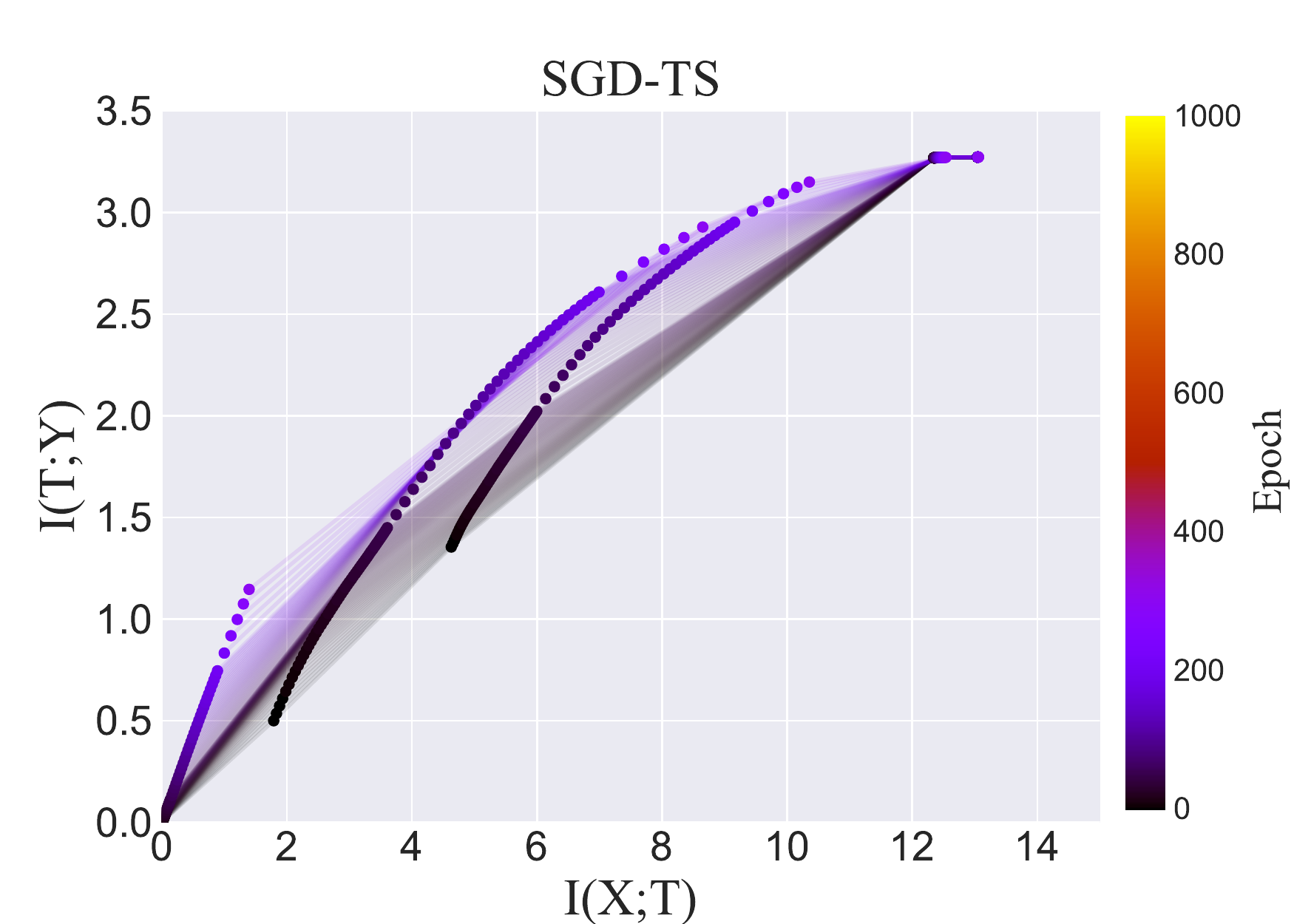}
	\end{subfigure} 
	\hfill
	\begin{subfigure}[t]{0.24\textwidth}
		\includegraphics[width=\textwidth]{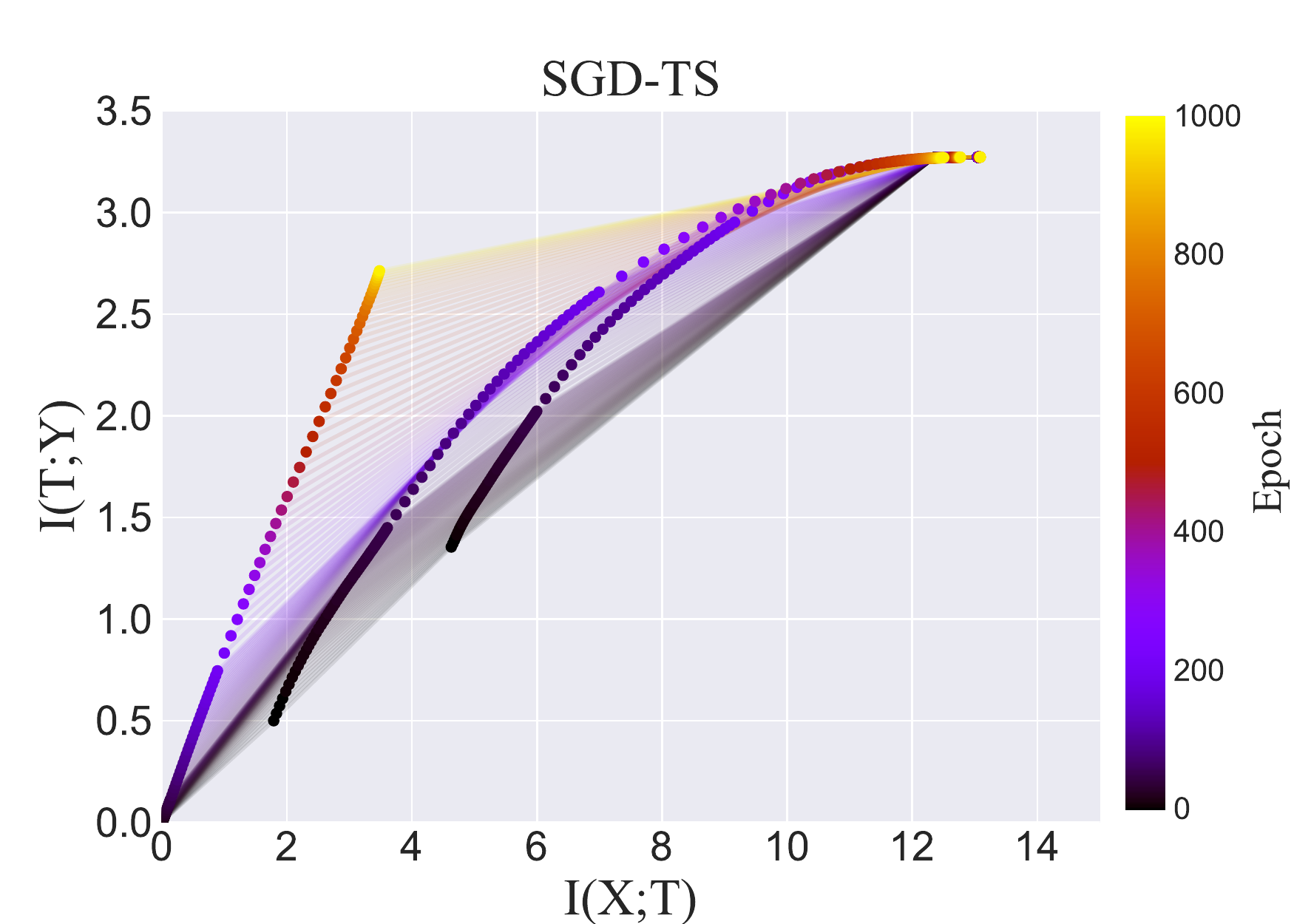}
	\end{subfigure}
	\begin{subfigure}[t]{\textwidth}
		\centering
		~\includegraphics[width=\textwidth]{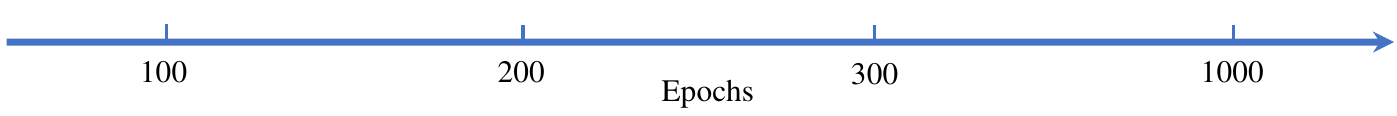}
	\end{subfigure}  
	\caption{The evolution of the layers information paths with different training epochs for Usps dataset, during the optimization of conventional minibatch SGD and typical batch SGD. Same training steps of different layers are connected by the gray thin lines. Left to right: at 100, 200, 300 and 1000 epochs. The first row shows the optimization process of conventional minibatch SGD, the second row shows the optimization process of typical batch SGD.}
	\label{fig:5}
\end{figure*}

\subsection{Experimental Setup}
\label{subsection41}
In this section, we verify the above theoretical analysis findings by the experimental results on both synthetic and real-world datasets. We compare the performance of typical batch SGD (denoted as SGD-TS) against conventional minibatch SGD in the information plane. Training is conducted with fixed batch size and step size. No tuning skill, including regularization, is used in both experiments.

To estimate the mutual information of each hidden layer with respect to the input and the label, we use the kernel density estimator approach proposed in \citep{kolchinsky2017estimating} which assumes the distribution of the activity of the hidden layer is a mixture of Gaussians. This assumption is suitable for this case since we add Gaussian white noise to each hidden layer to avoid infinite mutual information value, making the distribution a mixture of Gaussians \citep{saxe2019information}. 

\subsection{Experiments on Synthetic Dataset}
\label{subsection42}
We start with the experiments on the same synthetic dataset in line with the work \citep{shwartz2017opening}. This dataset that contains $4096$ different patterns is generated by assigning binary values to $12$ uniformly distributed points on a $2$D sphere. The neural network we train on this dataset consists of $7$ hidden layers of width 12-10-7-5-4-3-2. Each layer is followed by a tanh activation function except the last layer with a sigmoid function.

We first verify Assumption \ref{assumption1} about the squared L2 norm of sample gradients made in Section \ref{section3}. The numerical result in Figure \ref{fig:1} shows that for each layer of the network, the averaged L2 norm of gradients across the samples in subset $\mathcal{H}$ and $\mathcal{L}$ are barely the same during the whole optimization process, thus verifying our assumption. For more validations on real-world datasets, see Appendix \ref{App-A}.

Then we visualize the performance of typicality sampling on the synthetic dataset. The experimental results are shown in Figure \ref{fig:2}. Consistent with the observation in \citep{shwartz2017opening}, the existence of an initial fitting phase and the following compression phase is explicit. We find that the transition occurs at about $300$ epochs for typical batch SGD, while it is $500$ epochs for conventional minibatch SGD. This implies that the typicality sampling scheme significantly boosts the fitting phase.

\begin{figure*}[!t]
	\centering
	\includegraphics[width=\textwidth, scale=1]{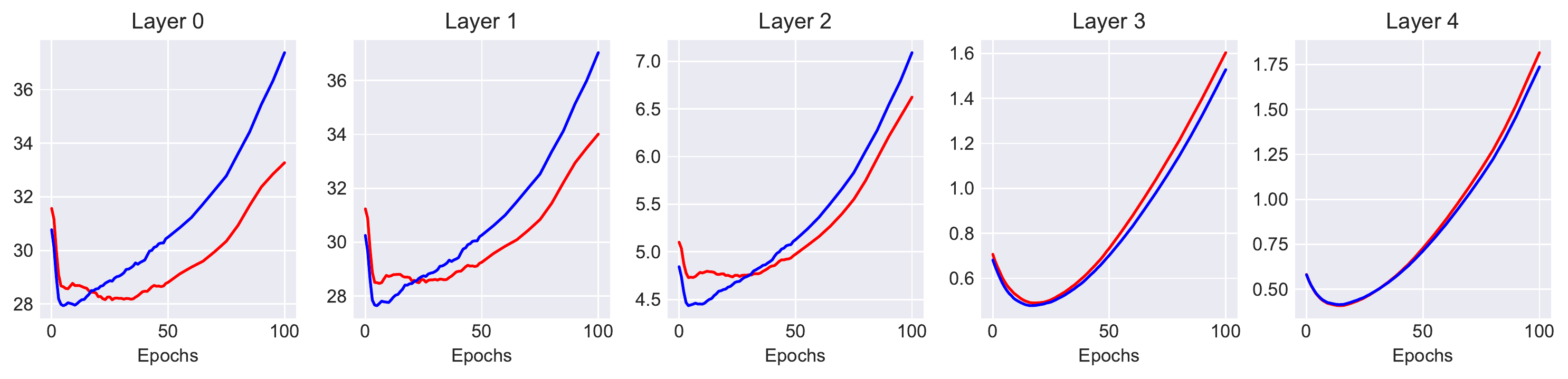}
	\caption{The averaged squared L2 norm of gradient across the samples in each subset $\mathcal{H}$ and $\mathcal{L}$ for Usps dataset. The red curves represent the results on subset  $\mathcal{H}$ and the blue curves represent the results on subset $\mathcal{L}$.}
	\label{fig:7}
\end{figure*}

\begin{figure*}[!t]
	\centering
	\includegraphics[width=\textwidth, scale=1]{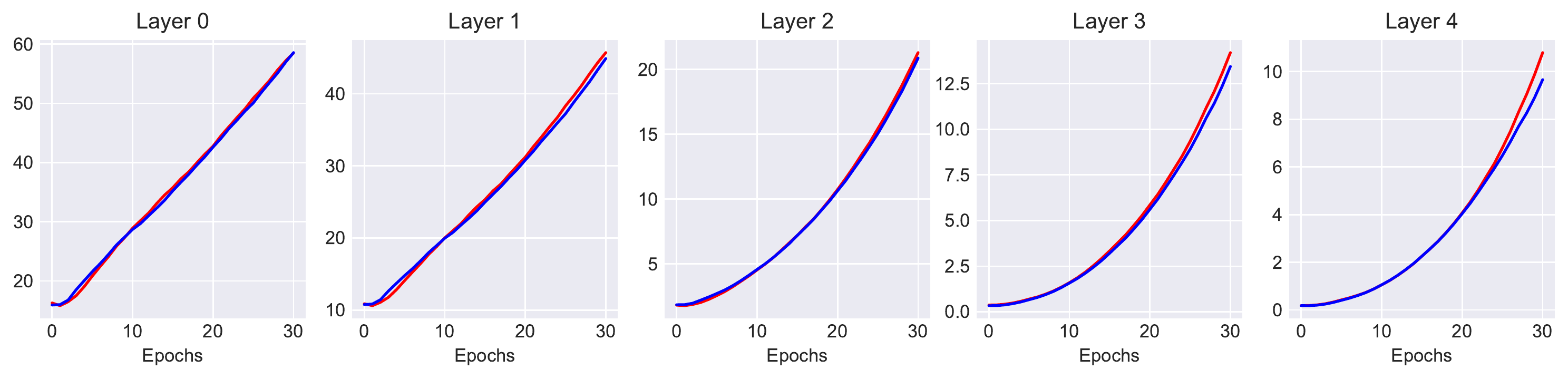}
	\caption{The averaged squared L2 norm of gradient across the samples in each subset $\mathcal{H}$ and $\mathcal{L}$ for MNIST dataset. The red curves represent the results on subset  $\mathcal{H}$ and the blue curves represent the results on subset $\mathcal{L}$.}
	\label{fig:8}
\end{figure*}

\subsection{Experiments on Real-world Dataset}
\label{subsection43}
To be more rigorous, we also evaluate typicality sampling on real-world datasets. To maintain consistency, the datasets we use here, including Pendigits \citep{Dua:2019}, Usps \citep{hull1994database}, and MNIST \citep{lecun1998gradient}, are the same as the ones in \citep{peng2019accelerating}. The detailed information about these datasets and the architectures of the corresponding neural networks we train on them are presented in Table \ref{tab:1}. Here we use ReLU instead of tanh as the activation function for the neural networks to valid if our theoretical analysis holds when the compression phase does not exist explicitly, as suggested in \citep{saxe2019information}.

The results are shown in Figure \ref{fig:3}--\ref{fig:5}. We can find that across all datasets, the fitting phases are still accelerated by the typicality sampling scheme, even though there are no compression phases in the optimization. The typical batch SGD converges to the same point in the information plane faster than conventional minibatch SGD, and this improvement is more notable for the deeper hidden layers. We also notice that for some datasets, e.g., Pendigits, typicality sampling can help the training model yield better prediction accuracy by capturing more information about the label at the end of the training. We also conduct an additional experiment on more complex CIFAR-10 \citep{krizhevsky2009learning} to further verify our findings. See Appendix \ref{App-B} for more details. All results above demonstrate that the typicality sampling scheme helps the training dynamics by speeding up the fitting phase of the training process, and thus accelerates minibatch SGD.

\section{Conclusion}
\label{section5}

From the viewpoint of the IB theory, this paper has given another explanation of the acceleration capability of typicality sampling for SGD. Different from the former work that mainly focuses on the averaged convergence rate and credits the improvement to the small noise in gradient approximation, we dug into the dynamics of learning to study the performance of typicality sampling in distinct phases of the training process. Our theoretical analysis showed that typicality sampling generates a higher SNR of gradient approximation that helps shorten the fitting phase and accelerate minibatch SGD. Experimental results on both synthetic and real-world demonstrated our findings. This finding also implies that the prior information of training set can help expedite the process of finding the minimal sufficient statistic for the deep learning system. The typicality sampling method may not be the best way to take advantage of the prior information, but it achieves good results with clear intuition and solid theoretical guarantee.

Besides, we empirically showed a new property of the typical samples that the information they contain about the L2 norm of the gradients is the same as the rest samples, although the typical samples can provide the most informative gradient in guiding the search direction. One potential area for future work is to theoretically prove this property and further reveal the structure of the typical samples.

\appendices
\section{Validation of Assumption \ref{assumption1} on Real-world Dataset}
\label{App-A}
{\color{black}Here we verify Assumption \ref{assumption1} on two real-world datasets, including Usps and MNIST, to confirm the generality of this assumption. All the experimental setup for these two datasets are the same as in Section \ref{section4}. Figure \ref{fig:7}--\ref{fig:8} show the results, respectively. We can see that all results support our assumption, especially at the early stage of the optimization which corresponds to the fitting phase in the IB theory.}

\section{Experiments on CIFAR-10 Dataset}
\label{App-B}
\begin{figure*}[!h]
	\centering
	\begin{subfigure}[t]{0.9\textwidth}
		\captionsetup[subfigure]{singlelinecheck=off,justification=raggedright}
		\begin{subfigure}[t]{\textwidth}
			\setlength{\abovecaptionskip}{-0.0cm}
			\setlength{\belowcaptionskip}{-0.3cm}
			\subcaption*{Layer $4$}
			\includegraphics[width=0.24\textwidth]{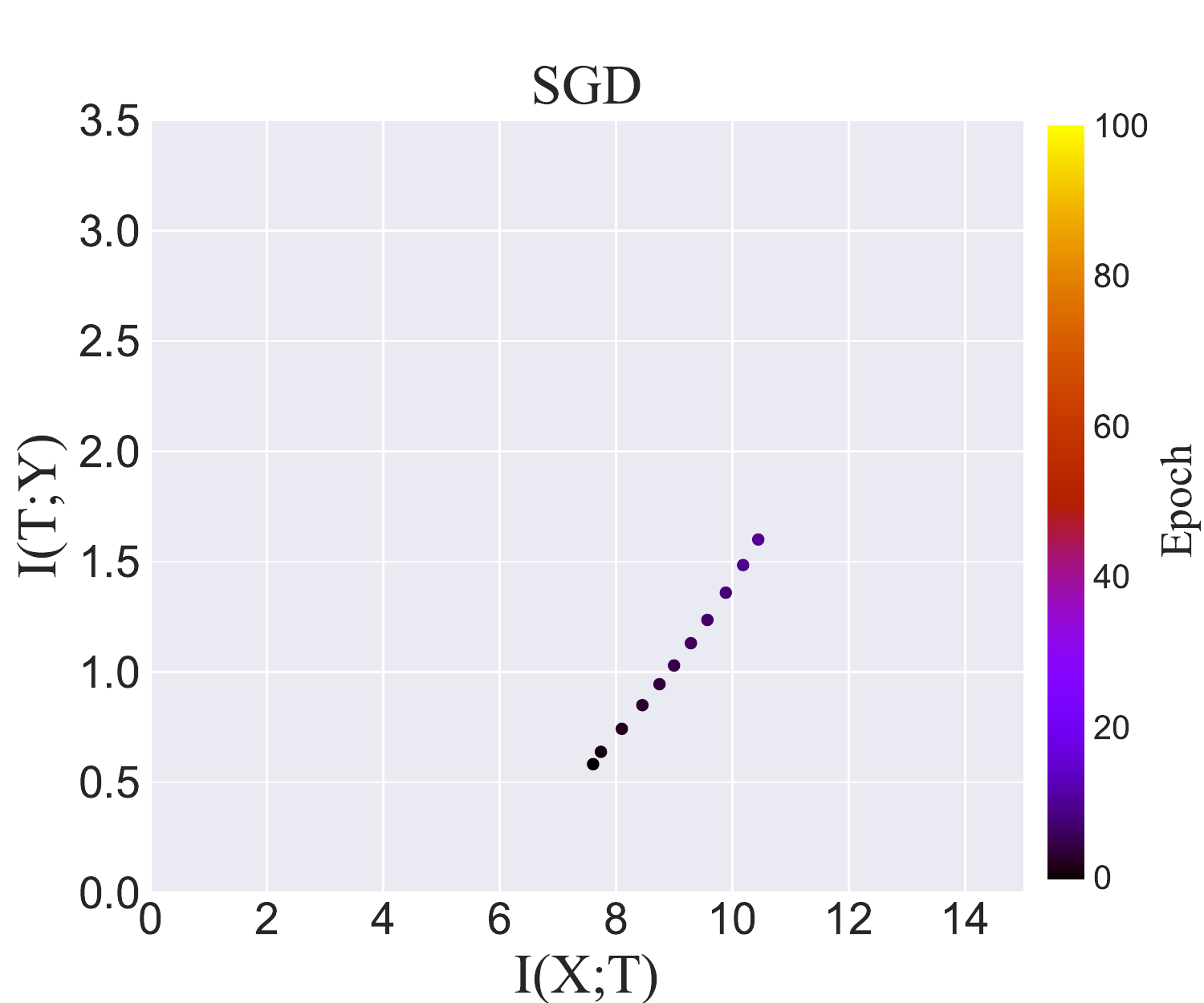}
			\includegraphics[width=0.24\textwidth]{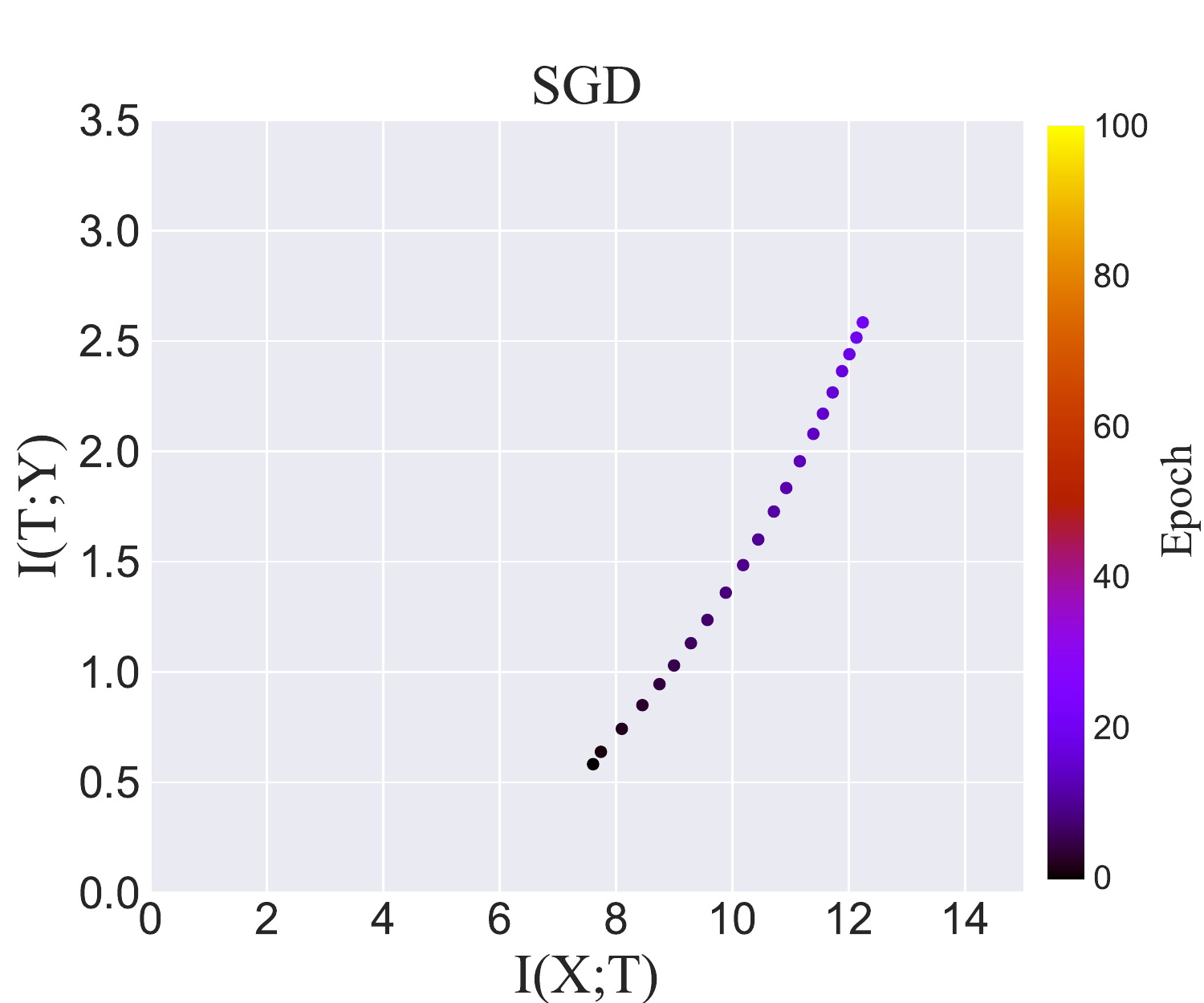}
			\includegraphics[width=0.24\textwidth]{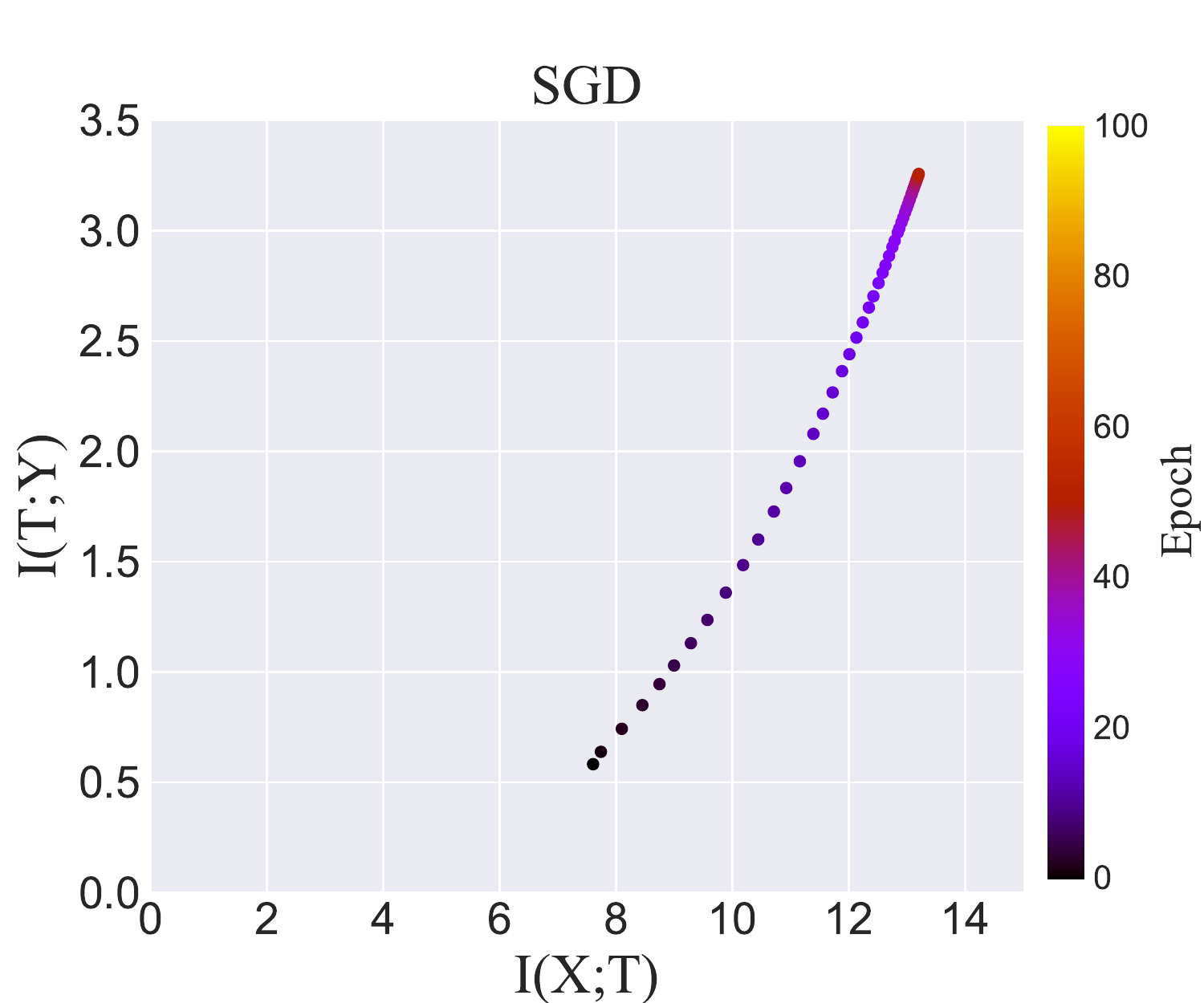}
			\includegraphics[width=0.24\textwidth]{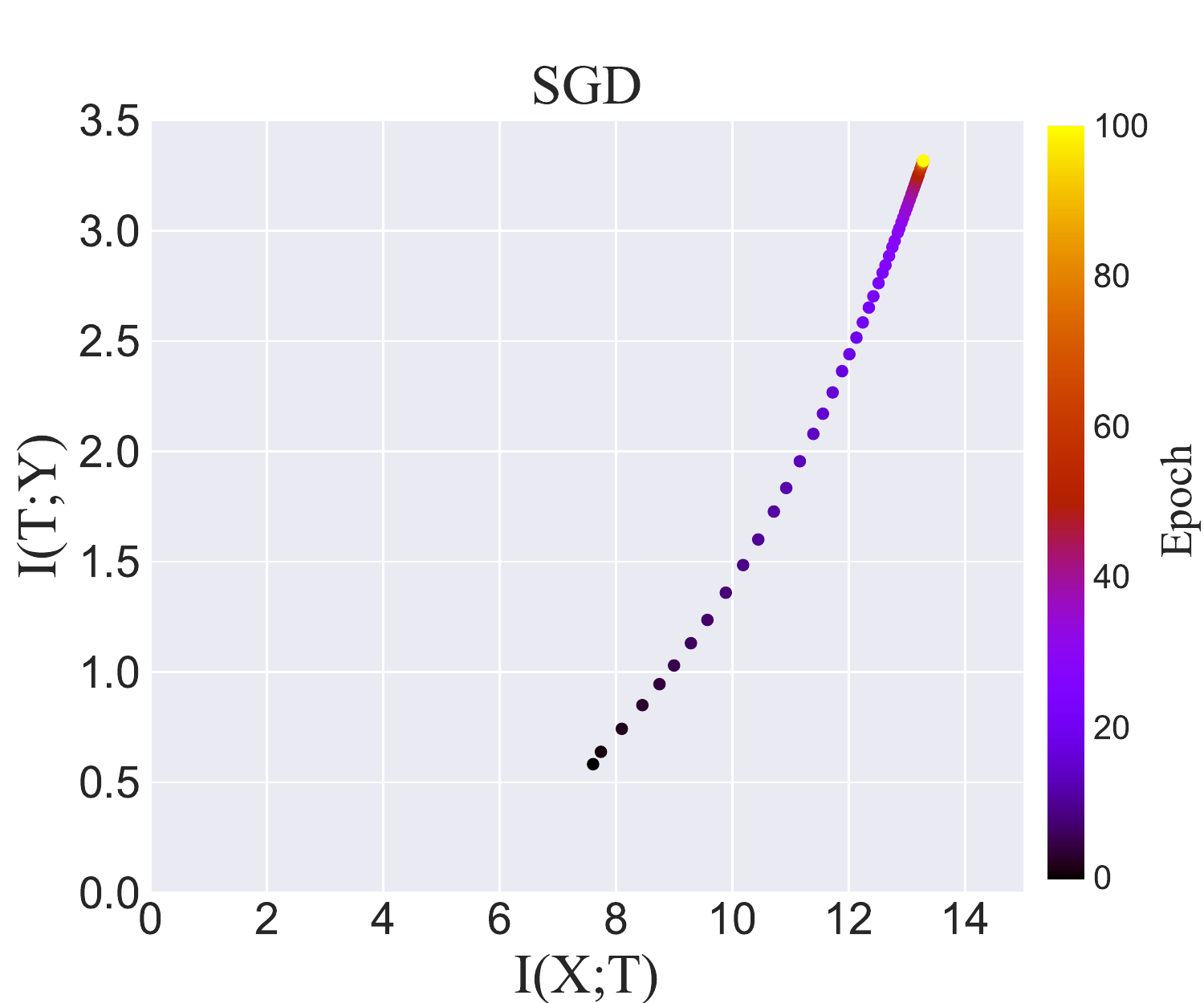}		
		\end{subfigure}
		\begin{subfigure}[t]{\textwidth}
			\includegraphics[width=0.24\textwidth]{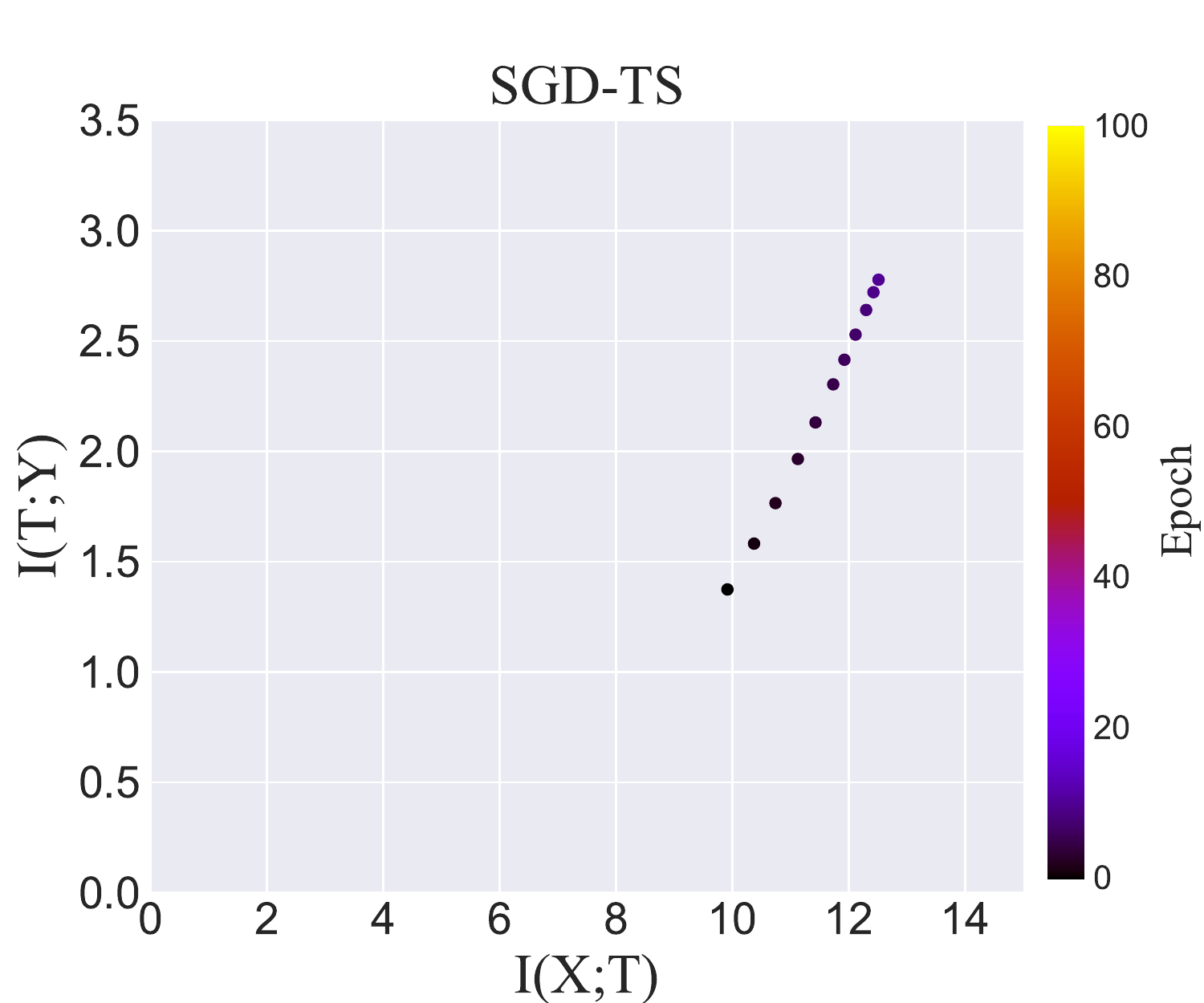}
			\includegraphics[width=0.24\textwidth]{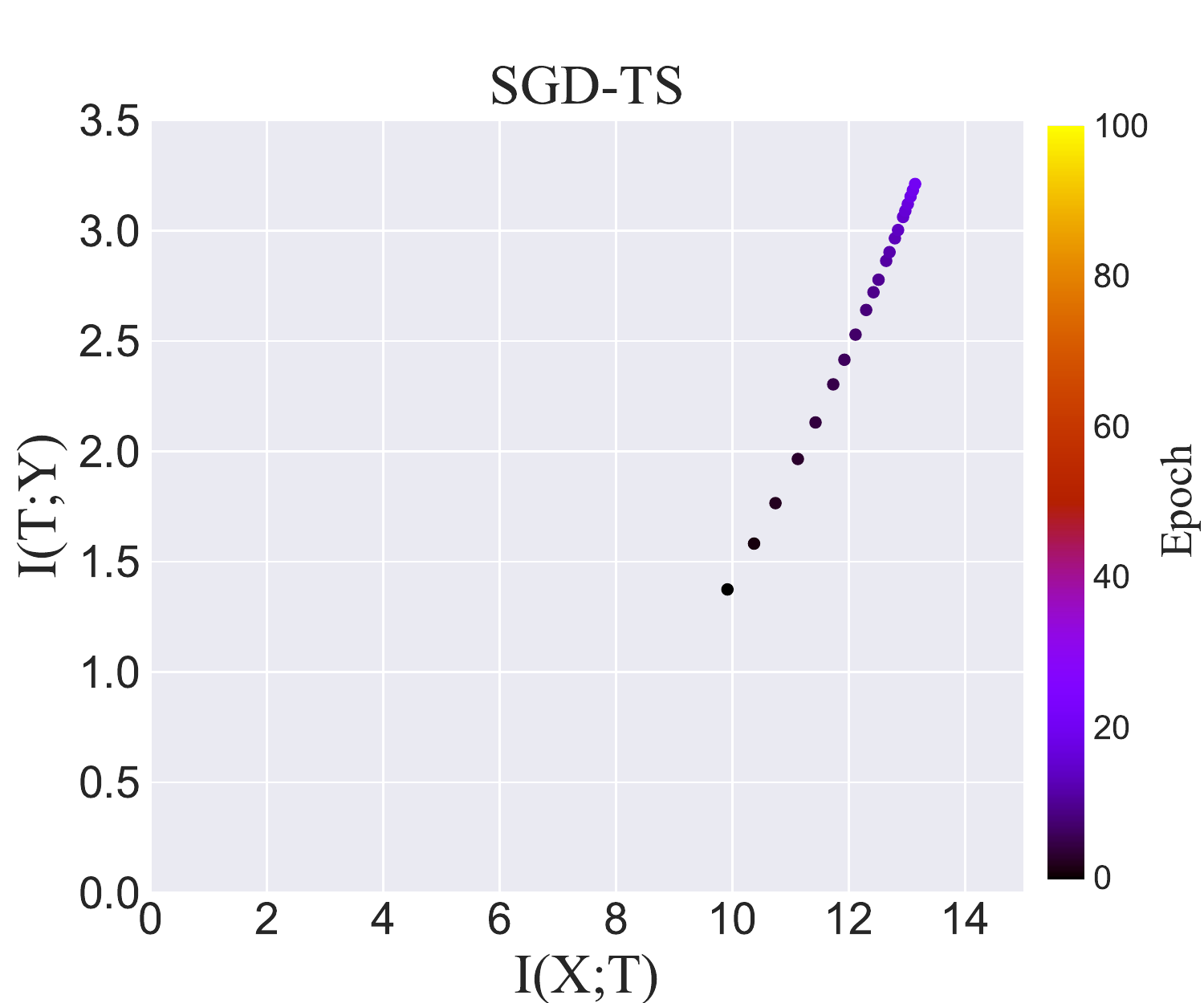}
			\includegraphics[width=0.24\textwidth]{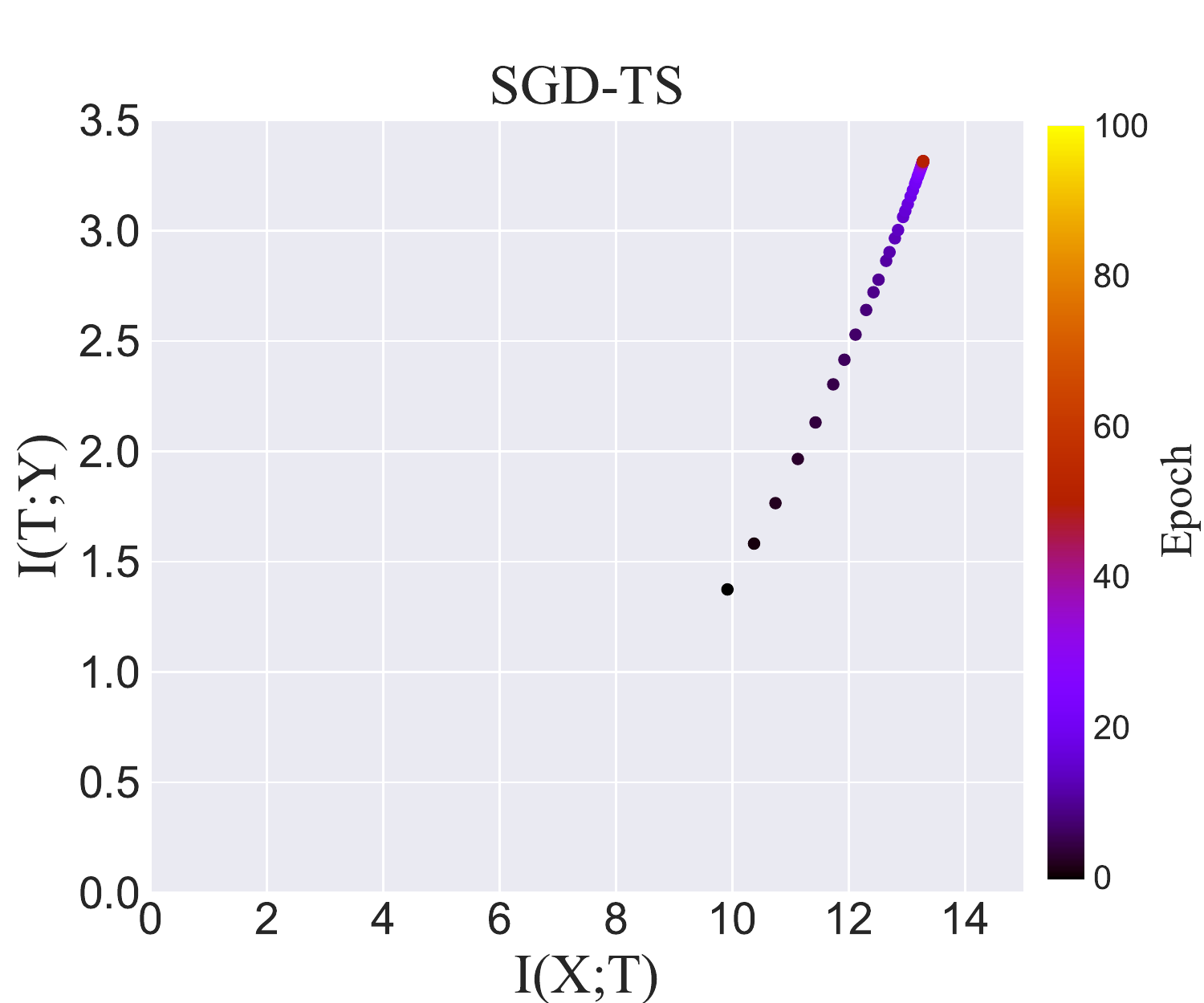}
			\includegraphics[width=0.24\textwidth]{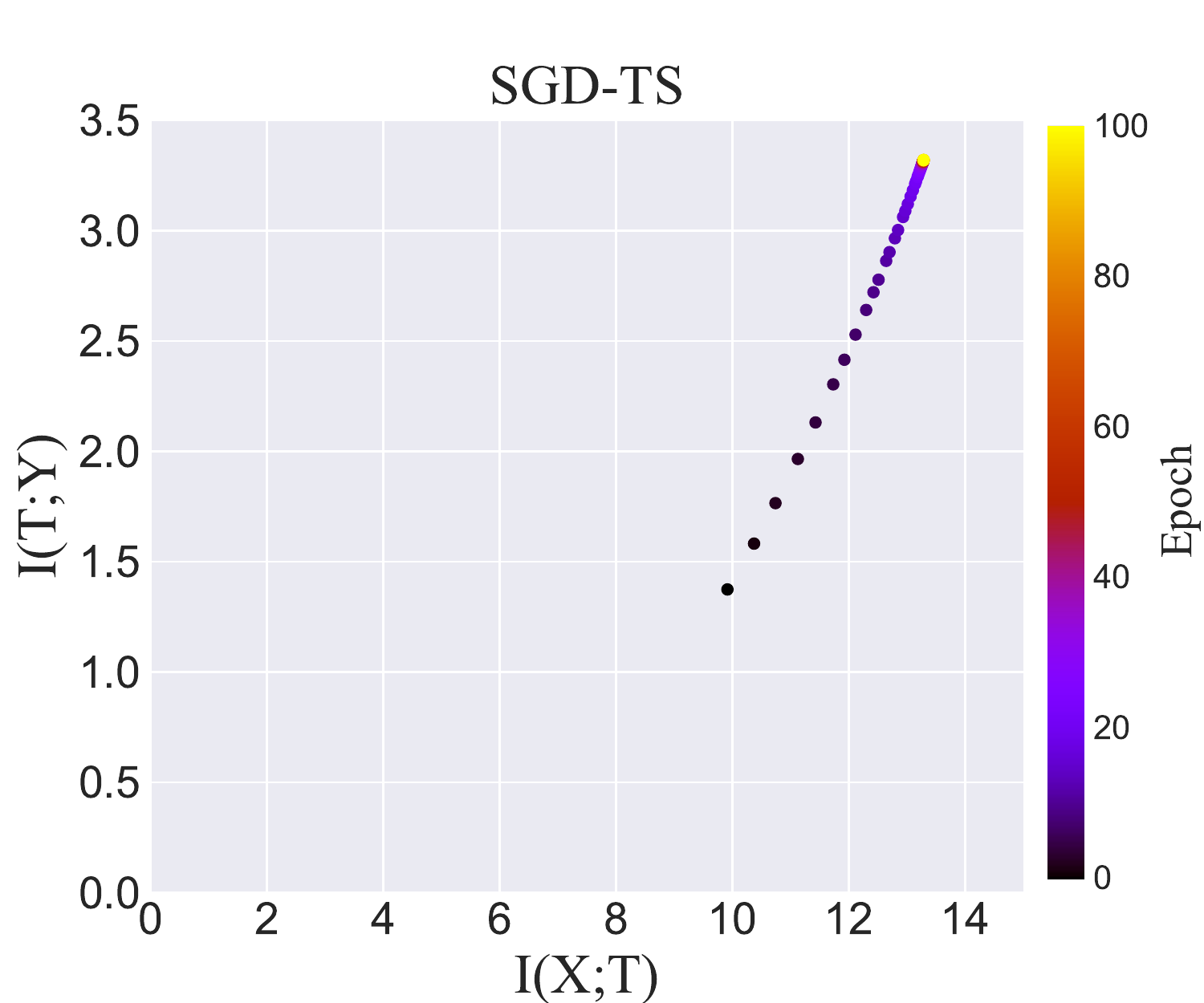}	
		\end{subfigure}
		\begin{subfigure}[t]{\textwidth}
			\setlength{\abovecaptionskip}{-0.0cm}
			\setlength{\belowcaptionskip}{-0.3cm}
			\subcaption*{Layer $5$}
			\includegraphics[width=0.24\textwidth]{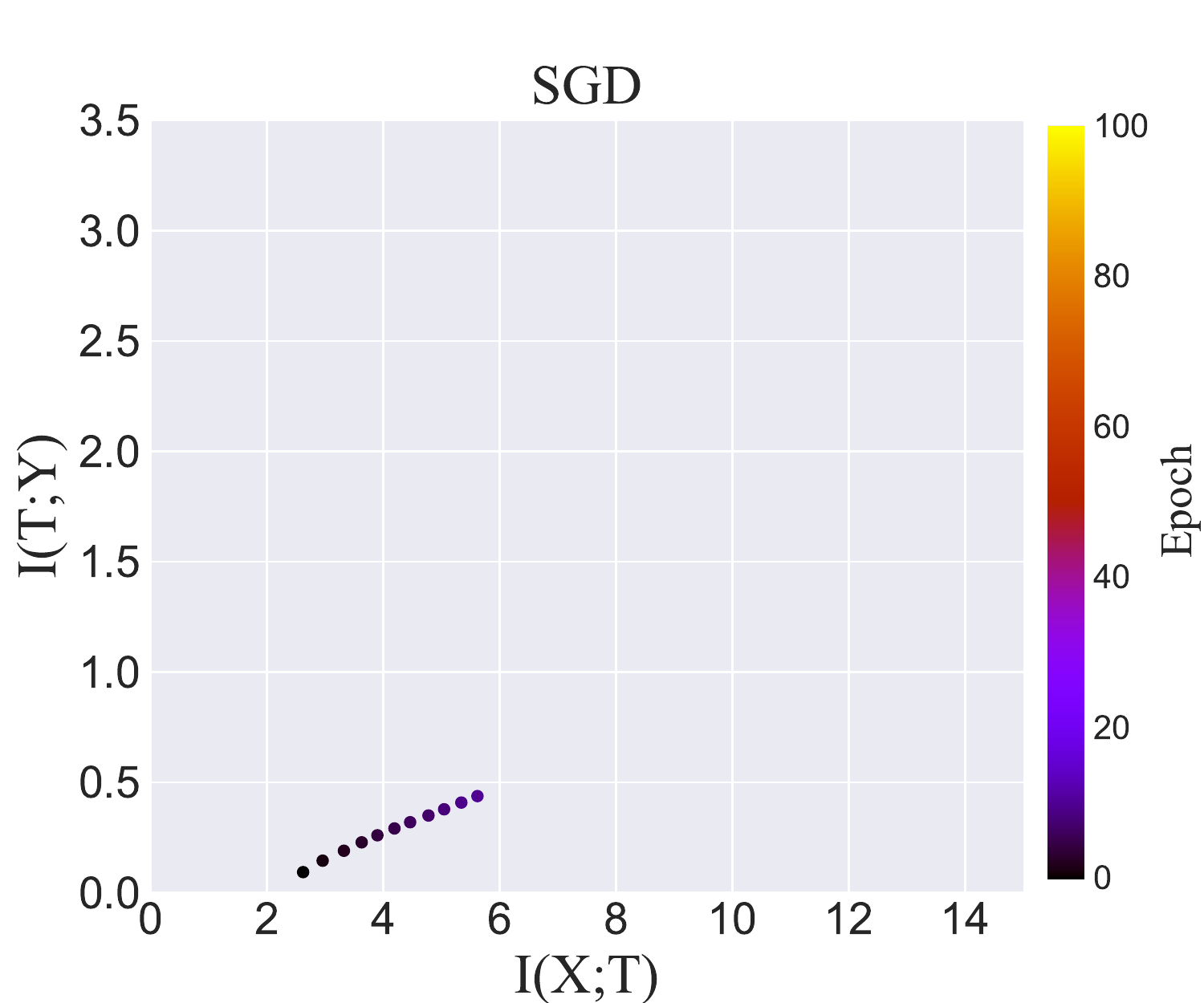}
			\includegraphics[width=0.24\textwidth]{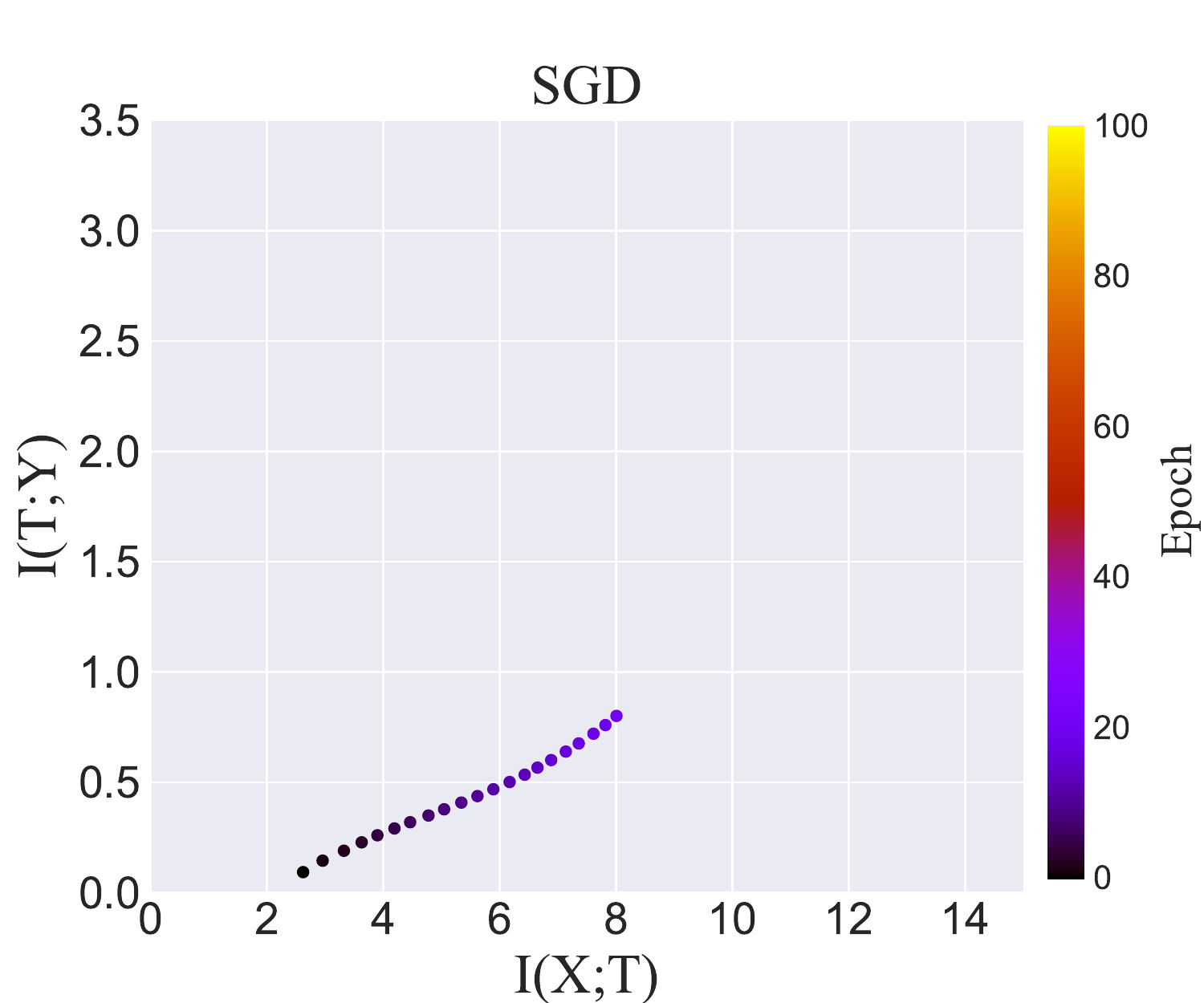}
			\includegraphics[width=0.24\textwidth]{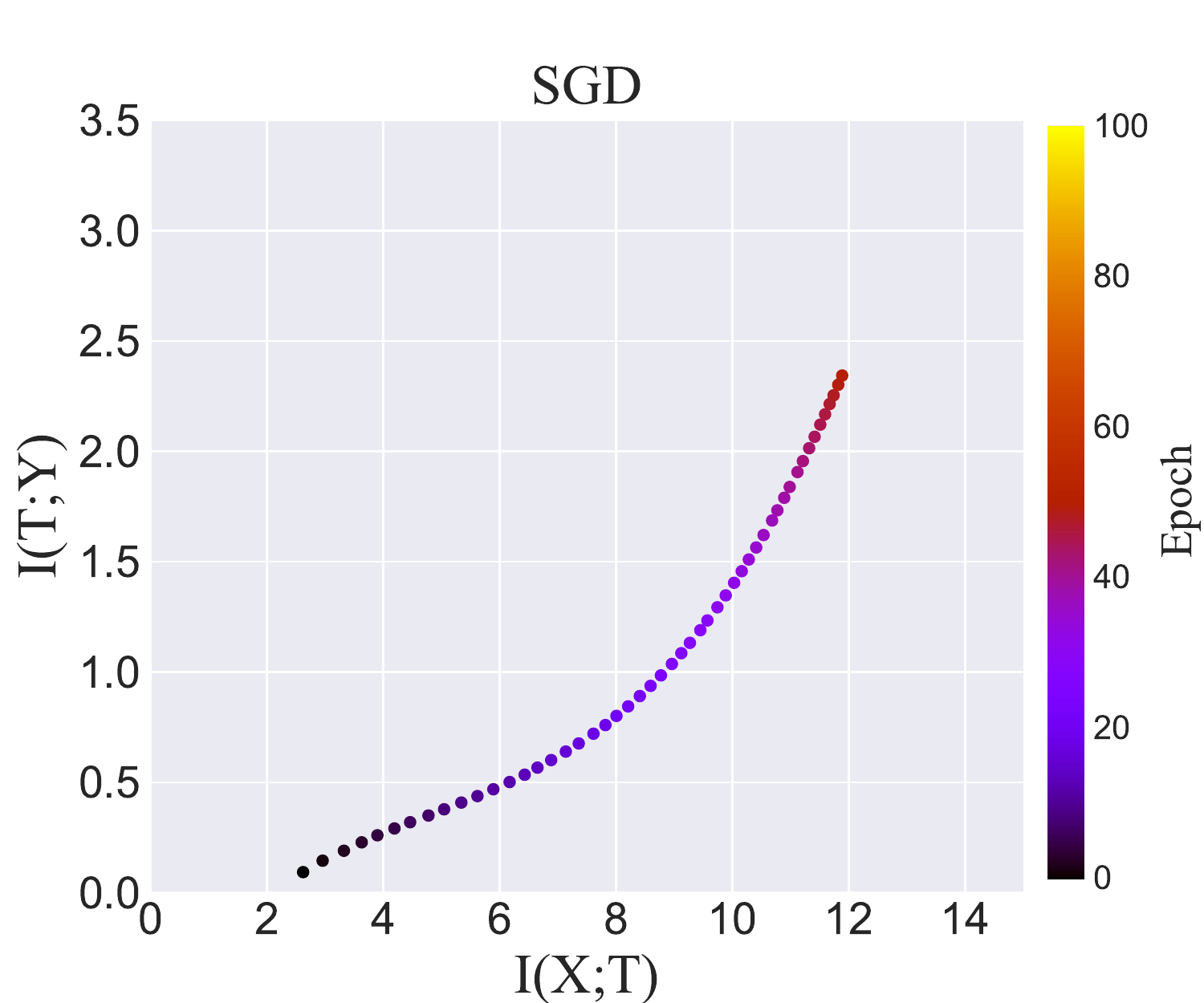}
			\includegraphics[width=0.24\textwidth]{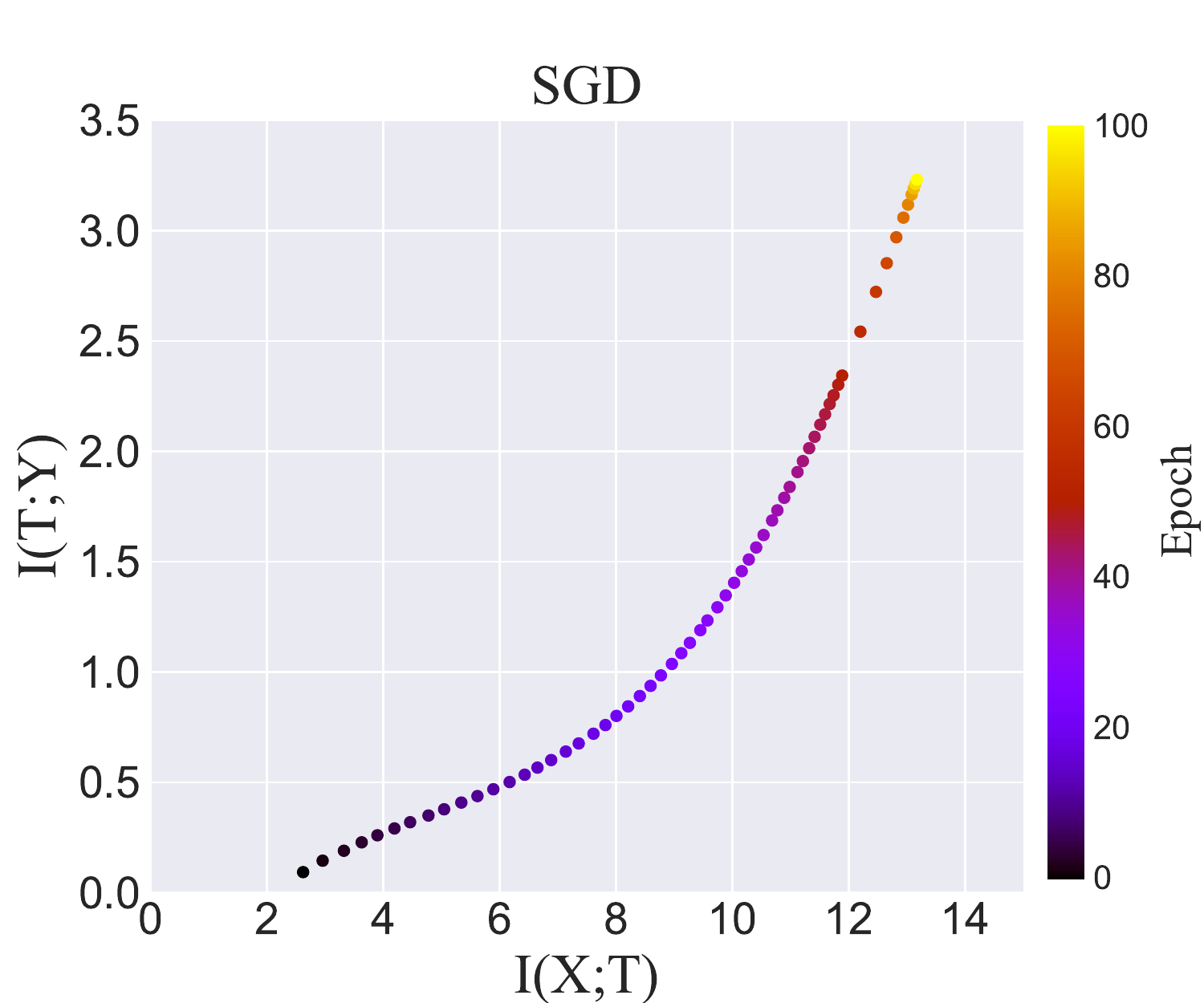}	
		\end{subfigure}
		\begin{subfigure}[t]{\textwidth}
			\includegraphics[width=0.24\textwidth]{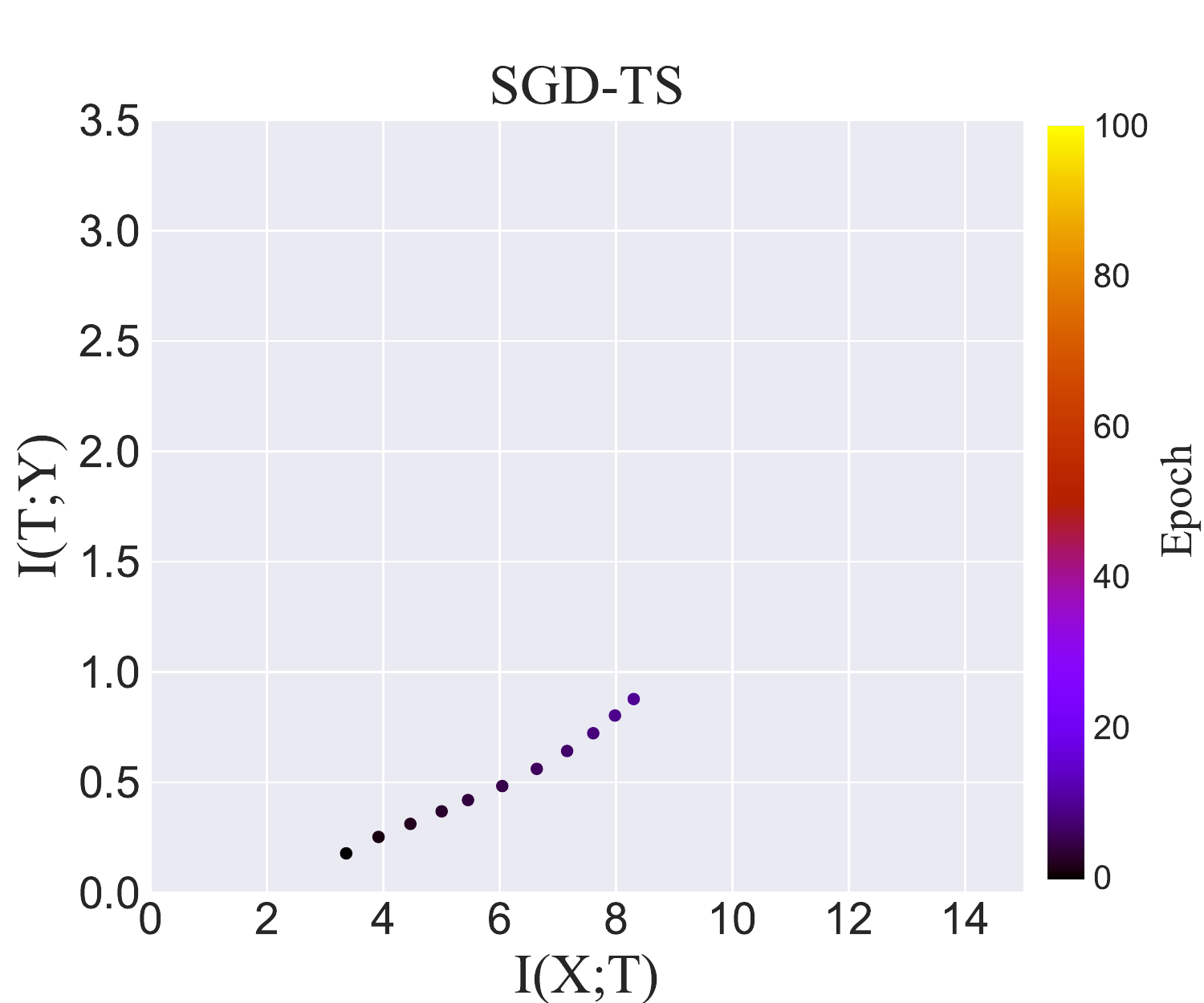}
			\includegraphics[width=0.24\textwidth]{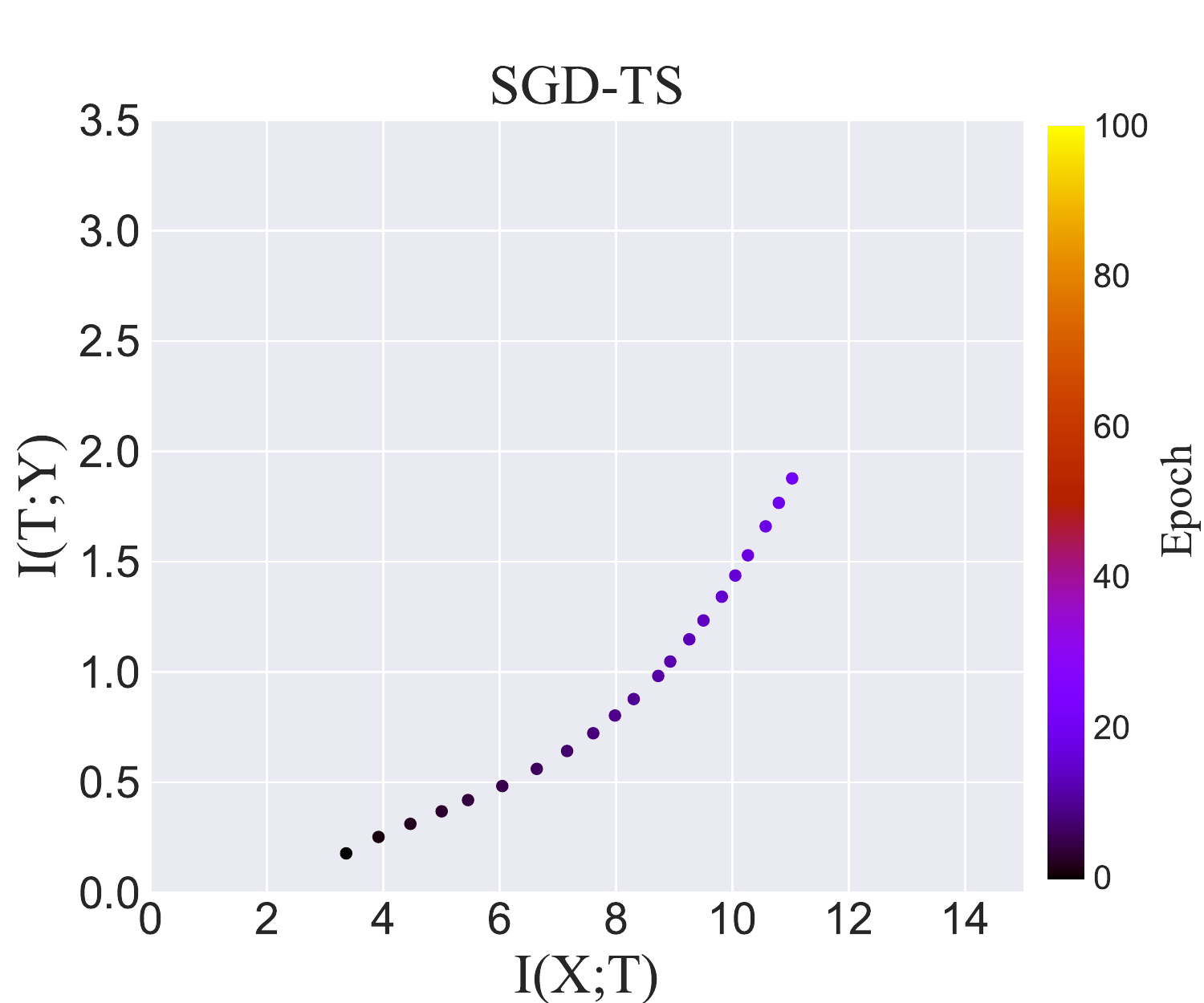}
			\includegraphics[width=0.24\textwidth]{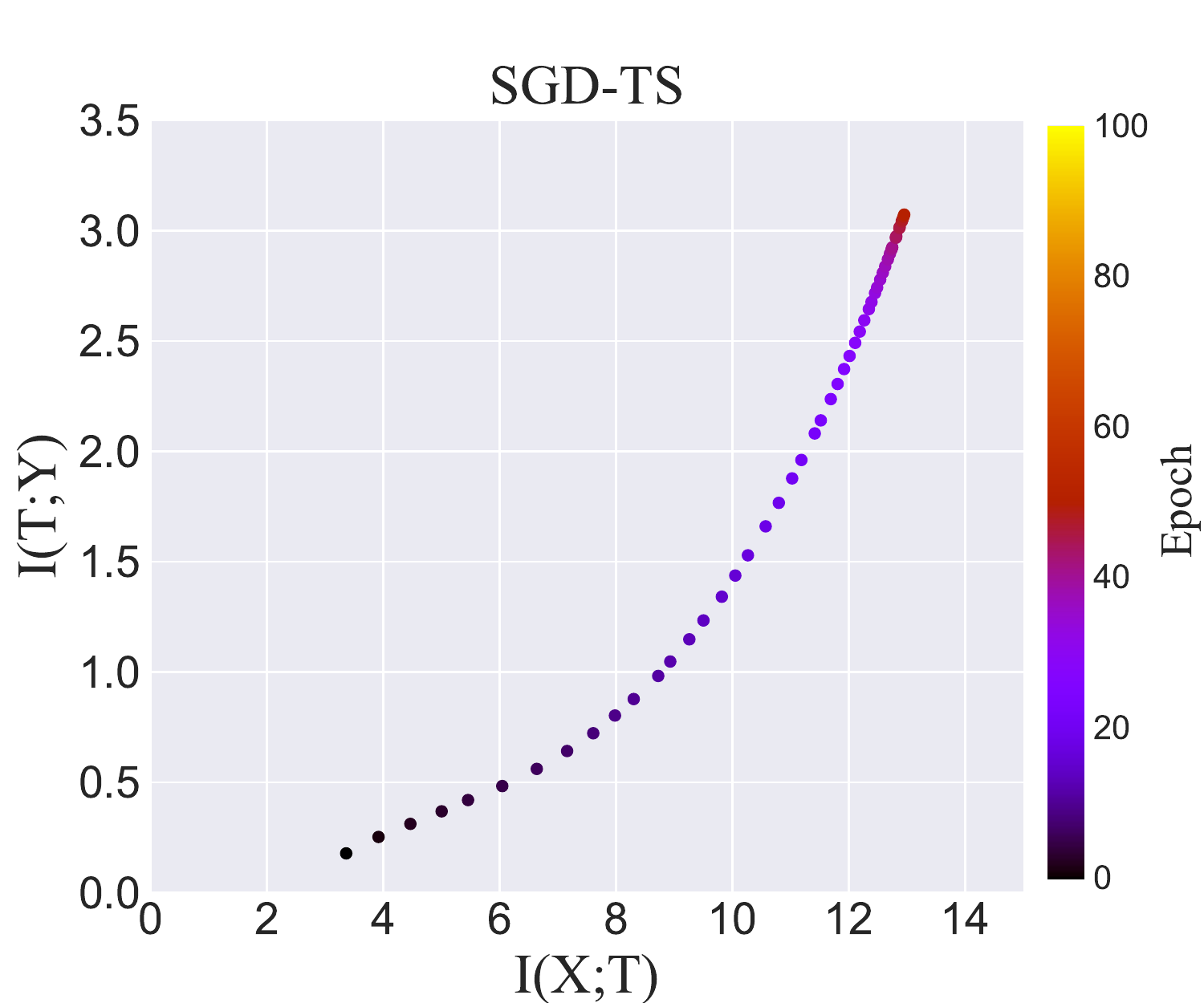}
			\includegraphics[width=0.24\textwidth]{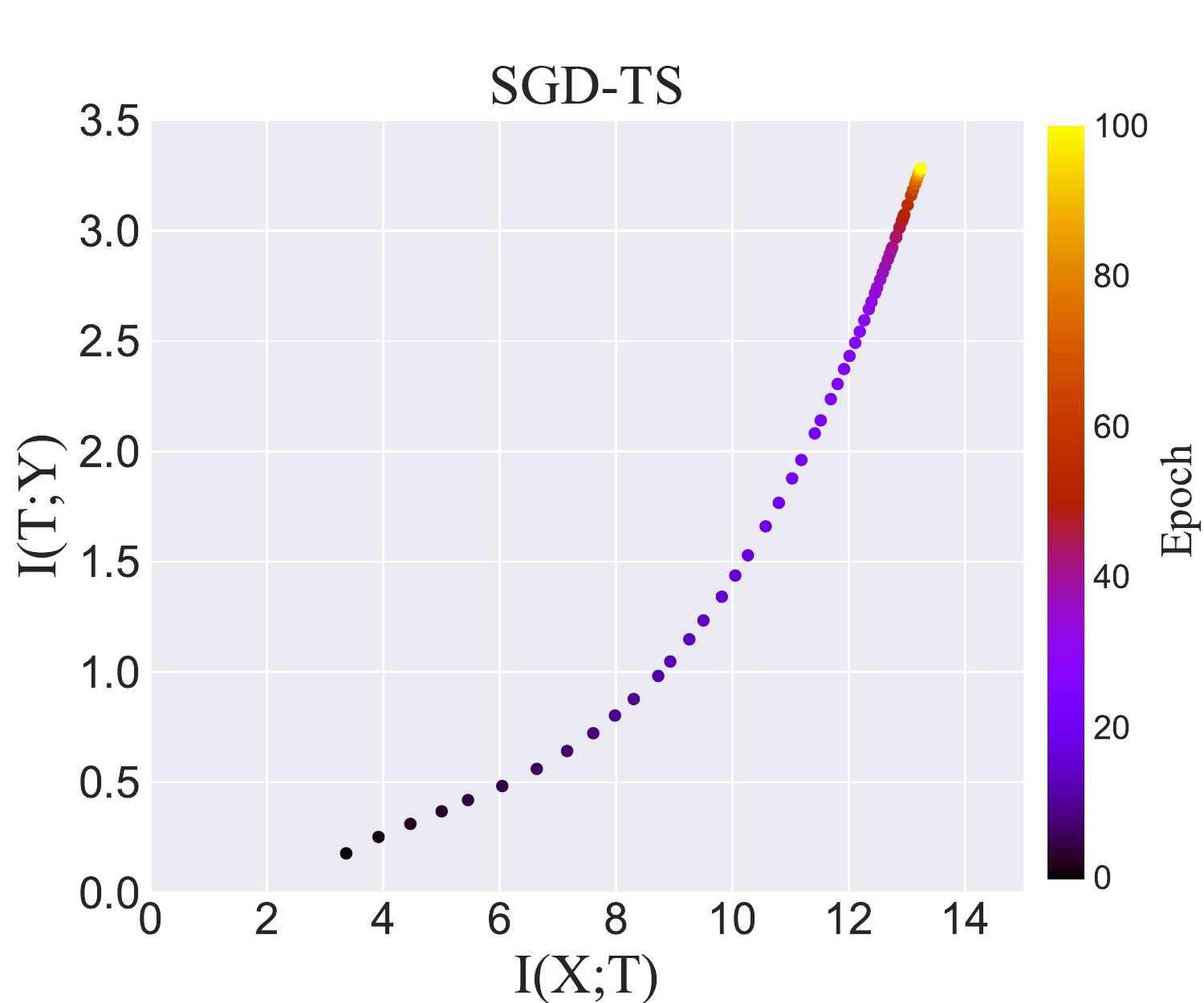}	
		\end{subfigure}
		\begin{subfigure}[t]{\textwidth}
			\setlength{\abovecaptionskip}{-0.0cm}
			\setlength{\belowcaptionskip}{-0.3cm}
			\subcaption*{Layer $6$}
			\includegraphics[width=0.24\textwidth]{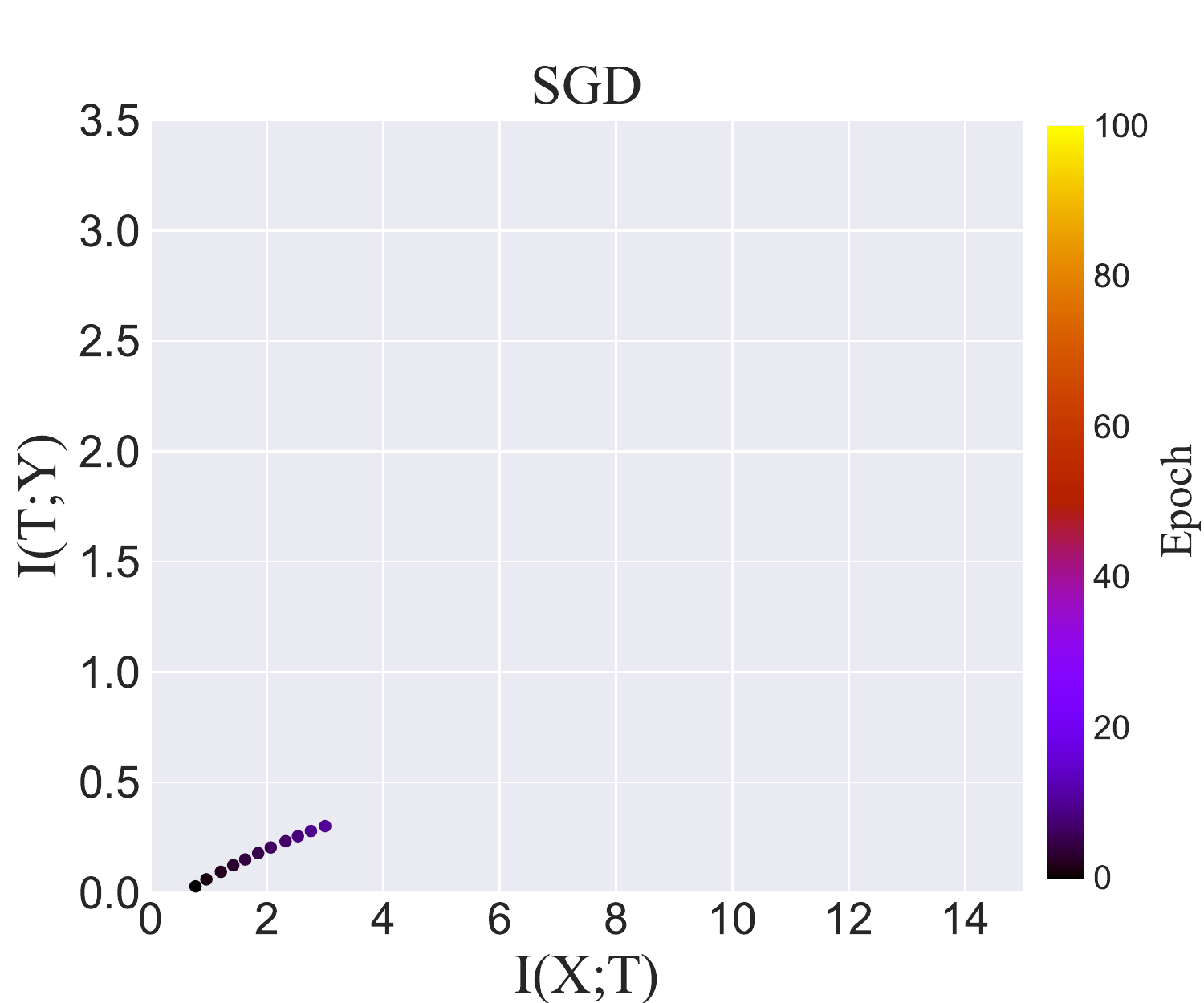}
			\includegraphics[width=0.24\textwidth]{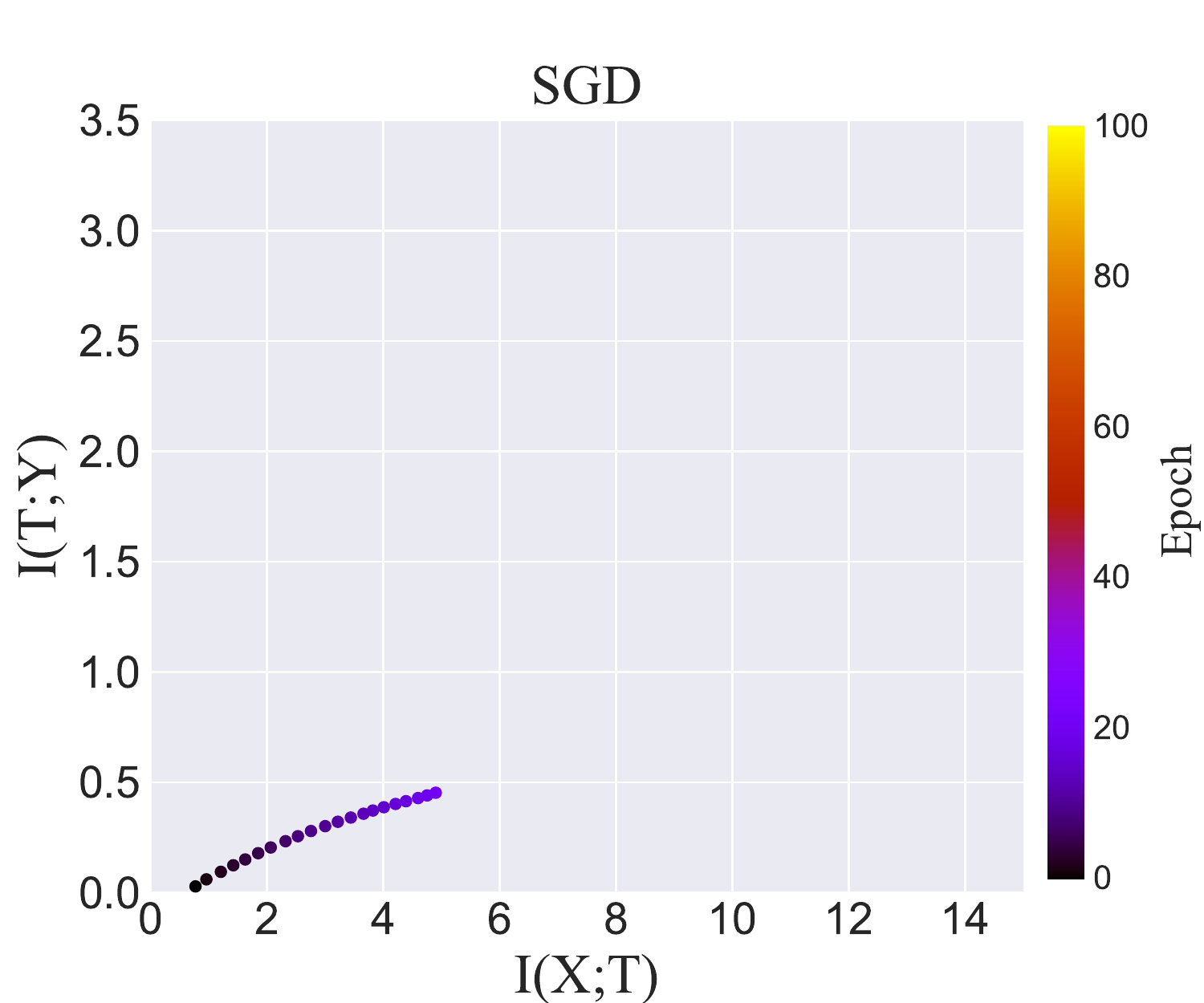}
			\includegraphics[width=0.24\textwidth]{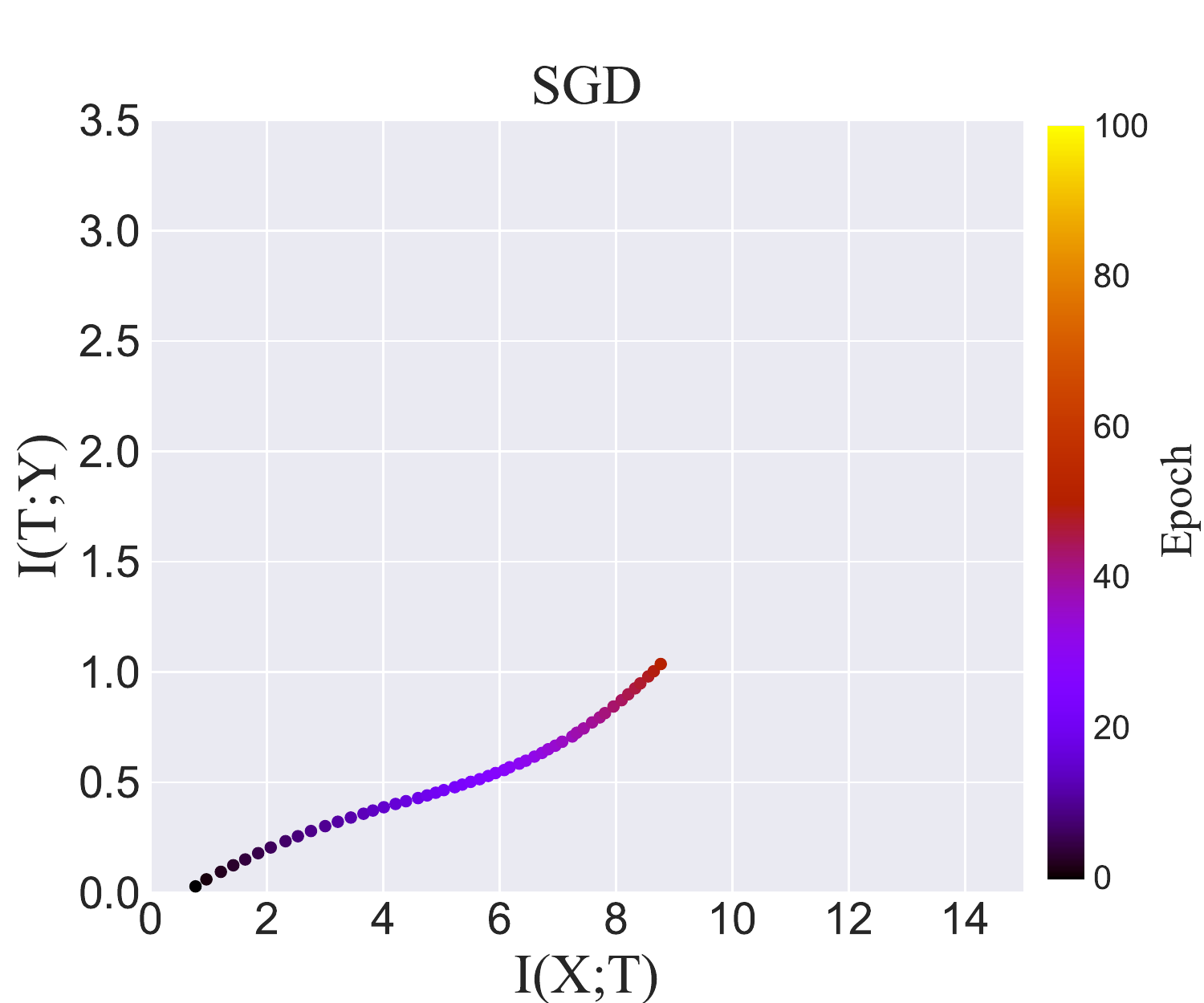}
			\includegraphics[width=0.24\textwidth]{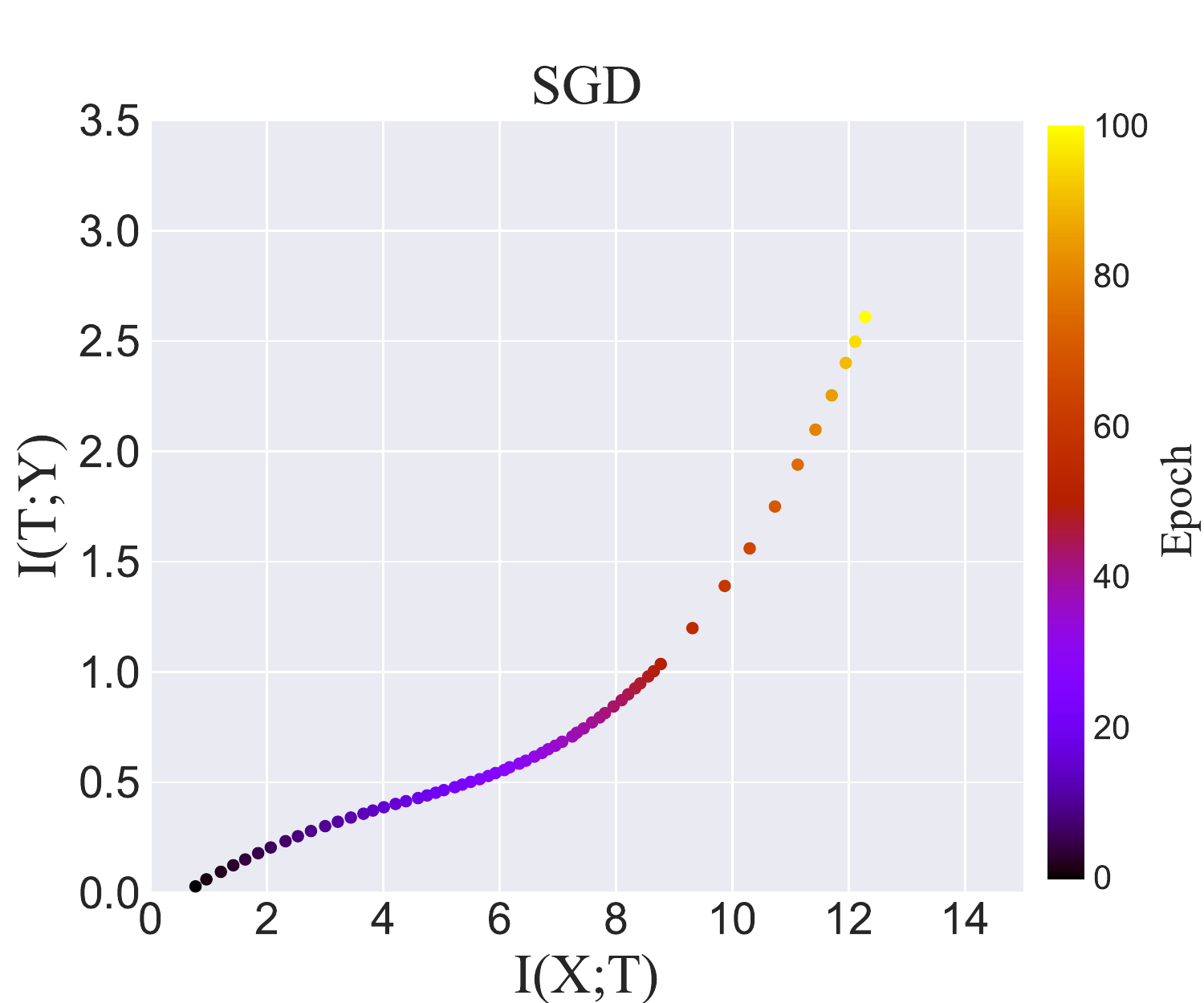}
		\end{subfigure}
		\begin{subfigure}[t]{\textwidth}
			\includegraphics[width=0.24\textwidth]{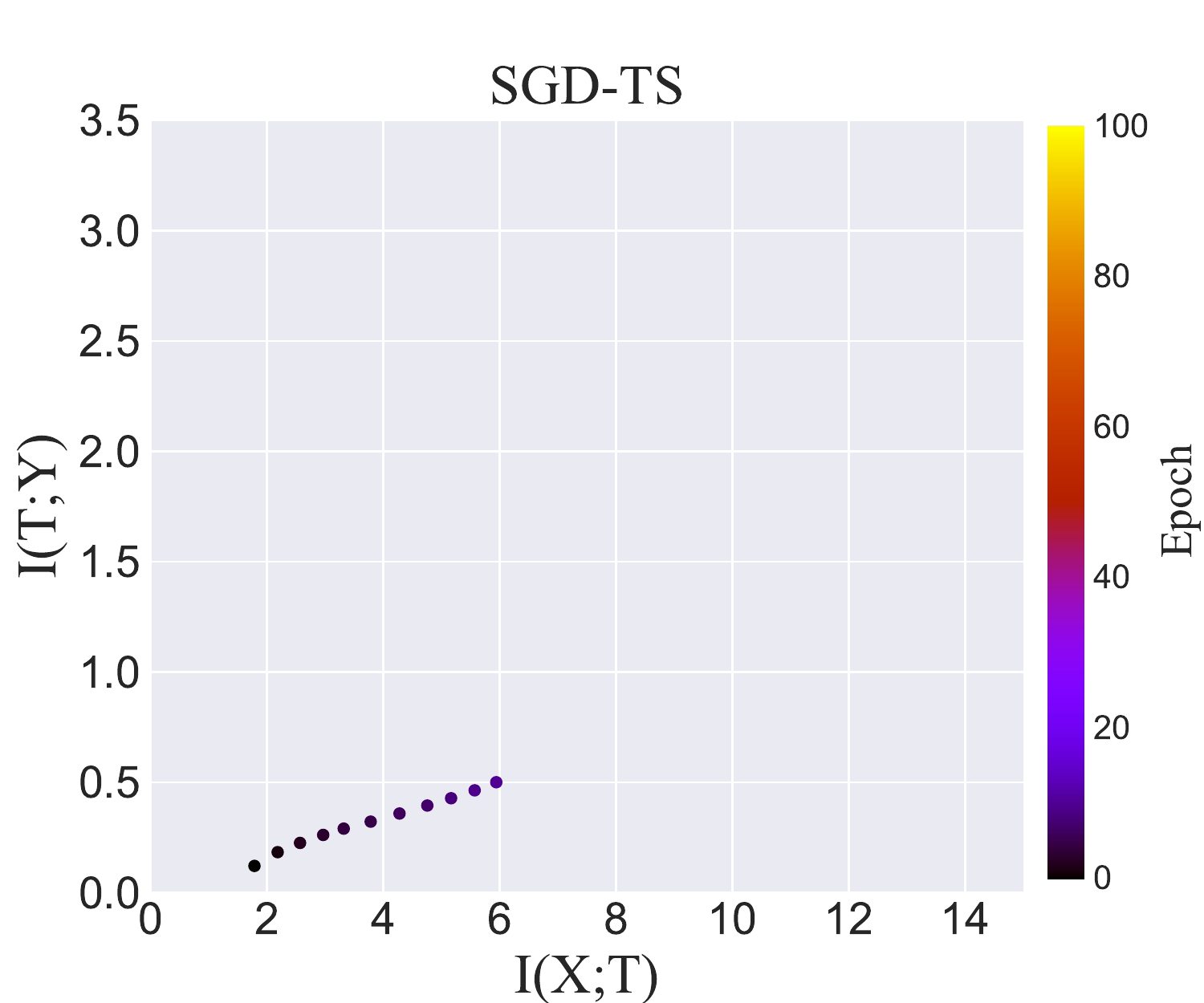}
			\includegraphics[width=0.24\textwidth]{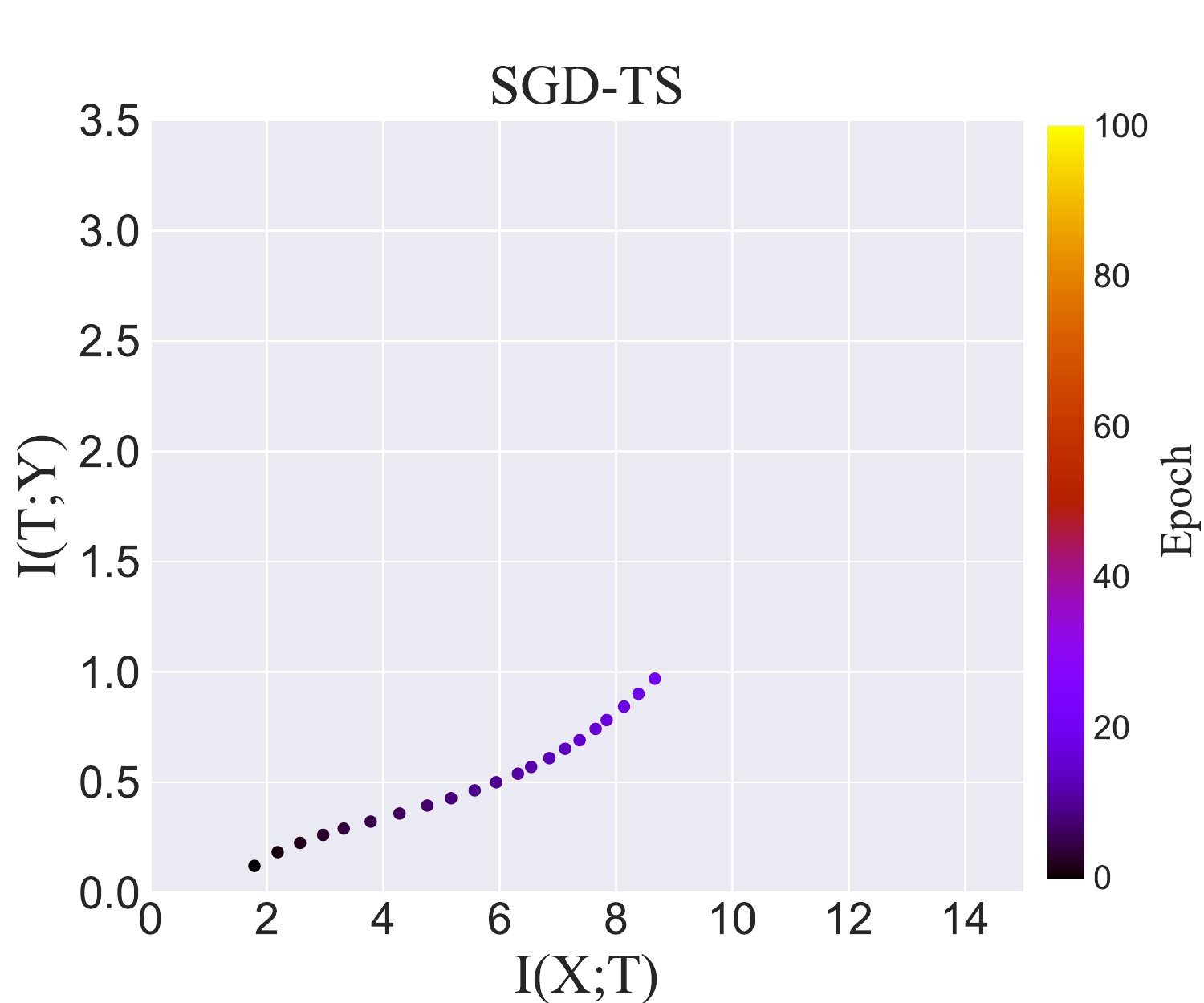}
			\includegraphics[width=0.24\textwidth]{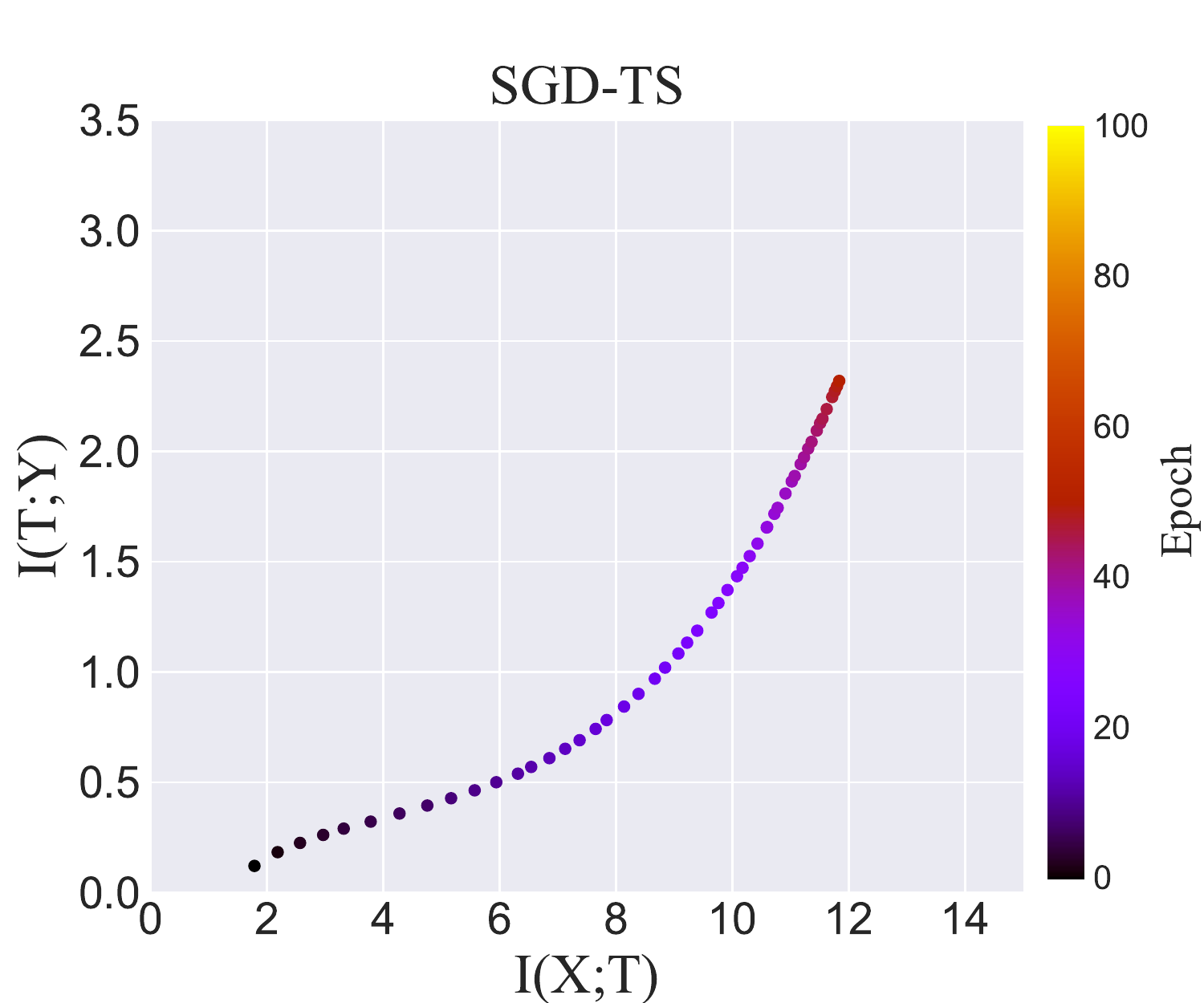}
			\includegraphics[width=0.24\textwidth]{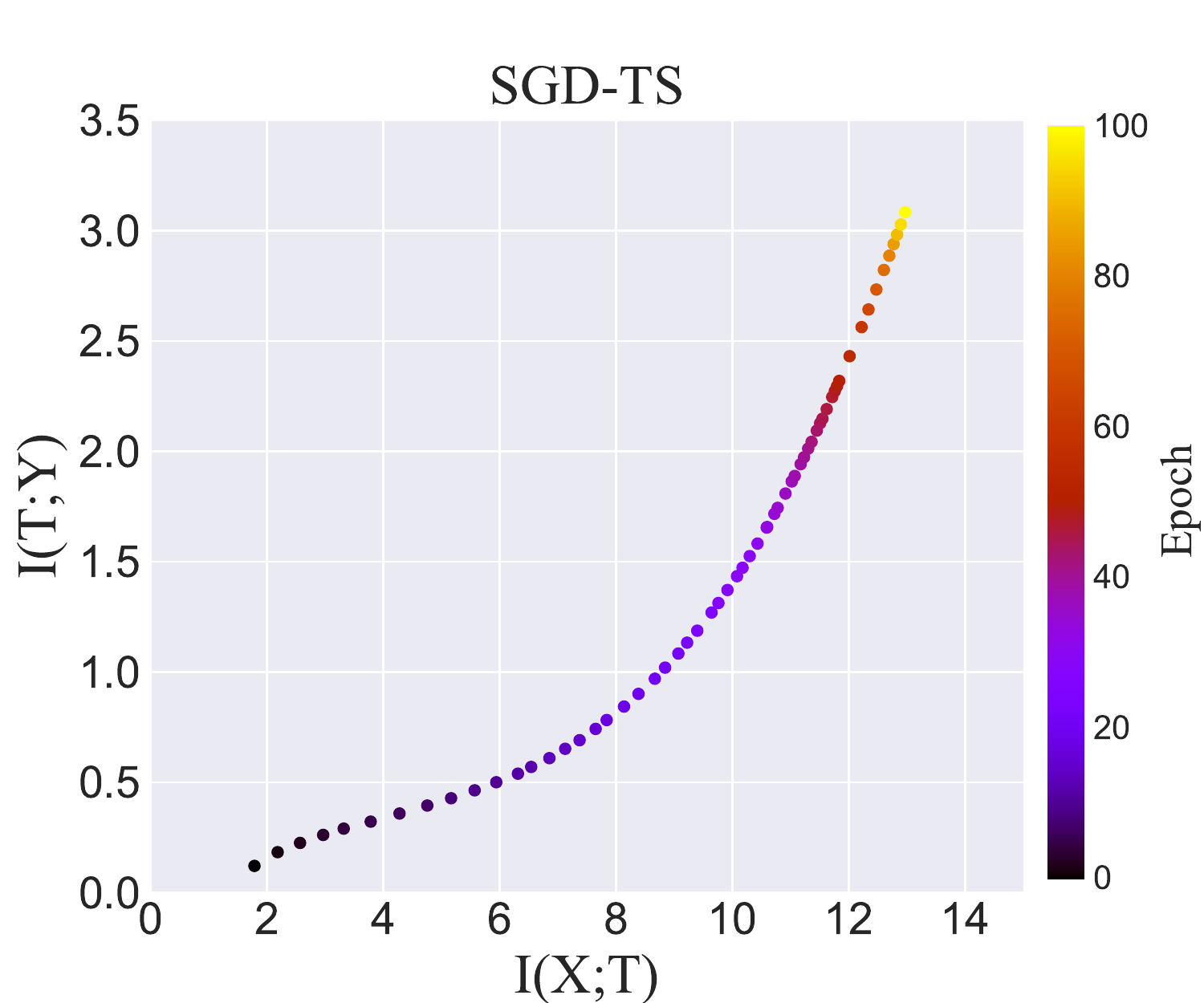}
		\end{subfigure}
		\begin{subfigure}[t]{0.98\textwidth}
			\centering
			\includegraphics[width=\textwidth]{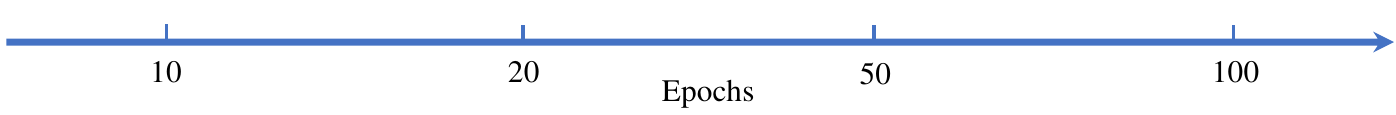}
		\end{subfigure}  
	\end{subfigure}
	\caption{The evolution of the last three layers information paths for CIFAR-$10$ dataset, during the optimization of conventional minibatch SGD and typical batch SGD. Left to right: at 10, 20, 50 and 100 epochs. For each layer, the first row shows the optimization process of conventional minibatch SGD and the second row shows the optimization process of typical batch SGD.}
	\label{fig:6}
\end{figure*}

{\color{black} In this section, we will present the experimental results on a more complex CIFAR-10 image classification dataset. This dataset contains $60000$ color images that are divided evenly into $10$ classes. Each tiny color image is of size $3*32*32$. In practice, we select $50000$ color images as the training set and the rest $10000$ images as the validation set. The model we train on CIFAR-10 is a multi-layer neural network of size 1024-200-200-100-50-20-20 and the activation function we use here is ReLU. Note that the first few layers barely move in the information plane during the convergence, so we only show the evolution of the last three layers information paths. The results are presented in Figure \ref{fig:6}. We can see that for all layers, the fitting phase is boosted by the typicality sampling scheme, which is consistent with our findings in Section \ref{section4}.}

\section*{Acknowledgment}
The authors would like to thanks Mr. Jiayang Li at Northwestern University for his helpful suggestions to further improve this paper.

\ifCLASSOPTIONcaptionsoff
  \newpage
\fi

\bibliographystyle{IEEEtran}
\bibliography{arXiv_typical_IB}




\end{document}